# APRICOT-Mamba: <u>A</u>cuity <u>Pred</u>iction in In<u>t</u>ensive Care Unit (ICU): Development and Validation of a Stability, Transitions, and Life-Sustaining Therapies Prediction Model

Miguel Contreras [1,6], Brandon Silva [2,6], Benjamin Shickel [3,6], Tezcan Ozrazgat-Baslanti [3,6], Yuanfang Ren [3,6], Ziyuan Guan [3,6], Jeremy Balch [4,6], Jiaqing Zhang [2,6], Sabyasachi Bandyopadhyay [5], Kia Khezeli [1,6], Azra Bihorac [3,6], Parisa Rashidi [1,6]**

[1]Department of Biomedical Engineering, University of Florida, Gainesville, FL, USA
[2]Department of Electrical and Computer Engineering, University of Florida, Gainesville, FL, USA
[3]Department of Medicine, University of Florida, Gainesville, FL, USA
[4]Department of Surgery, University of Florida, Gainesville, FL, USA
[5]Department of Medicine, Stanford University, Stanford, CA, USA
[6]Intelligent Clinical Care Center (IC3), University of Florida, Gainesville, FL, USA

**\*\* Correspondence:**
**Parisa Rashidi**
parisa.rashidi@ufl.edu

**Abstract**

The acuity state of patients in the intensive care unit (ICU) can quickly change from stable to unstable, sometimes leading to life-threatening conditions. On average, more than 5 million patients are admitted to ICUs in the US, with mortality rates ranging from 10 to 29%. Early detection of deteriorating conditions can result in providing more timely interventions and improved survival rates. While Artificial Intelligence (AI)-based models show potential for assessing acuity in a more granular and automated manner, they typically use mortality as a proxy of acuity in the ICU. Furthermore, these methods do not determine the acuity state of a patient (*i.e.*, stable, or unstable), the transition between acuity states, or the need for life-sustaining therapies. In this study, we propose APRICOT-M (Acuity Prediction in Intensive Care Unit-Mamba), a 150k-parameter state space-based neural network to predict acuity state, transitions, and the need for life-sustaining therapies in real-time in ICU patients. The model uses data obtained in the prior four hours in the ICU (including vital signs, laboratory results, assessment scores, and medications) as well as patient information obtained at admission (age, sex, race, and comorbidities) to predict the acuity outcomes in the next four hours. Our state space-based model can process sparse and irregularly sampled data without the need for manual imputation, reducing noise in input data and increasing deployment speed. We validated the APRICOT-M externally on data from hospitals not used in development (75,668 patients from 147 hospitals), temporally on data from a period not used in development (12,927 patients from one hospital in the period 2018-2019), and prospectively on data collected in real-time (215 patients from one hospital in the period 2021-2023) using three large datasets: the University of Florida Health (UFH) dataset, the electronic ICU Collaborative Research Database (eICU), and the Medical Information Mart for Intensive Care (MIMIC)-IV. The area under the receiver operating characteristic curve (AUROC) of APRICOT-M for mortality (external AUROC 0.94-0.95, temporal AUROC 0.97-0.98, and prospective AUROC 0.96-1.00) as well as for acuity (external AUROC 0.95-0.95, temporal AUROC 0.97-0.97, and prospective AUROC 0.96-0.96) shows comparable results to state-of-the-art prediction models. Furthermore, APRICOT-M can make predictions on transitions to instability (external AUROC 0.81-0.82, temporal AUROC 0.77-0.78, and prospective AUROC 0.68-0.75) and predictions on the need for life-sustaining therapies, such as the need for mechanical ventilation (external AUROC 0.82-0.83, temporal AUROC 0.87-0.88, and prospective AUROC 0.67-0.76), and the need for vasopressors (external AUROC 0.81-0.82, temporal AUROC 0.73-0.75, prospective AUROC 0.66-0.74). In both cases, its performance is comparable to state-of-the-art models. This tool allows for real-time acuity monitoring in critically ill patients and could help clinicians make timely interventions. Furthermore, the model predicts the transition between acuity states and can suggest life-sustaining therapies that the patient might need within the next four hours in the ICU.

## 1. Introduction

On average, more than 5 million patients are admitted annually to intensive care units (ICU) in the US [1], with mortality rates ranging from 10 to 29% depending on the patient's age, comorbidities, and illness severity [2]. In the context of healthcare, patient acuity implies the categorization of patients based on their care needs [3]. During an ICU stay, the acuity state of patients can quickly change, leading to life-threatening conditions. Early detection of such deteriorating conditions can result in providing more timely interventions and improved survival rates.

Given the frequent pace of monitoring in the ICU, many physiological and clinical measurements are recorded as time series, such as the vital signs, laboratory test results, medications, and assessment scores. The rich temporal data, in conjunction with static admission data (*e.g.*, age, gender, or comorbidities), provide transformative opportunities for developing data-driven and real-time approaches for assessing patient acuity and augmenting clinicians' decisions.

Several tools have been traditionally used to assess patient acuity in the ICU, such as Sequential Organ Failure Assessment (SOFA) [4], Acute Physiology And Chronic Health Evaluation (APACHE) [5], Simplified Acute Physiology Score (SAPS) [6], [7], and Modified Early Warning Score (MEWS) [8]. More recent approaches have used deep learning methods to exploit the vast amount of Electronic Health Record (EHR) data for automated acuity assessment. Such approaches have focused on mortality prediction as a proxy for real-time patient acuity scores [9], [10], [11] and have been shown to outperform manual scores while providing more timely and frequent acuity assessments. Despite their high performance and granularity, existing deep learning methods have the disadvantage of solely predicting mortality risk. Incorporating scores that capture a wider spectrum of patient acuity states besides the mortality risk can provide a more complete picture and better opportunities for timely interventions.

A recent computational phenotyping method developed by Ren et al. [12] labeled patient acuity states based on the use of four life-sustaining therapies: mechanical ventilation (MV), massive blood transfusion (BT) (defined as at least 10 units in past 24 hours), vasopressors (VP), and continuous renal replacement therapy (CRRT). A significant number of patients in the ICU require MV, with an incidence rate of 20-40% [1], while 30-50% of ICU patients require BT [13]. Vasopressors (VP) are used to manage blood pressure in about 27% of ICU patients [14], and CRRT is needed in patients with acute kidney injury, up to 14% of ICU patients [15]. Patient was considered as unstable if patient required at least one of these life supportive therapies, and was considered stable otherwise. These states, along with discharge and mortality indicators, provide a wider range for patient acuity assessment. Timely clinical intervention would require predicting such states in advance based on the more subtle changes in the patient's physiological state.

Recent advances in deep learning have shown that the Transformer architecture [16] can be very effective for time-series classification tasks [17], [18]. Specifically, this architecture has been adapted for clinical time-series with the development of novel embeddings that can handle the irregular sampling of this type of data [10], [19]. However, the emergence of state space models (SSM) has provided an alternative to the Transformer architecture with increased speed and performance on tasks with longer sequence lengths while using less parameters [20]. The use of this type of model can potentially lead to improved performance in patient acuity prediction compared to the existing deep learning methods.

In this study, we propose APRICOT-M (Acuity Prediction in Intensive Care Unit-Mamba), a novel state space-based model to predict real-time acuity (stable, unstable, deceased, discharge) in the ICU. APRICOT-M uses temporal data from the prior four hours in the ICU along with static data obtained at admission time to predict patient acuity within the next four hours in the ICU. Specifically, the algorithm predicts the probability of a patient being stable or unstable in the next four hours, along with the probability of mortality or being discharged. Furthermore, it predicts transitions between acuity states (*i.e.*, transition from stable to unstable, and transition from unstable to stable) and the need for three life-sustaining therapies: MV, VP, and CRRT. This algorithm provides the advantage of not requiring imputation for the temporal data, reducing input noise, and optimizing real-time deployment of the model. The APRICOT-M model was developed on data obtained from the University of Florida Health (UFH) between 2014-2017 and 52 hospitals from the electronic ICU Collaborative Research Database (eICU) [21]. The model was then externally validated on data from hospitals not used in the development phase consisting of the remaining 146 hospitals from eICU and data from the Medical Information Mart for Intensive Care (MIMIC)-IV [22], and temporally validated on data from a period not used in the development phase consisting of UFH data between 2018-2019. Finally, prospective validation was performed with data acquired in real-time from UFH between May 2021 and October 2023. Additionally, a Transformer variant of the model, APRICOT-T, was also developed and used along with three baseline models to demonstrate the advantages of the novel architecture. To the best of our knowledge, the proposed APRICOT-M model is the first real-time acuity prediction tool for patients in the ICU that incorporates prediction of stable and unstable states, mortality, and discharge, transition in patient acuity states, and the need for life-sustaining therapies, while being extensively validated on three large datasets externally, temporally, and prospectively. Furthermore, it is the first state space-based model to be applied to clinical data.

## 2. Methods

2.1 Data and Study Design

Three databases were used in this study: UFH, MIMIC, and eICU (cohort diagrams in Fig. 1). The UFH dataset was collected in both a retrospective (UFH-R) and prospective (UFH-P) manner, while the MIMIC and eICU datasets were publicly available and collected retrospectively. The UFH-R dataset was retrieved from the UF Integrated Data Repository and included adult patients admitted to the ICUs at the University of Florida (UF) Health Shands Hospital, Gainesville between 2014 to 2019, while UFH-P data obtained in real-time from adult patients admitted to the ICU at UFH between May 2021 and October 2023 (UFH-P). The MIMIC dataset is a publicly available dataset collected at the Beth Israel Deaconess Medical Center from 2008 to 2019 [22]. The eICU dataset contains data from ICU patients in 208 hospitals in the midwest, northeast, south, and west regions of the US from 2014 to 2015 [21]. In all four datasets, patients were excluded if they had missing records of basic information (*i.e.*, age, discharge location, sex, race, or BMI), outcomes (*i.e.*, acuity, transitions, life-sustaining therapies), or at least one of six routine vitals: heart rate (HR), respiratory rate (RR), systolic blood pressure (SBP), diastolic blood pressure (DBP), body temperature (Temp), and oxygen saturation ($SPO_2$). Also, patients who stayed in the ICU for less than 12 hours or more than 30 days were excluded. Data from UFH-R between 2014 and 2017 and data from 52 hospitals from the eICU database were used to create the development set. This set was used to train and tune the APRICOT-M model. Data from MIMIC and the remaining 146 hospitals in eICU were used for external validation to evaluate the generalizability of APRICOT-M to diverse hospital settings. Hospitals from eICU used for development contained at least 800 patients to have the largest number of patients possible in the fewest number of hospitals, allowing for a larger number of hospitals in the external validation set. Data from UFH-R between 2018 and 2019 was used for temporal validation to evaluate generalizability to different years of admission. Data from UFH-P was used for prospective validation to evaluate the application of APRICOT-M to a real-time setting.

The study design consisted of six steps: development, calibration, validation, subgroup analysis, interpretation, and prospective evaluation. This study follows the Transparent Reporting of a multivariable prediction model for Individual Prognosis Or Diagnosis (TRIPOD) Statement [23] (checklist available in Supplementary Material). First, we developed and tuned the APRICOT-M model on the development dataset (80% training, 20% validation), followed by calibration using a random sampling of 10% of each validation set, to obtain a risk probability more relevant to the true incidence probabilities of each outcome. Next, the model was externally validated on hospitals not included in the development stage, and temporally validated on data from years not included in the development set. We also performed a subgroup analysis of model performance based on age group (18-60 vs over 60), sex, and race, accompanied by detailed interpretability analyses, to find the most relevant predictors correlated with increased patient acuity. Finally, prospective validation of the model was performed on the data acquired using our clinical real-time platform [24] interfaced with the Epic® EHR system. The data was collected in real-time and retrieved retrospectively for model evaluation.

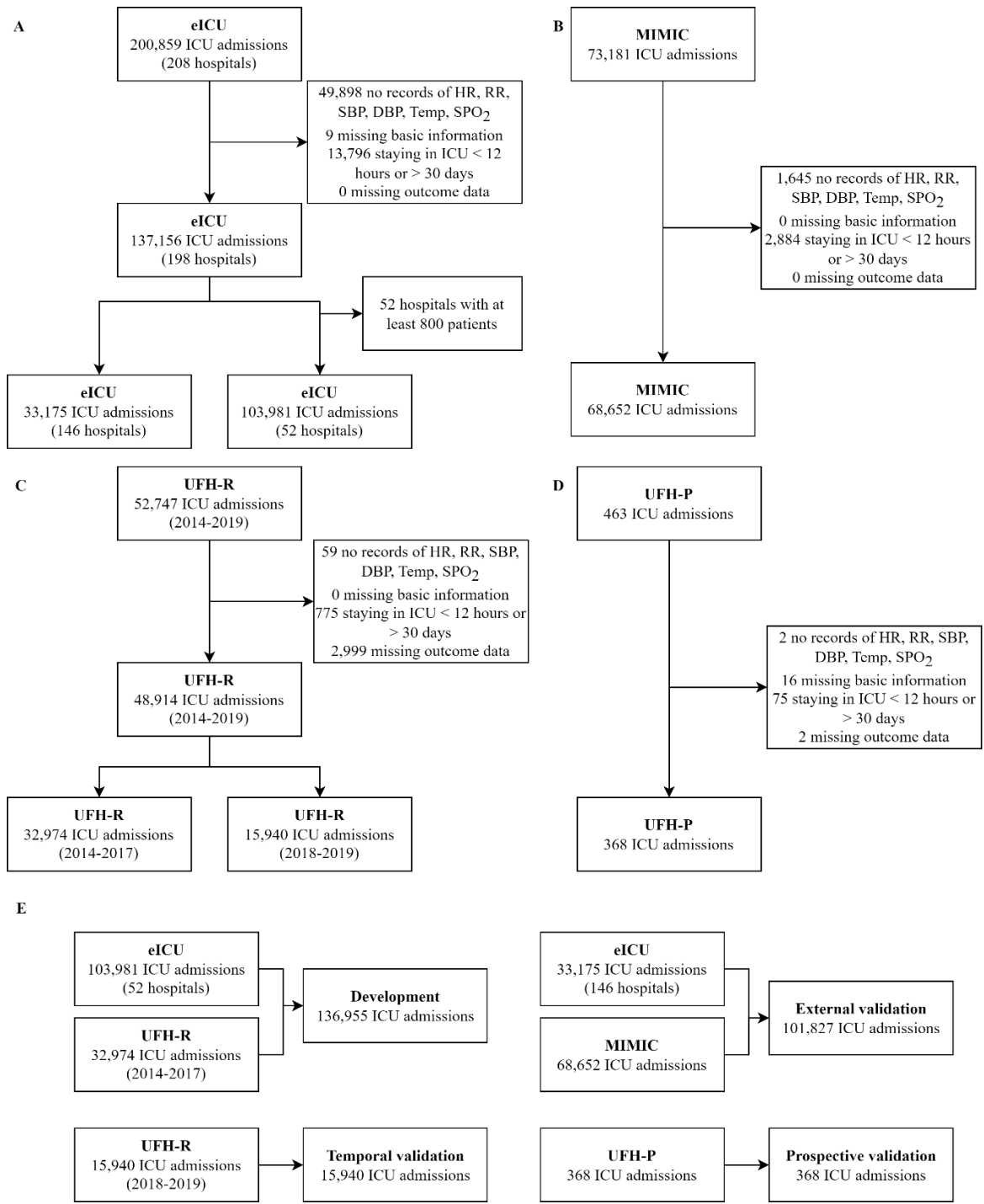

**Fig. 1 | Cohort flow diagram.** The (A) eICU, (B) MIMIC, (C) UFH-R, and (D) UFH-P datasets. (E) Final datasets were assembled from the four datasets for development (training and validation), external validation, temporal validation, and prospective validation. We created the development dataset by pooling data from 52 random eICU hospitals with data from UFH-R between 2014-2017. Data from the remaining 146 hospitals at eICU were pooled with data from MIMIC to form the external validation set. The temporal validation dataset was created from UFH-R between 2018 and 2019. The prospective validation dataset was created from real-time data in the UFH-P dataset.

## 2.2 Ethics Approval and Patient Consent

Retrospective data from UFH (UFH-R) was obtained by the University of Florida Institutional Review Board and Privacy Office approval as an exempt study with a waiver of informed consent (IRB202101013). Prospective data from UFH was obtained by consent under the approval granted by the University of Florida Institutional Review Board under the numbers IRB201900354 and IRB202101013. Before enrolling patients in the study, written informed consent was obtained from all participants. In cases where patients could not provide informed consent, consent was obtained from a legally authorized representative (LAR) acting on their behalf. Eligible participants were individuals aged 18 and older who were admitted to the ICU and expected to remain there for at least 24 hours. Patients who could not provide an LAR or self-consent were expected to be transferred or discharged from the ICU within 24 hours, and those necessitating contact or isolation precautions were excluded. Also excluded from this study were patients who expired within 24 hours of recruitment. The analysis using the eICU dataset is exempt from institutional review board approval due to the retrospective design, absence of direct patient intervention, and security schema. The data in the MIMIC dataset is de-identified, and the institutional review boards of the Massachusetts Institute of Technology and Beth Israel Deaconess Medical Center both approved using the database for research.

## 2.3 Features and Outcomes

The primary outcome predicted by APRICOT-M is the acuity state of a patient in the ICU in the next four hours obtained using computable phenotyping, defined by four levels of severity from the least to the most severe: discharge, stable, unstable, and deceased [12]. Discharge and deceased states were determined through indicators of whether a patient was discharged alive or passed away at the end of their stay in the ICU (Fig. 2d). Unstable and stable are determined by the presence of one of four life-sustaining therapies. Specifically, if a patient is under continuous renal replacement therapy (CRRT), on vasopressors (VP), on invasive mechanical ventilation (MV), or has had a massive blood transfusion (>10 units in the last 24 hours) (BT), the patient is considered unstable for the time intervals in which any of the therapies was present (Fig. 2c). Otherwise, the patient is considered stable. Given the inertia that is usually seen in clinical states (*i.e.*, tendency to remain unchanged), there is an imbalance in the number of unstable and stable cases that represent a transition in acuity state. Therefore, transitions to instability (*i.e.*, Stable-Unstable) and to stability (*i.e.*, Unstable-Stable) were labeled and added as outcomes to remove potential bias towards predicting the same state (*e.g.*, patient which is currently stable, and the model predicts the next state to be stable as well). Furthermore, onset of life-sustaining therapies (CRRT, MV, and VP) were labeled as well to predict potential need for one or more of these therapies (Fig. 2d). In the case of BT, prediction is not performed given the label is generated using information of the last 24 hours which would mean the patient has undergone the therapy by the time of prediction.

In order to predict the outcome for the next four hours, static patient information and clinical events in the previous four hours are used as predictive features. Clinical events consist of the timestamp converted to hours since admission, a variable code, and the value of the measured variable. In the case multiple features happen at the same time point, they are arbitrarily ordered. Events are assigned to their corresponding observation window, where an observation window is comprised of all clinical events in the 4 hours prior to the start of the current prediction window. The prediction window represents a four-hour interval in which the outcome is assessed (Fig. 2b). In particular, vital signs, medications, laboratory tests, and assessment scores are extracted from the previous four hours of ICU monitoring, while demographic and comorbidity information are extracted from patient history information. Variables occurring in less than 5% of all ICU stays are removed. Outliers are also removed based on an upper (99th percentile) and lower (1st percentile) bound (Fig. 2a). A complete list of the variables used for prediction can be found in the Supplementary Material (Supplementary Table S1). Variables are then scaled according to their specific range using minimum-maximum scaling based on the minimum and maximum values for each feature in the development cohort (Fig. 2b). Static variables are first imputed using the population mean for each feature and subsequently scaled using minimum-maximum scaling (both based in the development cohort) (Fig. 2b). Finally, the APRICOT-M model, which uses the embedding schema of a clinical time-series model [19] with incorporation of a state-space model (Mamba block, details in section 2.4) [20], is trained to predict probabilities for each acuity outcome, transition, and life-sustaining therapy onset and evaluated against ground truth labels (Fig. 2e).

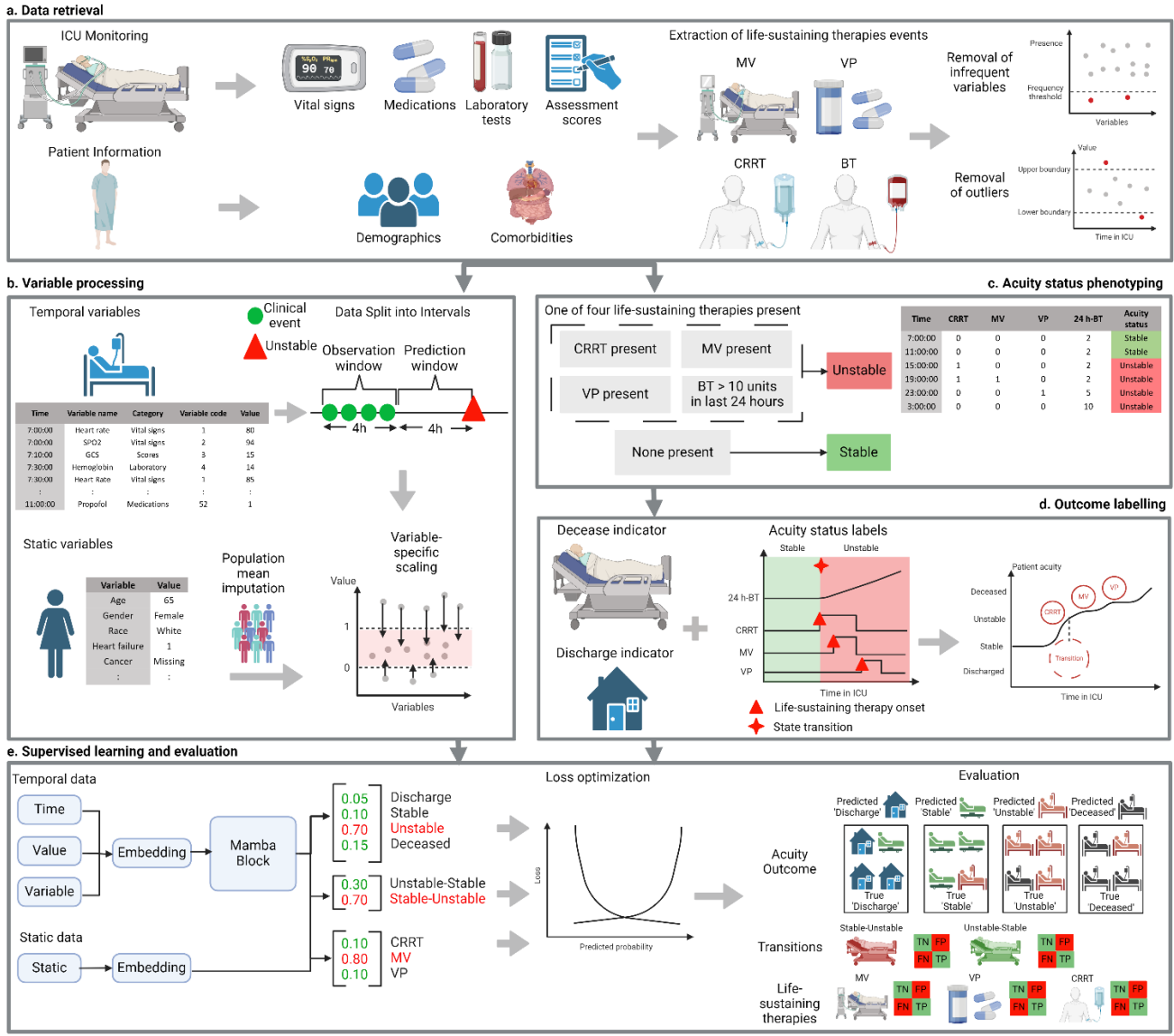

**Fig. 2 | APRICOT-M model development overview. (a)** Data from three different Intensive Care Unit (ICU) datasets were used for training and validating the APRICOT-M model (200 hospitals, 193,597 patients total). Each dataset consisted of ICU stay information (*i.e.*, vital signs, medications, laboratory test results, and assessment scores) and admission information (*i.e.*, patient demographics and comorbidities). The presence of four life-sustaining therapies events were extracted: mechanical ventilation (MV), vasopressors (VP), continuous renal replacement therapy (CRRT), and massive blood transfusion (BT). Highly infrequent and outlier features are removed. **(b)** Temporal variables are processed into a list of clinical events for each patient, where each variable is assigned a code according to its order of appearance. Clinical events are assigned to 4-hour observation windows and used in conjunction with static patient information (which is imputed using the population mean) to predict the outcome in the next 4-hour window. After data is split into time windows, each variable is scaled. **(c)** The acuity status was determined every 4 hours in the ICU, with a patient being labeled as 'unstable' for a 4-hour interval if at least one of the four life-sustaining therapies extracted was present. **(d)** The outcome for each 4-hour window was determined by combining acuity status labels with decease/discharge indicators, providing a complete acuity trajectory for each patient including transition between states and onset of three life-sustaining therapies: MV, VP, and CRRT. **(e)** A state space-based model was employed for acuity, transition, and life-sustaining therapy onset prediction, with two separate embeddings used for temporal and static data. The first

embedding is created by a Mamba block which takes the sequence of clinical events, from which the output is then combined with the second embedding, which is created from static patient data, to calculate the probabilities of each acuity outcome, transition, and life-sustaining therapy. The loss of the model is optimized by comparing the predicted probabilities with the ground truth labels. Finally, the model is evaluated by calculating the classification error for each outcome individually.

2.4 Model Development and Performance

After splitting the three datasets into development, external, temporal, and prospective, we further split the development dataset by randomly selecting 80% of the patients for training, and 20% for validation and tuning hyperparameters. The APRICOT-M model was developed based on the embedding schema of a clinical Transformer architecture [19] with the incorporation of a Mamba block [20]. The original architecture is modified to create an embedding for temporal data using a sequence of triplets of time, value, and variable code recordings. The time and value recordings are passed through one-dimensional (1D)-Convolutional layers and the variable code recordings through an Embedding layer. The results are fused as one temporal embedding through addition and positional encoding is incorporated to preserve the sequence order information. This approach removes the need for imputation that other time-series models require to maintain fixed dimensions in the input. The embedding schema presented here uses padding/truncation to match the maximum sequence length set for the model, so no feature values are imputed. For static data, an embedding is created using two linear layers. A Mamba block is then employed to add context to the temporal embedding, from which the resulting vector is combined with the static embedding. Finally, the fused vector is passed through individual linear layers to calculate the probability of each acuity primary outcome, transition, and life-sustaining therapy (Fig. 3).

To predict the overall acuity status, a logic schema (shown in Fig. 4) is implemented to the model outputs. For training the model, class weights are applied to the loss of each outcome to account for the class imbalance. Four algorithms were used as baseline models for comparison: categorical boosting (CatBoost) [25], gated recurrent unit (GRU) [26], a vanilla Transformer [16], and APRICOT-T. For CatBoost, statistical features (*i.e.*, minimum, maximum, mean, and standard deviation values) were extracted to convert temporal data into a static representation. For the GRU and vanilla Transformer, temporal features were resampled to 20-minute resolution and tabularized to have the same sequence length and dimensions. For APRICOT-T, the same data processing as APRICOT-M is applied as well as the same architecture with the only difference of replacing the Mamba blocks with Transformer encoder layers. For evaluation, six metrics were used: the Area Under the Receiver Operating Characteristic (AUROC), the Area Under the Precision-Recall Curve (AUPRC), sensitivity, specificity, Positive Predictive Value (PPV), and Negative Predictive Value (NPV). The Youden index [27] was used to find the optimal threshold to calculate sensitivity, specificity, PPV and NPV. Both AUROC and AUPRC were used as metrics to tune the model architecture and hyperparameters on the development test.

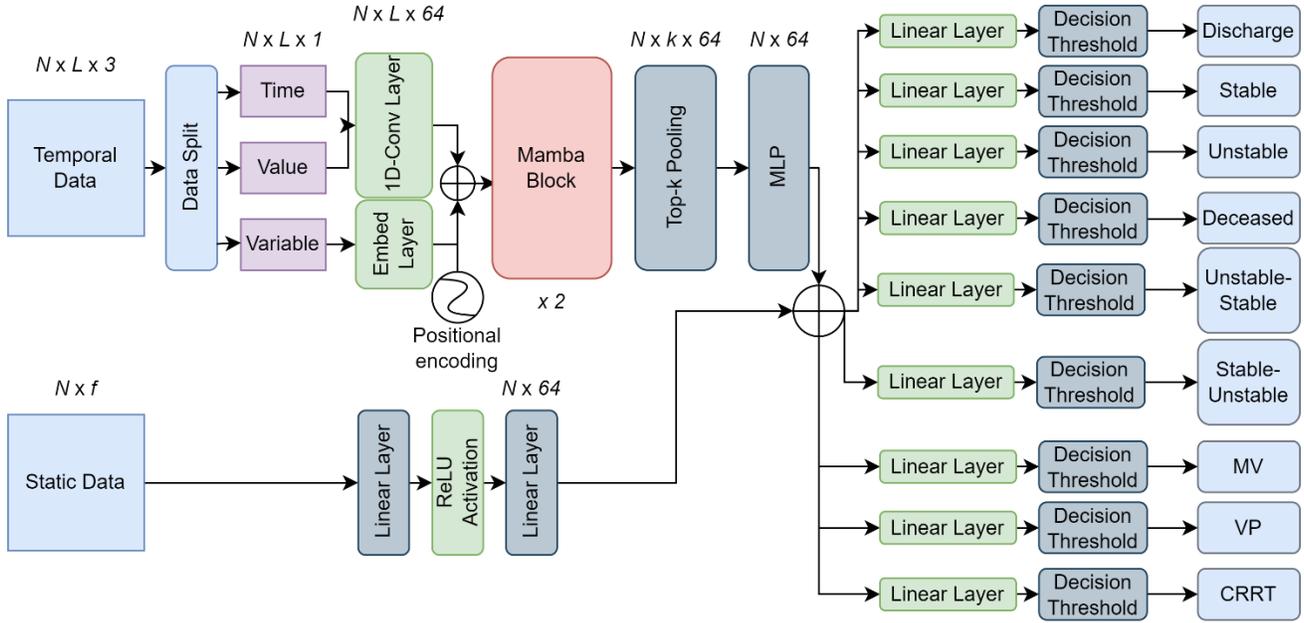

**Fig. 3 | APRICOT-M model architecture.** State space model adapted to a clinical setting. The temporal data of dimensions $N$ (number of samples) $\times$ $L$ (sequence length) $\times$ 3 (triplet recording for each event), is split into individual matrices for time, value, and variable code of dimensions $N \times L$ each. The time and value matrices go through a 1D-Convolutional Layer to create one embedding, while the variable code matrix goes through a look-up embedding layer to retrieve the embedding for the specific variable. Both embeddings are fused, and a positional encoding is added to conserve the sequence order information. The fused embedding is passed through two Mamba blocks, from which top-$k$ pooling is applied to retrieve the $k$-most important states. The vector is then passed through a multi-layer perceptron (MLP) to create an $N \times 64$ embedding fused with the static embedding resulting from passing the static data of dimensions $N \times f$ (static features) through two linear layers. Finally, the resulting vector is passed through four linear layers, which yield the probability for each acuity primary outcome (discharge, stable, unstable, decease), from which a decision threshold is applied to each outcome. The vector is also passed through two and three linear layers along with application of a decision threshold to determine state transition and need of life-sustaining therapies (CRRT, MV, VP) respectively. The acuity status for the next four hours is determined by applying a logic schema.

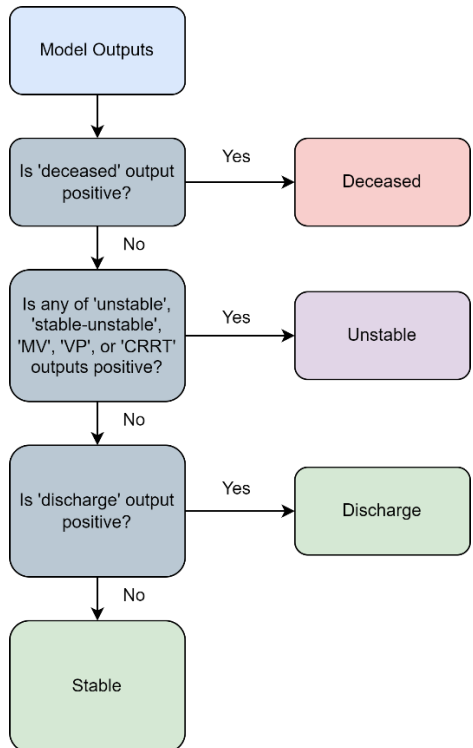

**Fig. 4 | Prediction logic.** The model outputs are processed to determine the acuity status through the use of a prediction logic. Deceased state is determined through the presence of a positive output for the 'deceased' prediction head and is given priority given it is the most severe state. Unstable state is determined if 'deceased' output is negative and through the presence of a positive label for any of 'unstable', 'stable-unstable', 'MV', 'VP', or 'CRRT' prediction heads. The presence of a positive label for any of the life-sustaining therapies (MV, VP, or CRRT) would indicate an unstable status. Discharge state is determined if both 'deceased' and 'unstable' outputs are negative and if the 'discharge' output is positive. Finally, stable state is determined if all other three states are negative.

2.5 Calibration and Subgroup Analysis

As the outcome rates in the three study cohorts were different, the APRICOT-M model was calibrated using an isotonic regression algorithm with three-fold cross-validation in order to obtain a risk probability more relevant to the true incidence probabilities. A random sampling of 10% for each validation set was used for model calibration. Calibration performance was evaluated by the Brier score and calibration curve. We evaluated model performance by comparing important demographic subgroups in all cohorts, including sex (male and female), age (young [18-60] and old [over 60]), and race (Black, White, and Others).

2.6 Model Interpretability

For the interpretation of the APRICOT-M model outputs, the integrated gradients method was used [28]. Integrated gradients attributions were calculated for each temporal variable in the embedding layer as well as for each static variable in the static embedding. This method allows computation of the importance of a feature as well as the time step, obtaining an overall score for each input feature to better understand risk predictors for acuity.

2.7 Statistical Analysis

To determine if the difference in performance between baseline models and APRICOT-M was statistically significant, all metric values between algorithms were compared using a Wilcoxon rank sum test. A 100-iteration bootstrap was performed to calculate the 95% confidence interval (CI) for each performance metric, and the median across the bootstrap was used to represent the overall value of each metric.

# 3. Results

3.1 Patient Characteristics

Patient characteristics for this study are presented in Table 1 in terms of ICU admissions. The external cohort had the highest median age (65.0) while the prospective cohort had the lowest (60.0). The percentage of female patients had no significant differences between the development, external, and temporal cohorts, but was lower in the prospective cohort (39.9%). The prospective cohort had the highest ICU length of stay (6.2 days) while the external cohort had the lowest (1.9 days). The incidence of mechanical ventilation and vasopressors was higher in the prospective set (41.6% and 50.5% respectively) compared to all other sets. The temporal cohort had the highest mortality rate (5.9%) while the prospective cohort had the lowest (2.7%). Comparisons of stable patients (*i.e.*, patients that did not experience instability in their ICU stay and survived) against patients that received MV, VP, and CRRT therapies and deceased patients, are provided in the Supplement for all cohorts (Supplementary Tables S2-S5).

Table 1. Baseline characteristics for ICU admissions of four study cohorts.

| Item | Development (*n* = 136,955) | External (*n* = 101,827) | Temporal (*n* = 15,940) | Prospective (*n* = 368) |
|---|---|---|---|---|
| **Basic information** | | | | |
| Number of patients | 104,787 | 75,668 | 12,927 | 215 |
| Number of hospital encounters | 125,630 | 93,718 | 15,940 | 325 |
| Age, years, median (IQR) | 64.0 (52.0-75.0) [b,c,d] | 65.0 (53.0-76.0) [a,c,d] | 62.0 (49.0-72.0) [a,b,d] | 60.0 (48.0-68.0) [a,b,c] |
| Female, n (%) | 62,013 (45.3%) [d] | 46,304 (45.5%) [d] | 7,158 (44.9%) | 147 (39.9%) [a,b] |
| BMI, kg/m$^2$, median (IQR) | 27.5 (23.7-32.5) [c,d] | 27.5 (26.0-29.1) [c,d] | 27.5 (23.3-32.3) [a,b,d] | 26.6 (23.0-30.5) [a,b,c] |
| ICU length of stay, days, median (IQR) | 2.0 (1.1-3.9) [b,c,d] | 1.9 (1.1-3.6) [a,c,d] | 3.1 (1.7-6.2) [a,b,d] | 6.2 (3.1-12.1) [a,b,c] |
| CCI, median (IQR) | 0.0 (0.0-1.0) [b,c,d] | 0.0 (0.0-0.0) [a,c,d] | 2.0 (1.0-4.0) [a,b] | 2.0 (1.0-4.0) [a,b] |
| **Race, n (%)** | | | | |
| Black | 19,378 (14.1%) [b,c] | 10,432 (10.2%) [a,c,d] | 3,046 (19.1%) [a,b,d] | 53 (14.4%) [b,c] |
| White | 104,213 (76.1%) [b,c] | 72,920 (71.6%) [a,c,d] | 11,881 (74.5%) [a,b] | 285 (77.4%) [b] |
| Other | 13,364 (9.8%) [b,c] | 18,475 (18.1%) [a,c,d] | 1,013 (6.4%) [a,b] | 30 (8.2%) [b] |
| **Comorbidities, n (%)** | | | | |
| Congestive heart failure | 12,121 (8.9%) [b,c,d] | 7,015 (6.9%) [a,c,d] | 4,806 (30.2%) [a,b,d] | 88 (23.9%) [a,b,c] |
| Chronic obstructive pulmonary disease | 13,347 (9.7%) [b,c,d] | 6,916 (6.8%) [a,c,d] | 4,761 (29.9%) [a,b,d] | 73 (19.8%) [a,b,c] |
| Renal disease | 9,865 (7.2%) [b,c,d] | 8,807 (8.6%) [a,c,d] | 3,577 (22.4%) [a,b,d] | 109 (29.6%) [a,b,c] |
| **Life-sustaining therapies, n (%)** | | | | |
| MV | 46,741 (34.1%) [b,c,d] | 35,241 (34.6%) [a,c,d] | 4,595 (28.8%) [a,b,d] | 153 (41.6%) [a,b,c] |
| VP | 25,806 (18.8%) [b,c,d] | 24,805 (24.4%) [a,c,d] | 4,586 (28.8%) [a,b,d] | 186 (50.5%) [a,b,c] |
| CRRT | 1,507 (1.1%) [b,c] | 1,684 (1.7%) [a,c] | 392 (2.5%) [a,b] | 4 (1.1%) |
| **Outcomes, n (%)** | | | | |
| Mortality | 7050 (5.1%) [b,c,d] | 5716 (5.6%) [a,d] | 942 (5.9%) [a,d] | 10 (2.7%) [a,b,c] |

Abbreviations: BMI: Body Mass Index; CCI: Charlson Comorbidity Index; CRRT: Continuous Renal Replacement Therapy; IQR: interquartile range; MV: Mechanical Ventilation; VP: Vasopressors.
P-values for continuous variables are based on pairwise Wilcoxon rank sum test. P-values for categorical variables are based on pairwise Pearson's chi-squared test for proportions.
[a] p-value < 0.05 compared to development cohort.
[b] p-value < 0.05 compared to external cohort.
[c] p-value < 0.05 compared to temporal cohort.
[d] p-value < 0.05 compared to prospective cohort.

## 3.2 State Transitions

The probability for each state transition across all study cohorts was calculated and shown in Fig. 5. For patients that are stable, there is a 97.10-97.83% probability of remaining stable, 1.26-1.66% of transitioning to an unstable state, 0.50-1.50% of transitioning to discharge, and 0.02-0.21% of transitioning to deceased. For patients that are unstable, there is a 92.12-93.94% probability of remaining unstable, 5.58-7.12% of transitioning to stable, 0.03-0.21% of transitioning to discharge, and 0.14-0.56% of transitioning to deceased.

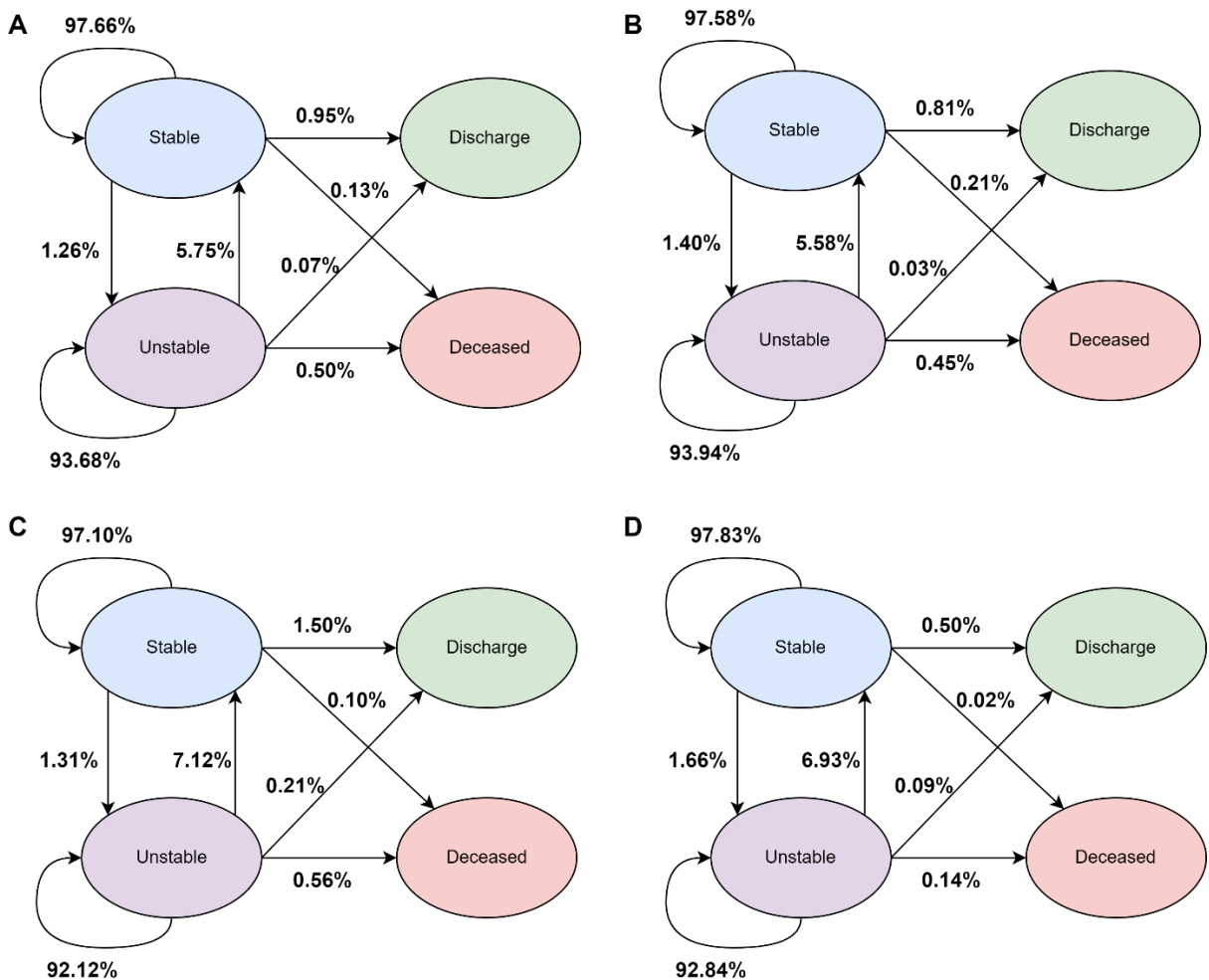

**Fig. 5 | State transition probability.** (A) Development cohort. (B) External cohort. (C) Temporal cohort. (D) Prospective cohort. The acuity state transition probability is shown for every 4-h window. The probability is calculated based on the current state of the patient (either stable or unstable) and the probability to transition to each of the acuity states. For discharge and deceased, the probabilities are not calculated since they are end outcomes.

3.2 Model Performance, Calibration, and Subgroup Analysis

The performance of APRICOT-M measured in AUROC is shown in Table 2 for prediction of the outcomes within the next 4 hours. Performance for discharge prediction was highest in the external cohort (0.88 CI: 0.87-0.88) and lowest in the temporal cohort (0.80 CI: 0.80-0.81). Prediction of deceased state had the highest performance in both temporal and prospective cohorts (0.98 CI: 0.97-0.98, and 0.98 CI: 0.96-1.00) and lowest performance in the development cohort (0.93 CI: 0.92-0.94), while prediction of stable-unstable transition had the highest performance in the development cohort (0.84 CI: 0.83-0.85) and lowest performance in the prospective cohort (0.72 CI: 0.68-0.75). Prediction for life-sustaining therapies showed highest performance on temporal cohort for MV (0.87 CI: 0.87-0.88), development cohort for VP (0.83 CI: 0.82-0.84), and temporal cohort for CRRT (0.96 CI: 0.95-0.96). All other metrics and comparisons to baseline models are shown in Supplementary Table S6. Calibration curves and brier scores for each cohort are shown in Supplementary Figures S1-3. Subgroup analysis based on age, gender, and race groups is shown in Supplementary Table S7. This analysis could not be performed in the prospective cohort due to absence of positive labels for some groups given the small size of the cohort.

**Table 2. APRICOT-M performance for 4-h predictions in terms of AUROC for four study cohorts.**

| Outcomes | Development | External | Temporal | Prospective |
|---|---|---|---|---|
| Primary | | | | |
| Discharge | 0.87 (0.86-0.87) [b,c] | 0.88 (0.87-0.88) [a,c,d] | 0.80 (0.80-0.81) [a,b,d] | 0.84 (0.80-0.87) [b,c] |
| Stable | 0.94 (0.94-0.94) [b,c,d] | 0.95 (0.94-0.95) [a] | 0.95 (0.95-0.95) | 0.95 (0.95-0.96) [a] |
| Unstable | 0.95 (0.95-0.95) [c,d] | 0.95 (0.95-0.95) [c,d] | 0.97 (0.97-0.97) | 0.96 (0.96-0.96) [c] |
| Deceased | 0.93 (0.92-0.94) [b,c,d] | 0.95 (0.94-0.95) [a,c,d] | 0.98 (0.97-0.98) [a,b] | 0.98 (0.96-1.00) [a,b] |
| Transition | | | | |
| Unstable-Stable | 0.74 (0.73-0.74) [b,c,d] | 0.79 (0.79-0.79) [a,c,d] | 0.91 (0.91-0.92) [a,b,d] | 0.83 (0.81-0.87) [a,b,c] |
| Stable-Unstable | 0.84 (0.83-0.85) [b,c,d] | 0.82 (0.81-0.82) [a,c,d] | 0.78 (0.77-0.78) [a,b,d] | 0.72 (0.68-0.75) [a,b,c] |
| Life-sustaining therapy | | | | |
| MV | 0.85 (0.84-0.86) [b,c,d] | 0.82 (0.82-0.83) [a,c,d] | 0.87 (0.87-0.88) [a,b,d] | 0.72 (0.67-0.76) [a,b,c] |
| VP | 0.83 (0.82-0.84) [c,d] | 0.82 (0.81-0.82) [c,d] | 0.74 (0.73-0.75) [a,b] | 0.71 (0.66-0.74) [a,b] |
| CRRT | 0.89 (0.86-0.91) [c] | 0.88 (0.87-0.89) [c] | 0.96 (0.95-0.96) [a,b] | 0.91 (0.86-0.98) |

Abbreviations: CRRT: Continuous Renal Replacement Therapy; MV: Mechanical Ventilation; VP: Vasopressors. Performance is shown as the median AUROC across 100-iteration bootstrap with 95% Confidence Intervals in parenthesis.
P-values are based on pairwise Wilcoxon rank sum tests.
[a] p-value < 0.05 compared to internal cohort.
[b] p-value < 0.05 compared to external cohort.
[c] p-value < 0.05 compared to temporal cohort.
[d] p-value < 0.05 compared to prospective cohort.

3.3 False Positive Analysis

To investigate the false positives in deceased and transition to instability predictions, a false positive analysis was conducted. For this analysis, the number of false positives occurring more than four hours before a true positive (*i.e.*, false positives happening on earlier 4-h windows) or after the true positive (*i.e.*, false positives happening in 4-h windows after a true positive) are quantified. For mortality, out of all false positives, 21.4%, 30.0%, 35.0%, and 27.1% occurred before a true positive on the development, external, temporal, and prospective sets respectively. The distribution of deceased false positives in the hours before death were quantified and sensitivity and PPV were adjusted to account for false positives occurring more than 4 hours before a positive (Fig. 6). This resulted in significant improvements in sensitivity (86% to 98% on development set, 85% to 97% on external set, 92% to 99% on temporal set, and 92 to 100% on prospective set) and PPV (1% to 10% on development set, 2% to 14% on external set, 3% to 16% on temporal set, and 1 to 11% on prospective set). For transition to instability, 38.0%, 43.9%, 56.9%, and 83.9% of false positives occurred before or after a true positive on the development, external, temporal, and prospective sets respectively. Similarly, as in mortality, distribution of false positives was quantified with respect to the closest positive label. Sensitivity and PPV were also adjusted to account for false positives occurring more than 4 hours before a true positive (Fig. 7). This also resulted in significant improvements in sensitivity (75% to 92% on development set, 65% to 85% on external set, 65% to 91% on temporal set, and 83% to 99% on prospective set) and PPV (2% to 11% on development set, 3% to 17% on external set, 2% to 21% on temporal set, and 2% to 45% on prospective set).

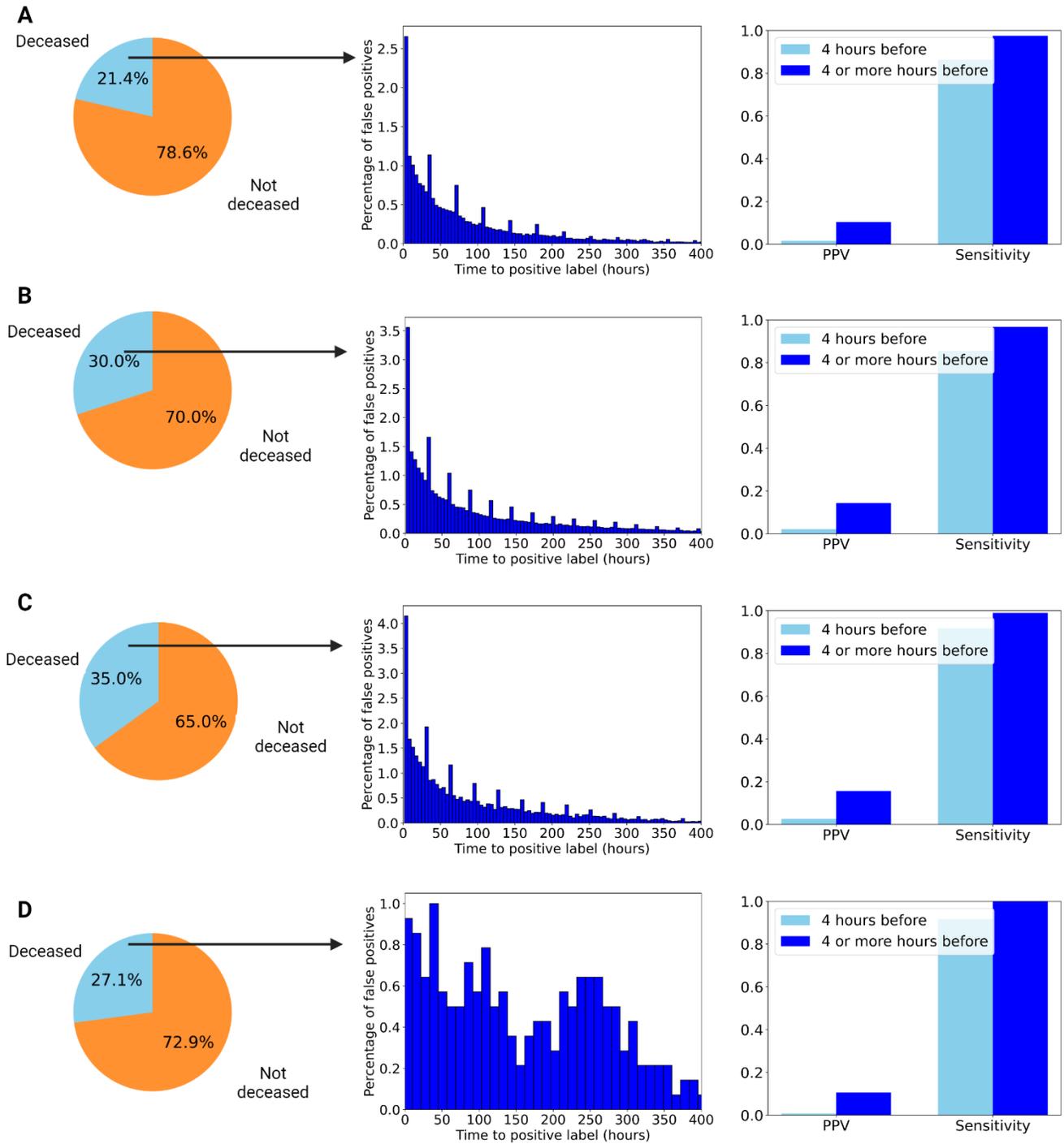

**Fig. 6 | False positive analysis for mortality prediction.** (A) Development set. (B) External set. (C) Temporal set. (D) Prospective set. The proportion of all false positives is quantified for deceased and not deceased patients. The distribution of false positives on deceased patients is quantified according to the number of hours before the time of death where the false positives occurred. Furthermore, sensitivity and PPV are adjusted by counting predictions with 4 or more hours lead as true positives.

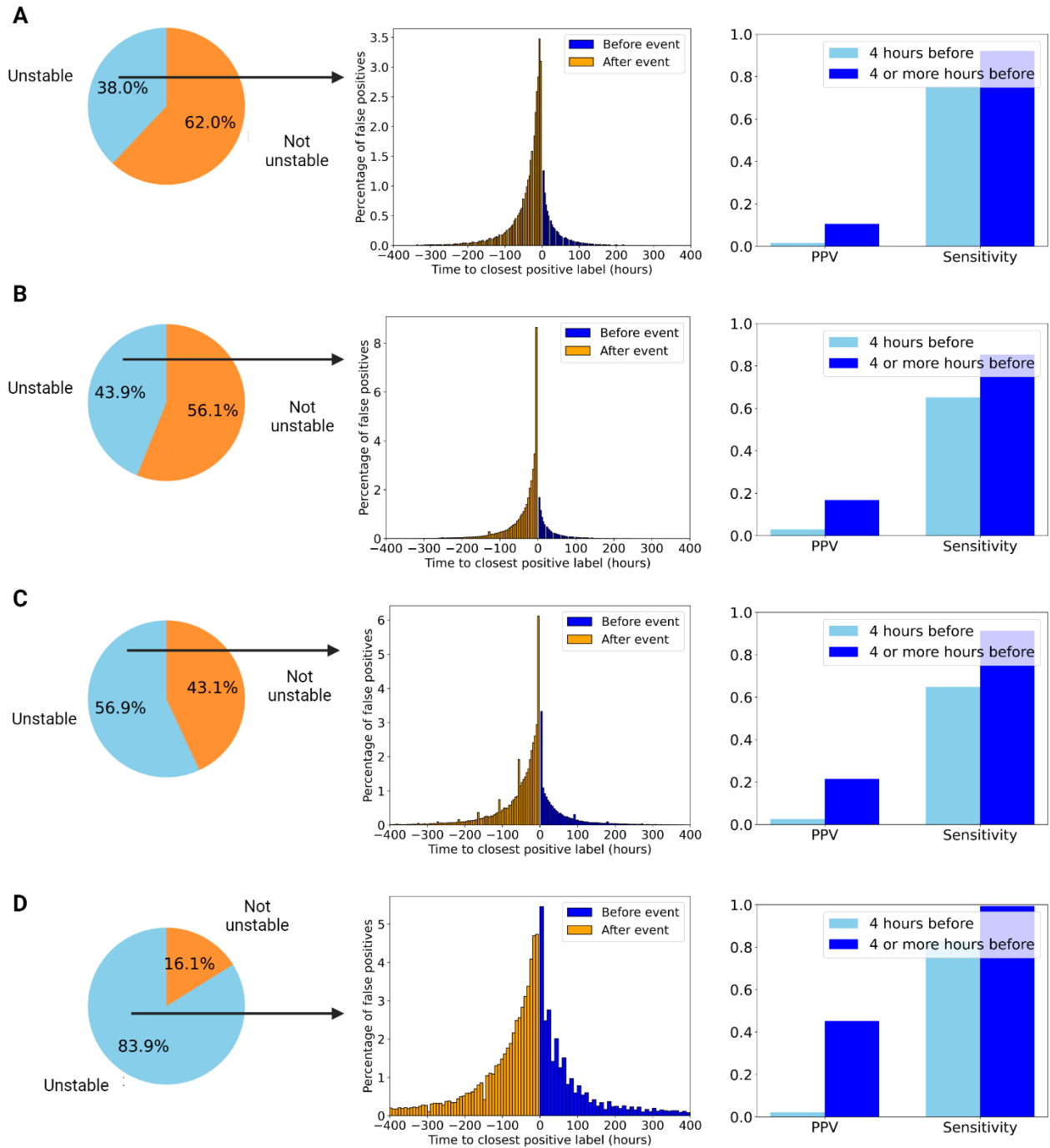

**Fig. 7 | False positive analysis for transition to instability.** (A) Development set. (B) External set. (C) Temporal set. (D) Prospective set. The proportion of false positives occurring in unstable and not unstable patients is quantified. False positives occurring in patients that transitioned to an unstable state were plotted to determine the proportion that occurred before and after a true positive. Furthermore, sensitivity and PPV are adjusted by counting predictions with 4 or more hours lead as true positives.

3.4 Predictions distribution

To further evaluate the performance of APRICOT-M on predicting acuity status, a confusion matrix was generated for the four primary outcomes: discharge, stable, unstable, and deceased. The proportions of predicted acuity status (Fig. 8) show that across all cohorts, the sensitivity for predicting discharge and mortality ranges between 36-76% and 86-92% respectively. For prediction of stable and unstable states, the sensitivity range is 11-35% and 66-90% respectively, with 14-34% of stability cases being predicted as discharge and 9-28% of instability cases being predicted as deceased. Furthermore, the proportions of each acuity state across each day in the ICU are shown in Fig. 9 compared to the predictions by APRICOT-M. This figure shows higher number of discharge, unstable, and deceased states predictions compared to the ground truth number of these states.

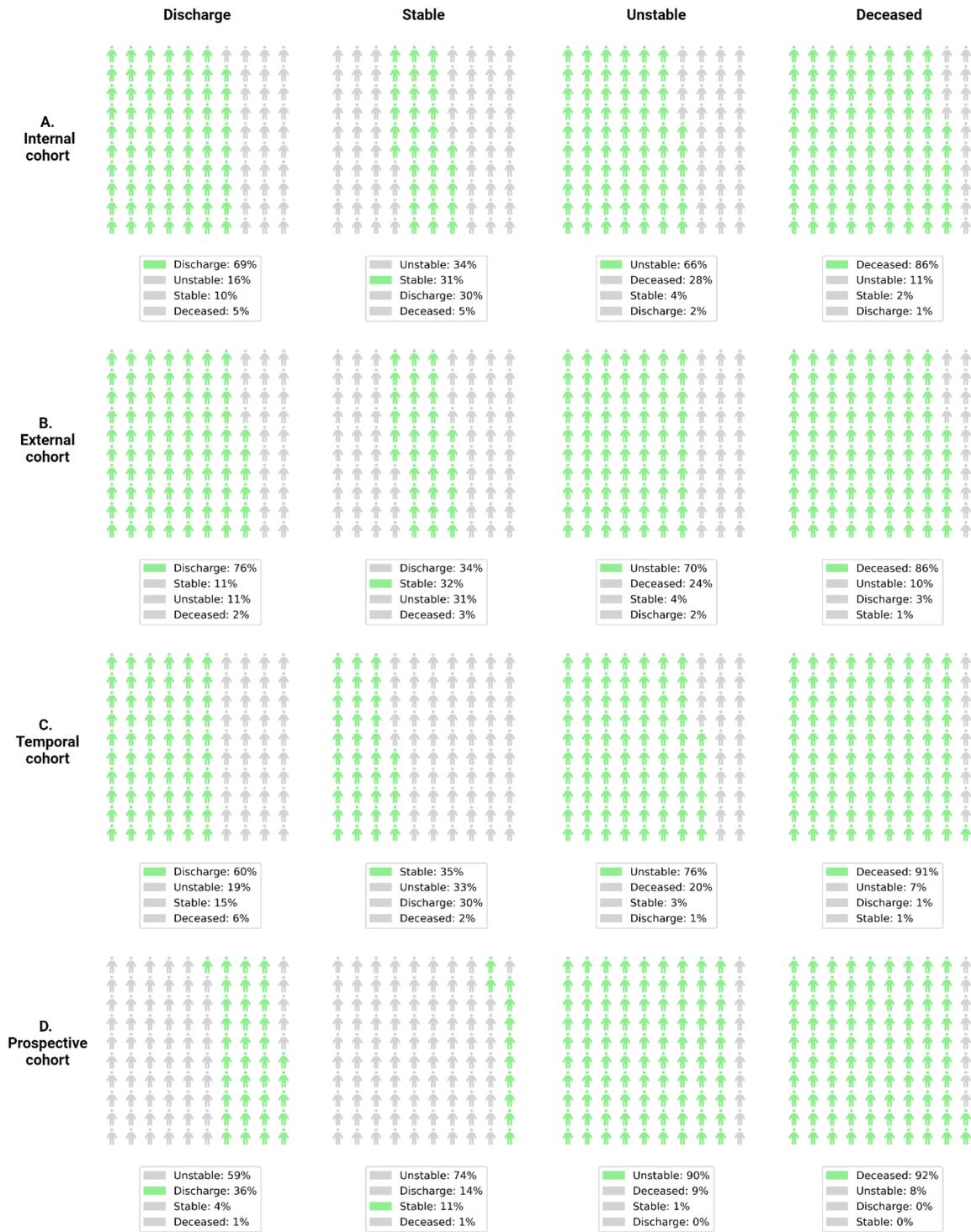

**Fig. 8 | Proportions of acuity status predictions.** (A) Development cohort. (B) External cohort. (C) Temporal cohort. (D) Prospective cohort. Each plot shows the proportion of predicted acuity states for each class according to the column: discharge, stable, unstable, and deceased. Green represents a correct prediction (*i.e.*, true positives) for those who had the specific outcome, while gray represents an incorrect prediction (*i.e.*, any state other than the correct one was predicted).

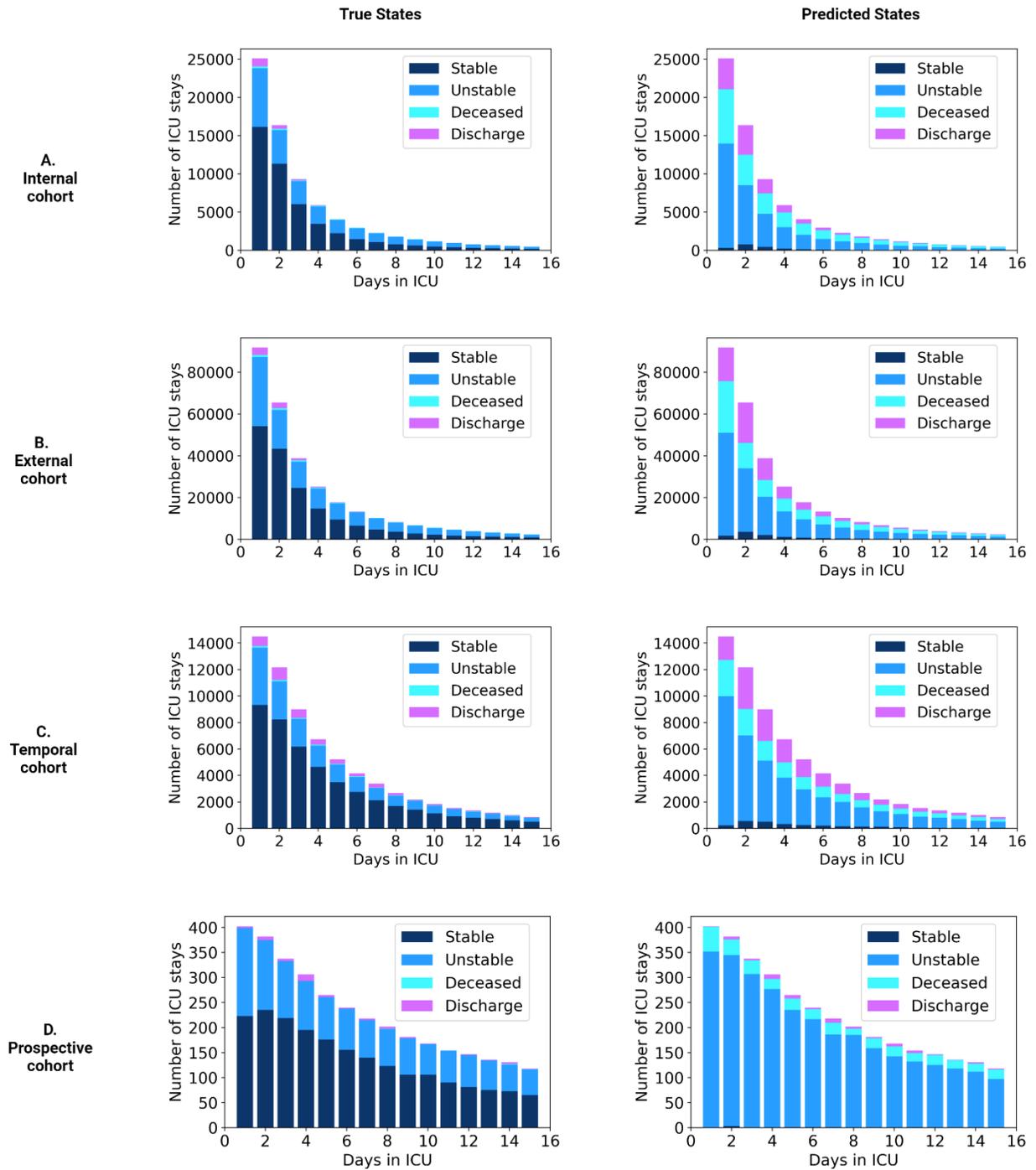

**Fig. 9 | Distribution of true and predicted acuity states across days in ICU.** (A) Development cohort. (B) External cohort. (C) Temporal cohort. (D) Prospective cohort. The left side shows the true distribution of acuity states in the first 15 days of ICU admission for each cohort, while the right side shows the distribution of predicted acuity states in comparison.

## 3.5 Feature importance

The relevance of each feature for prediction of each outcome was determined through the use of integrated gradients. The top 15 features for primary outcomes (i.e., discharge, deceased, stable, and unstable) prediction across all cohorts are presented in Fig. 10. Among the top features common in all cohorts are age, Glasgow coma scale (GCS) score, oxygen (O2) flow, vasopressors (*i.e.*, epinephrine, norepinephrine, phenylephrine, vasopressin, dopamine), and mechanical ventilator settings (*i.e.*, positive end-expiratory pressure (PEEP), tidal volume). A complete feature ranking for all prediction tasks across all cohorts is provided in the Supplement (Supplementary Figures S4-S6).

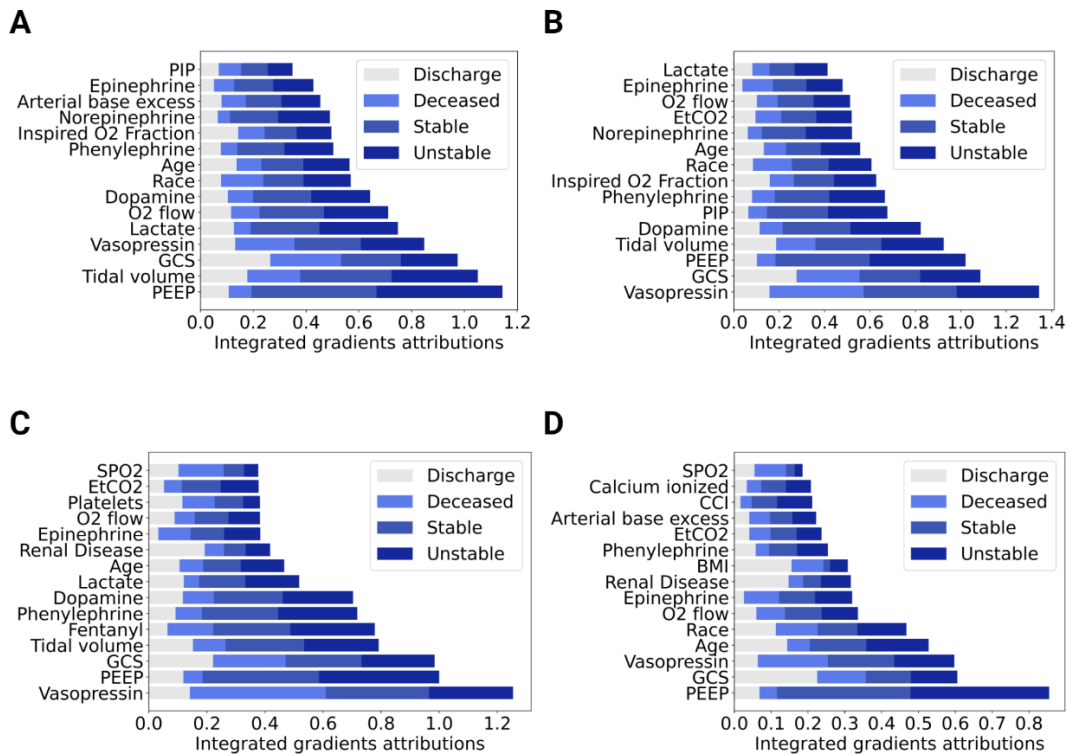

**Fig. 10 | Feature importance.** (A) Development cohort. (B) External cohort. (C) Temporal cohort. (D) Prospective cohort. The top 15 features for primary outcome prediction. Features are ordered according to the highest sum of integrated gradients attributions across the four primary outcomes: discharge, deceased, stable, and unstable. The feature importance for each primary outcome is represented with a different color.

## 3.6 Examples of Real-Time Evaluation

Two example patients were extracted from the external cohort to provide a demonstration of APRICOT-M real-time capabilities. The first example was a patient who passed away at the end of their ICU stay. The patient transitioned from stable to unstable at around 136 hours after ICU admission when they went into mechanical ventilation (MV). About 8 hours later, the patient entered vasopressor (VP) therapy. The patient remained unstable until death at around 280 hours after ICU admission. The APRICOT-M model detected increased mortality risk at around 200 hours (Fig. 11A), which kept increasing until the time of death. The vitals chart on Fig. 10B shows the integrated gradient attribution to each time step for each vital sign along the patient's trajectory in the ICU. The second example was a patient who was discharged to their home at the end of their ICU stay. This patient transitioned from stable to unstable at around 20 hours after ICU admission when they went into MV and VP therapies simultaneously. The patient remained unstable until about 156 hours after admission where they transitioned back to stable. After 24 hours of this transition, the patient was discharged. The APRICOT-M model detected decreased instability risk at around 140 hours (Fig. 12A), which is about 16 hours before the patient transitioned to a stable condition. Also, mortality risk remained low at all times during the ICU stay.

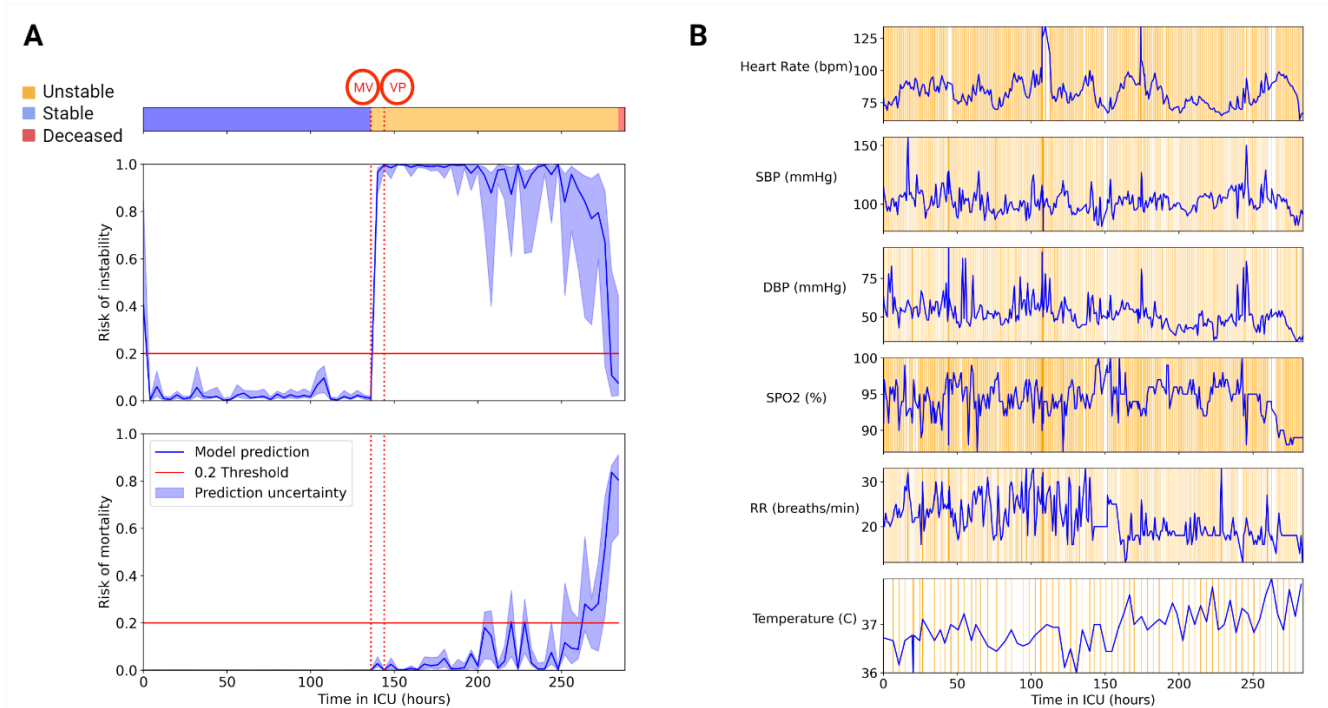

**Fig. 11 | APRICOT-M real-time monitoring on deceased patient.** Example of real-time monitoring of a deceased patient in the external cohort. (A) Acuity trajectory of the patient, with onset of vasopressor (VP) and mechanical ventilation (MV) therapies, along with instability and mortality risk predicted by APRICOT-M throughout ICU admission. (B) Vitals trajectory of patient including Heart Rate (HR), Diastolic Blood Pressure (DBP), Systolic Blood Pressure (SBP), Respiratory Rate (RR), Oxygen Saturation (SPO2), and Body Temperature. Orange shading represents integrated gradients attributions at each time step, with higher color intensity representing a higher attribution.

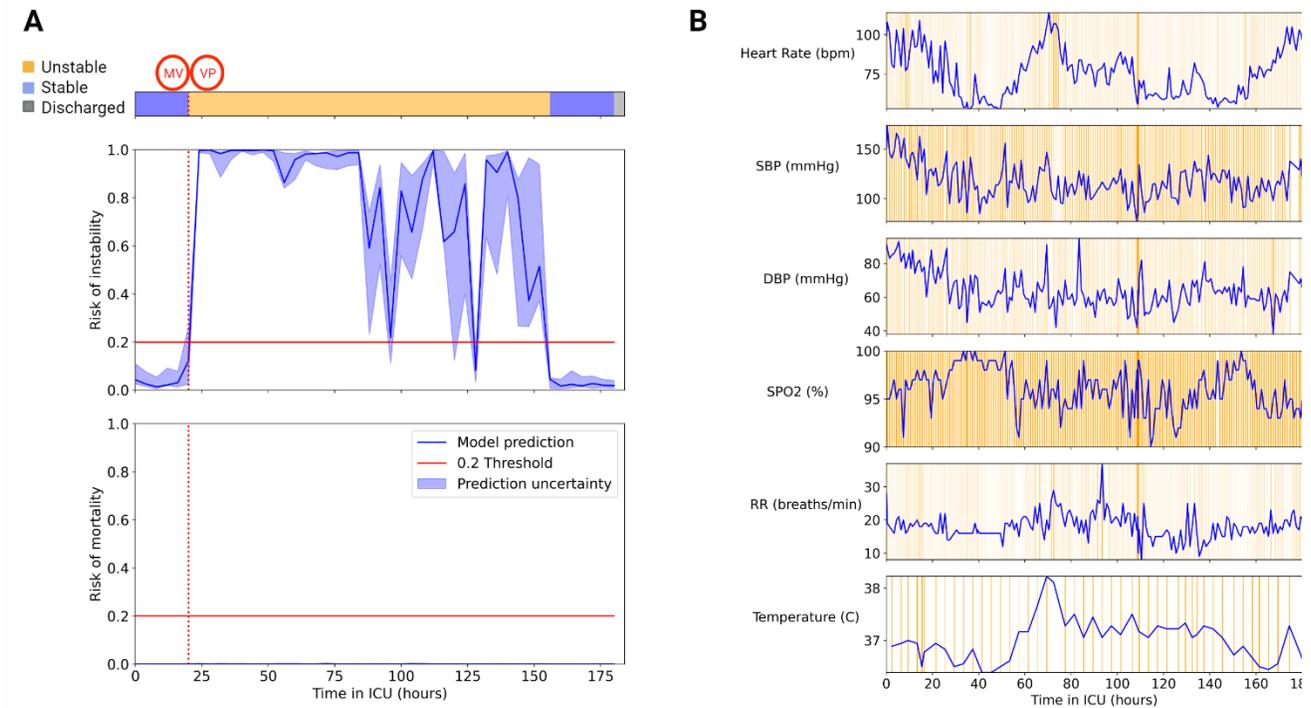

**Fig. 12 | APRICOT-M real-time monitoring on discharged patient.** Example of real-time monitoring of a discharged patient in the external cohort. (A) Acuity trajectory of the patient, with onset of vasopressor (VP) and mechanical ventilation (MV) therapies, along with instability and mortality risk predicted by APRICOT-M throughout ICU admission. (B) Vitals trajectory of patient including Heart Rate (HR), Diastolic Blood Pressure (DBP), Systolic Blood Pressure (SBP), Respiratory Rate (RR), Oxygen Saturation (SPO2), and Body Temperature. Orange shading represents integrated gradients attributions at each time step, with higher color intensity representing a higher attribution.

## 4. Discussion

The results presented in this study show the development and validation of APRICOT-M as a tool for ICU patient acuity monitoring. To the best of our knowledge, this is the first study to predict acuity in the ICU based on life-sustaining therapies (mechanical ventilation, vasopressors, continuous renal replacement therapy, and blood transfusion). Furthermore, discharge and mortality outcomes are included to obtain a full spectrum of acuity levels. Previous studies to predict patient acuity in the ICU have focused on mortality prediction as a proxy of acuity [9], [10]. Comparing mortality prediction in this study, performance on all sets (development AUROC 0.92-0.94, external AUROC 0.94-0.95, temporal AUROC 0.97-0.98, and prospective AUROC 0.96-1.00) exceeds DeepSOFA (AUROC 0.90-0.91) [9] and is comparable to the Transformer acuity estimation model (AUROC 0.98) [10]. Furthermore, this study provides the advantage of validating mortality prediction performance on patients from multiple hospitals, which proves the generalizability of APRICOT-M. Also, instability transition prediction had similar performance on the external set (AUROC 0.81-0.82) to one study which performs acuity prediction (AUROC 0.81-0.82) [34], although this study used nurse intensity care as a proxy of acuity.

The model presented here provides the advantage of predicting the type of life-sustaining therapy that the patient will need in addition to predicting overall acuity. Although no studies predict the need for CRRT to the best of our knowledge, some studies have predicted the use of vasopressors and mechanical ventilation. For vasopressor prediction, the performance in development and external sets (AUROC 0.82-0.84 and AUROC 0.81-0.82) was higher than seen in the literature (AUROC 0.80-0.81) [33], while the temporal and prospective sets showed lower performance (AUROC 0.73-0.75 and AUROC 0.66-0.74 respectively) which could be due to higher proportions of mechanical ventilation on patients receiving vasopressors in the development and external sets (68.6% and 73.2% respectively) compared to temporal and prospective sets (61.4% and 65.6% respectively) (Supplementary Tables S2-S5). For mechanical ventilation prediction, performance reported in the literature is varied, with some studies reporting as low as AUROC 0.68 [31] and others reporting as high as AUROC 0.97 [32]. It is important to note that these studies perform mechanical ventilation prediction on COVID-19 patients which could be a potential bias. Furthermore, the models are developed and validated on data from only one hospital. The diverse settings and reporting guidelines of each hospital can make it difficult to fully capture ventilation events and extract with high precision the start and end times of these events. The performance shown in this study falls within the range of what is seen in the literature (development AUROC 0.84-0.86, external AUROC 0.82-0.83, temporal AUROC 0.87-0.88, and prospective AUROC 0.67-0.76) and is tested in data from multiple hospitals. The lower performance in ventilation prediction in the prospective cohort could be explained by the higher length of stay for ICU admissions in this cohort. A longer ICU admission could lead to more false positives at the 4-h window level, which reduces AUROC.

The high count of false positives for transition to unstable and deceased predictions can be a concern in model reliability. However, a false positive analysis revealed that between 38-84% of all false positives for transition to unstable and 21-35% for deceased across the four study cohorts occurred in an ICU admission where the patient eventually presented the outcome. The plots in Fig. 6 and Fig. 7 show that many false positives occur earlier than the 4-hour window. Adjusting for false positives happening earlier than the 4-hour prediction window significantly increased both sensitivity and PPV for transition to unstable and deceased predictions. It is also important to note that timely intervention from the caregivers in the ICU would prevent the patient from transitioning to a higher acuity state. This could partially explain the inflated false positive rate of the model.

The distribution of predictions shown in Fig. 8 demonstrates that APRICOT-M correctly identifies the primary outcomes with high sensitivity for deceased (86-92%) and unstable (66-90%). Although sensitivity for stability (11-35%) and discharge (36-76%) is lower, many false negatives are predicted as unstable for both stable (31-74%) and discharged (11-59%) patients. These numbers reflect the high count of false positives alarms for unstable state in these patients. This is also reflected in Fig. 9, where a significant proportion of the model's output is unstable predicted labels. As previously shown in the false positive analysis, in many cases the model alerts several hours before a patient transition to unstable state, which inflates the number of false positives.

The two example cases presented in Fig. 10 and Fig. 11, show the potential use of APRICOT-M in real-time in the ICU. For the deceased patient example, the model recognized increased mortality risk several days before the patient passed away. On the other hand, the model recognized decreased instability risk in a similar way for the discharged patient example. These two examples show the capability of APRICOT-M to provide early warnings for more timely interventions.

The use of integrated gradients revealed the most important features for acuity prediction. The importance of age for this type of prediction is consistent from the knowledge that older patients are at a higher risk in the ICU [35]. The GCS and CAM scores, which are used to determine coma and delirium in patients, are also correlated to patient acuity [36], [37]. Both the presence of vasopressors (*i.e.*, epinephrine, norepinephrine, phenylephrine, vasopressin, dopamine) and mechanical ventilator settings (*i.e.*, PEEP, tidal volume) were relevant to acuity prediction. The fact that stability and instability are defined by presence of these therapies can account for the high relevance of these features.

The prospective evaluation of APRICOT-M on real-time ICU patient data provides a validation method of this model in the real world. The performance seen in this cohort was comparable to all three retrospective cohorts, which shows APRICOT-M can be readily deployed and employed in real-time to assist clinicians in decision-making. However, deployment of the model in real-time and in larger prospective cohorts are necessary to fully validate the real-time performance of the model.

Some of the limitations of this study include the challenge of labeling outcomes in different hospital settings. Given the high number of hospitals included in this study, labeling of procedures such as mechanical ventilation and continuous renal replacement therapy can be inconsistent. Different hospitals have different reporting guidelines, meaning some of these procedures might not be recorded, and even if they are, start and end times could be inaccurate. This could limit the performance of the model due to inconsistency in the labels. The small number of features can limit model performance as well. Large models can usually benefit from a large number of features. The present study used a limited subset of features present on all three databases (UFH, MIMIC, and eICU). Since these features were manually mapped between datasets, some features could have been missed. Future studies will employ the Observational Medical Outcomes Partnership (OMOP) Common Data Model to map features more effectively and accurately between datasets.

In summary, we have developed and validated both retrospectively and prospectively APRICOT-M, a state space model to predict patient acuity in the ICU. This tool allows for real-time acuity monitoring of a patient and can provide helpful information to clinicians to make timely interventions. Furthermore, the model can suggest life sustaining therapies that the patient might need in the next hours.

## 5. Acknowledgement

A.B, P.R., and T.O.B. were supported by NIH/NINDS R01 NS120924, NIH/NIBIB R01 EB029699. P.R. was also supported by NSF CAREER 1750192.

# APRICOT-Mamba: <u>A</u>cuity <u>Pr</u>ed<u>i</u>c<u>t</u>i<u>o</u>n i<u>n</u> I<u>nt</u>ensive <u>C</u>are Unit (ICU): Development and Validation of a Stability, Transitions, and Life-Sustaining Therapies Prediction Model

Miguel Contreras [1,6], Brandon Silva[2,6], Benjamin Shickel[3,6], Tezcan Ozrazgat-Baslanti[3,6], Yuanfang Ren[3,6], Ziyuan Guan[3,6], Jeremy Balch[4,6], Jiaqing Zhang[2,6], Sabyasachi Bandyopadhyay[5], Kia Khezeli[1,6], Azra Bihorac[3,6], Parisa Rashidi[1,6]**

[1]Department of Biomedical Engineering, University of Florida, Gainesville, FL, USA
[2]Department of Electrical and Computer Engineering, University of Florida, Gainesville, FL, USA
[3]Department of Medicine, University of Florida, Gainesville, FL, USA
[4]Department of Surgery, University of Florida, Gainesville, FL, USA
[5]Department of Medicine, Stanford University, Stanford, CA, USA
[6]Intelligent Clinical Care Center (IC3), University of Florida, Gainesville, FL, USA

**SUPPLEMENTARY TABLES**

**Supplementary Table S1. Feature names.**

| Category | Type | Name |
|---|---|---|
| Basic information | Static | Age, Body mass index (BMI), Gender, Race |
| Comorbidities | Static | Charlson comorbidity index (CCI), Acquired immunodeficiency syndrome, Cancer, Cerebrovascular disease, Congestive heart failure, Chronic obstructive pulmonary disease, Dementia, Diabetes without complications, Diabetes with complications, Myocardial infarction, Metastatic carcinoma, Mild liver disease, Moderate severe liver disease, Paraplegia hemiplegia, Peptic ulcer disease, Peripheral vascular disease, Renal disease, Rheumatologic disease |
| Assessment scores | Temporal | Confusion Assessment Method (CAM), Glasgow Coma Scale (GCS), Richmond agitation sedation scale (RASS) |
| Laboratory values | Temporal | Alanine transaminase (ALT), Aspartate aminotransferase (AST), Basophils, Eosinophils, Lymphocytes, Monocytes, Albumin, Anion gap, Arterial base excess, Arterial CO2 pressure, Arterial O2 Saturation, Arterial O2 pressure, Brain natriuretic peptide (BNP), C reactive protein (CRP), Calcium non-ionized, Chloride, Creatinine, Bilirubin direct, Glucose, Hematocrit, Hemoglobin, Prothrombin time (INR), Calcium ionized, Lactate, Arterial pH, Platelets, Potassium, Sodium, Specific gravity urine, Bilirubin total, Troponin-T, White Blood Cell (WBC) |
| Medications | Temporal | Heparin sodium, Propofol, Phenylephrine, Folic acid, Norepinephrine, Amiodarone, Fentanyl, Dexmedetomidine, Digoxin, Vasopressin, Epinephrine, Dopamine |
| Vital signs | Temporal | End-tidal carbon dioxide (EtCO2), Heart rate (HR), Inspired O2 Fraction, Diastolic blood pressure (DBP), Systolic blood pressure (SBP), Oxygen (O2) flow, Oxygen saturation (SPO2), Peak inspiratory pressure (PIP), Respiratory rate (RR), Body temperature (T), Tidal volume, Positive end-expiratory pressure (PEEP) |

**Supplementary Table S2. Development cohort baseline characteristics.**

|  |  | Overall (*n* = 136,955) | Stable (*n* = 81,355) | MV (*n* = 46,741) | VP (*n* = 25,806) | CRRT (*n* = 1,507) | Deceased (*n* = 7,050) |
|---|---|---|---|---|---|---|---|
|  | Number of patients | 104,787 | 66,967 | 40,413 | 23,487 | 1,447 | 7,050 |
|  | Number of hospital encounters | 125,630 | 76,818 | 44,405 | 25,003 | 1,464 | 7,050 |
| **Basic information** | | | | | | | |
|  | Age, years | 64.0 (52.0-75.0) | 64.0 (51.0-75.0) | 64.0 (53.0-74.0) | 65.0 (55.0-75.0)* | 62.0 (52.0-71.0)* | 68.0 (57.0-78.0)* |
|  | BMI, kg/m^2 | 27.5 (23.7-32.5) | 27.5 (23.7-32.1) | 27.7 (23.8-33.2)* | 27.5 (23.6-32.8) | 28.9 (24.6-35.2)* | 27.3 (23.0-32.3)* |
|  | Female | 62,013 (45.3%) | 37,871 (46.6%) | 19,922 (42.6%)* | 11,004 (42.6%)* | 619 (41.1%)* | 3,134 (44.5%)* |
|  | ICU length of stay, days | 2.0 (1.1-3.9) | 1.6 (1.0-2.7) | 3.5 (1.9-7.0)* | 4.0 (2.1-7.8)* | 7.8 (3.9-13.9)* | 3.1 (1.4-6.7)* |
| **Race** | | | | | | | |
|  | Black | 19,378 (14.1%) | 11,689 (14.4%) | 6,689 (14.3%) | 2,955 (11.5%)* | 314 (20.8%)* | 996 (14.1%) |
|  | Other | 13,364 (9.8%) | 7,622 (9.4%) | 4,910 (10.5%)* | 2,737 (10.6%)* | 155 (10.3%) | 712 (10.1%)* |
|  | White | 104,213 (76.1%) | 62,044 (76.3%) | 35,142 (75.2%)* | 20,114 (77.9%)* | 1,038 (68.9%)* | 5,342 (75.8%) |
| **Comorbidity index** | | | | | | | |
|  | CCI | 0.0 (0.0-1.0) | 0.0 (0.0-1.0) | 0.0 (0.0-1.0) | 0.0 (0.0-2.0) | 0.0 (0.0-3.0) | 0.0 (0.0-2.0) |
| **Comorbidities** | | | | | | | |
|  | Acquired immunodeficiency syndrome | 26 (0.0%) | 11 (0.0%) | 13 (0.0%) | 4 (0.0%) | 0 (0.0%) | 7 (0.1%)* |
|  | Cancer | 6,654 (4.9%) | 4,147 (5.1%) | 1,859 (4.0%)* | 1,378 (5.3%) | 101 (6.7%)* | 519 (7.4%)* |
|  | Cerebrovascular disease | 9,057 (6.6%) | 5,623 (6.9%) | 2,806 (6.0%)* | 1,796 (7.0%) | 61 (4.0%)* | 779 (11.0%)* |
|  | Chronic obstructive pulmonary disease | 13,347 (9.7%) | 7,595 (9.3%) | 4,764 (10.2%)* | 2,978 (11.5%)* | 187 (12.4%)* | 747 (10.6%)* |
|  | Congestive heart failure | 12,121 (8.9%) | 6,588 (8.1%) | 4,412 (9.4%)* | 3,322 (12.9%)* | 309 (20.5%)* | 926 (13.1%)* |
|  | Dementia | 1,145 (0.8%) | 696 (0.9%) | 319 (0.7%)* | 272 (1.1%)* | 16 (1.1%) | 86 (1.2%)* |
|  | Diabetes with complications | 2,920 (2.1%) | 1,830 (2.2%) | 767 (1.6%)* | 795 (3.1%)* | 122 (8.1%)* | 193 (2.7%)* |
|  | Diabetes without complications | 7,690 (5.6%) | 4,712 (5.8%) | 2,258 (4.8%)* | 1,924 (7.5%)* | 158 (10.5%)* | 515 (7.3%)* |

| | | | | | | |
|---|---|---|---|---|---|---|
| Metastatic carcinoma | 2,127 (1.6%) | 1,324 (1.6%) | 544 (1.2%)* | 483 (1.9%)* | 29 (1.9%) | 210 (3.0%)* |
| Mild liver disease | 3,221 (2.4%) | 1,681 (2.1%) | 1,203 (2.6%)* | 1,044 (4.0%)* | 190 (12.6%)* | 397 (5.6%)* |
| Moderate severe liver disease | 1,100 (0.8%) | 502 (0.6%) | 452 (1.0%)* | 439 (1.7%)* | 109 (7.2%)* | 201 (2.9%)* |
| Myocardial infarction | 7,155 (5.2%) | 4,490 (5.5%) | 2,080 (4.5%)* | 1,738 (6.7%)* | 125 (8.3%)* | 409 (5.8%) |
| Paraplegia hemiplegia | 2,165 (1.6%) | 1,137 (1.4%) | 753 (1.6%)* | 724 (2.8%)* | 13 (0.9%) | 246 (3.5%)* |
| Peptic ulcer disease | 797 (0.6%) | 430 (0.5%) | 260 (0.6%) | 250 (1.0%)* | 23 (1.5%)* | 52 (0.7%)* |
| Peripheral vascular disease | 5,991 (4.4%) | 3,190 (3.9%) | 2,119 (4.5%)* | 2,027 (7.9%)* | 153 (10.2%)* | 373 (5.3%)* |
| Renal disease | 9,865 (7.2%) | 5,593 (6.9%) | 3,280 (7.0%) | 2,636 (10.2%)* | 447 (29.7%)* | 796 (11.3%)* |
| Rheumatologic disease | 958 (0.7%) | 570 (0.7%) | 291 (0.6%) | 245 (0.9%)* | 26 (1.7%)* | 78 (1.1%)* |
| **Vital signs** | | | | | | |
| Body temperature | 37.0 (36.6-37.5) | 36.8 (36.6-37.2) | 37.1 (36.6-37.6)* | 37.1 (36.6-37.6)* | 36.9 (36.4-37.4)* | 37.0 (36.4-37.6)* |
| DBP | 63.0 (55.0-74.0) | 66.0 (57.0-76.0) | 62.0 (54.0-72.0)* | 60.0 (52.0-70.0)* | 57.0 (49.0-66.0)* | 58.0 (50.0-68.0)* |
| EtCO2 | 33.0 (30.0-38.0) | 33.0 (30.0-37.0) | 33.0 (30.0-38.0) | 33.0 (30.0-37.0) | 32.0 (27.0-37.0)* | 31.0 (26.0-36.0)* |
| Heart rate | 85.0 (73.0-98.0) | 82.0 (71.0-95.0) | 87.0 (75.0-100.0)* | 87.0 (75.0-100.0)* | 91.0 (79.0-104.0)* | 91.5 (78.0-106.0)* |
| Inspired O2 Fraction | 40.0 (30.0-50.0) | 29.7 (24.9-40.0) | 40.0 (35.0-50.0)* | 40.0 (30.0-50.0)* | 40.0 (40.0-50.0)* | 40.0 (40.0-60.0)* |
| O2 flow | 2.0 (2.0-4.0) | 2.0 (2.0-3.0) | 2.0 (2.0-4.0) | 2.0 (2.0-4.0) | 2.0 (2.0-4.0) | 3.0 (2.0-4.0)* |
| PEEP | 5.0 (5.0-8.0) | 5.0 (5.0-5.0) | 5.0 (5.0-8.0) | 5.0 (5.0-8.0) | 5.0 (5.0-10.0) | 5.0 (5.0-10.0) |
| PIP | 20.0 (15.0-26.0) | 13.0 (0.0-20.0) | 22.0 (16.0-27.0)* | 20.0 (15.0-26.0)* | 23.0 (17.0-29.0)* | 24.0 (17.0-30.0)* |
| Respiratory rate | 18.0 (15.0-22.0) | 18.0 (16.0-22.0) | 18.0 (15.0-23.0) | 18.0 (14.0-22.0) | 19.0 (15.0-24.0)* | 20.0 (16.0-25.0)* |
| SBP | 119.0 (105.0-137.0) | 123.0 (108.0-140.0) | 118.0 (104.0-135.0)* | 114.0 (100.0-131.0)* | 109.0 (95.0-125.0)* | 110.0 (96.0-127.0)* |
| SPO2 | 97.3 (95.0-99.9) | 97.0 (95.0-99.0) | 98.0 (95.0-100.0)* | 98.0 (95.0-100.0)* | 98.0 (95.0-100.0)* | 98.0 (95.0-100.0)* |
| Tidal volume | 500.0 (430.0-530.0) | 480.0 (360.0-490.0) | 500.0 (430.0-530.0)* | 500.0 (450.0-550.0)* | 480.0 (420.0-500.0) | 480.0 (420.0-510.0) |

**Assessment scores**

| | | | | | | |
|---|---|---|---|---|---|---|
| CAM | 0.0 (0.0-0.0) | 0.0 (0.0-0.0) | 0.0 (0.0-0.0) | 0.0 (0.0-0.0) | 0.0 (0.0-0.0) | 0.0 (0.0-0.0) |
| GCS | 14.0 (11.0-15.0) | 15.0 (14.0-15.0) | 11.0 (9.0-15.0)* | 13.0 (9.0-15.0)* | 11.0 (8.0-15.0)* | 8.0 (5.0-11.0)* |
| RASS | 0.0 (-1.0-0.0) | 0.0 (0.0-0.0) | -1.0 (-2.0-0.0)* | -1.0 (-2.0-0.0)* | -2.0 (-3.0-0.0)* | -2.0 (-4.0-0.0)* |

**Laboratory values**

| | | | | | | |
|---|---|---|---|---|---|---|
| ALT | 29.0 (17.0-63.0) | 26.0 (16.0-50.0) | 33.0 (18.0-76.0)* | 34.0 (18.0-84.0)* | 55.0 (24.0-187.0)* | 48.0 (23.0-143.0)* |
| AST | 34.0 (20.0-71.0) | 28.0 (18.0-55.0) | 39.0 (22.0-84.0)* | 42.0 (23.0-96.0)* | 73.0 (34.0-215.0)* | 70.0 (33.0-194.0)* |
| Albumin | 2.7 (2.3-3.2) | 2.8 (2.4-3.3) | 2.7 (2.2-3.1)* | 2.6 (2.2-3.0)* | 2.5 (2.1-3.0)* | 2.5 (2.1-3.0)* |
| Anion gap | 11.0 (8.0-13.0) | 10.0 (8.0-13.0) | 10.5 (8.0-13.4)* | 11.0 (8.0-14.0)* | 12.0 (9.0-16.0)* | 12.0 (9.0-16.0)* |
| Arterial CO2 pressure | 40.1 (35.0-46.7) | 41.3 (35.0-50.3) | 40.1 (35.0-46.3)* | 40.0 (34.9-46.0)* | 40.4 (35.0-46.6)* | 40.0 (34.0-47.5)* |
| Arterial O2 Saturation | 95.8 (15.8-98.8) | 95.0 (74.8-98.0) | 96.0 (15.5-98.9)* | 95.0 (14.2-98.9) | 94.2 (12.9-98.6) | 94.1 (14.5-98.6) |
| Arterial O2 pressure | 103.0 (77.3-145.0) | 87.0 (70.1-116.0) | 105.0 (78.5-147.0)* | 108.0 (79.3-150.0)* | 104.0 (76.2-147.0)* | 101.0 (74.1-147.0)* |
| Arterial PH | 7.4 (7.3-7.4) | 7.4 (7.3-7.4) | 7.4 (7.3-7.4) | 7.4 (7.3-7.4) | 7.4 (7.3-7.4) | 7.4 (7.3-7.4) |
| Arterial base excess | 1.0 (-2.9-3.9) | 1.3 (-2.0-5.0) | 1.0 (-2.9-3.8)* | 0.6 (-3.5-3.4)* | -0.4 (-5.0-2.1)* | -0.6 (-5.9-2.7)* |
| BNP | 632.1 (249.0-1675.5) | 621.1 (243.6-1649.0) | 611.0 (243.0-1616.1) | 770.0 (297.0-2250.0)* | 994.0 (346.0-1954.5)* | 789.6 (307.2-2176.0)* |
| Basophils | 0.0 (0.0-0.3) | 0.0 (0.0-0.3) | 0.0 (0.0-0.2) | 0.0 (0.0-0.1) | 0.0 (0.0-0.1) | 0.0 (0.0-0.1) |
| Bilirubin direct | 0.3 (0.2-0.9) | 0.2 (0.2-0.6) | 0.3 (0.2-1.1)* | 0.4 (0.2-1.2)* | 1.2 (0.4-3.6)* | 0.6 (0.2-2.4)* |
| Bilirubin total | 0.7 (0.4-1.3) | 0.6 (0.4-1.1) | 0.7 (0.4-1.5)* | 0.8 (0.5-1.7)* | 1.7 (0.8-4.5)* | 1.1 (0.6-2.9)* |
| CRP | 61.8 (12.0-163.0) | 23.3 (6.0-125.0) | 78.2 (16.9-175.3)* | 93.5 (28.1-186.5)* | 87.7 (23.0-186.9)* | 91.1 (20.7-198.1)* |
| Calcium ionized | 4.5 (4.2-4.8) | 4.6 (4.3-4.8) | 4.5 (4.2-4.8)* | 4.5 (4.2-4.8)* | 4.6 (4.2-4.8) | 4.5 (4.2-4.8)* |
| Calcium non-ionized | 8.4 (7.9-8.8) | 8.5 (8.0-8.9) | 8.3 (7.8-8.8)* | 8.2 (7.7-8.7)* | 8.2 (7.7-8.7)* | 8.1 (7.6-8.6)* |
| Chloride | 104.0 (100.0-108.0) | 103.0 (99.0-107.0) | 104.0 (100.0-109.0)* | 105.0 (100.0-109.0)* | 104.0 (100.0-107.0)* | 105.0 (101.0-110.0)* |
| Creatinine | 1.0 (0.7-1.6) | 0.9 (0.7-1.4) | 1.0 (0.7-1.7)* | 1.1 (0.7-1.9)* | 1.9 (1.2-3.0)* | 1.4 (0.9-2.3)* |
| Eosinophils | 0.3 (0.0-2.0) | 0.7 (0.0-2.0) | 0.3 (0.0-1.8)* | 0.1 (0.0-1.0)* | 0.1 (0.0-0.7)* | 0.0 (0.0-0.3)* |
| Glucose | 132.0 (108.0-167.0) | 132.0 (106.0-174.0) | 132.0 (110.0-164.0) | 132.0 (109.0-163.0) | 134.0 (109.0-167.0) | 139.0 (113.0-173.0)* |
| Hematocrit | 29.6 (26.0-34.3) | 31.0 (26.7-36.0) | 28.9 (25.5-33.1)* | 28.2 (25.1-32.3)* | 26.2 (23.8-29.4)* | 28.0 (24.7-32.5)* |
| Hemoglobin | 9.7 (8.5-11.3) | 10.1 (8.7-11.9) | 9.4 (8.3-10.9)* | 9.2 (8.2-10.6)* | 8.6 (7.8-9.7)* | 9.2 (8.1-10.6)* |
| INR | 1.3 (1.2-1.8) | 1.3 (1.1-1.8) | 1.4 (1.2-1.8)* | 1.4 (1.2-1.9)* | 1.6 (1.3-2.3)* | 1.6 (1.3-2.3)* |
| Lactate | 1.6 (1.1-2.7) | 1.4 (1.0-2.2) | 1.7 (1.1-3.0)* | 1.7 (1.1-3.0)* | 2.3 (1.4-5.3)* | 2.5 (1.5-5.5)* |
| Lymphocytes | 10.0 (5.0-16.7) | 12.0 (6.0-19.6) | 8.7 (4.0-14.2)* | 6.7 (2.0-12.0)* | 4.0 (1.0-8.5)* | 4.6 (1.5-9.0)* |
| Monocytes | 6.0 (1.0-9.0) | 7.0 (1.4-9.5) | 5.0 (0.9-8.5)* | 2.0 (0.6-7.0)* | 1.1 (0.5-6.0)* | 1.9 (0.6-6.0)* |
| Platelets | 189.0 (130.0-262.0) | 198.0 (145.0-263.0) | 182.0 (120.0-261.0)* | 170.0 (109.0-249.0)* | 104.0 (60.0-179.0)* | 134.0 (75.0-211.0)* |
| Potassium | 4.0 (3.7-4.3) | 4.0 (3.7-4.3) | 4.0 (3.7-4.4) | 4.0 (3.7-4.4) | 4.2 (3.9-4.6)* | 4.1 (3.7-4.6)* |
| Sodium | 139.0 (136.0-142.0) | 138.0 (135.0-141.0) | 140.0 (137.0-144.0)* | 140.0 (137.0-144.0)* | 140.0 (137.0-142.0)* | 141.0 (137.0-145.0)* |
| Specific gravity urine | 1.0 (1.0-1.0) | 1.0 (1.0-1.0) | 1.0 (1.0-1.0) | 1.0 (1.0-1.0) | 1.0 (1.0-1.0) | 1.0 (1.0-1.0) |
| Troponin-T | 0.1 (0.0-0.4) | 0.1 (0.0-0.4) | 0.1 (0.0-0.4) | 0.1 (0.0-0.4) | 0.2 (0.1-0.4)* | 0.2 (0.0-0.6)* |
| WBC | 10.2 (7.4-13.9) | 9.2 (6.8-12.4) | 11.0 (8.1-14.9)* | 11.2 (8.0-15.4)* | 12.0 (7.9-17.2)* | 12.4 (8.4-17.5)* |
| **Medications** | | | | | | |
| Amiodarone | 4,460 (3.3%) | 1,274 (1.6%) | 2,774 (5.9%)* | 2,450 (9.5%)* | 240 (15.9%)* | 558 (7.9%)* |
| Dexmedetomidine | 3,070 (2.2%) | 286 (0.4%) | 2,697 (5.8%)* | 1,736 (6.7%)* | 95 (6.3%)* | 212 (3.0%)* |
| Digoxin | 2,891 (2.1%) | 1,404 (1.7%) | 1,205 (2.6%)* | 902 (3.5%)* | 60 (4.0%)* | 220 (3.1%)* |
| Dopamine | 1,687 (1.2%) | 6 (0.0%) | 1,150 (2.5%)* | 1,676 (6.5%)* | 128 (8.5%)* | 456 (6.5%)* |
| Epinephrine | 2,190 (1.6%) | 21 (0.0%) | 1,970 (4.2%)* | 2,153 (8.3%)* | 295 (19.6%)* | 779 (11.0%)* |
| Fentanyl | 3,468 (2.5%) | 250 (0.3%) | 3,130 (6.7%)* | 1,741 (6.7%)* | 136 (9.0%)* | 603 (8.6%)* |
| Folic acid | 7,349 (5.4%) | 3,961 (4.9%) | 2,796 (6.0%)* | 1,814 (7.0%)* | 125 (8.3%)* | 318 (4.5%) |
| Heparin sodium | 26,695 (19.5%) | 14,341 (17.6%) | 9,505 (20.3%)* | 8,266 (32.0%)* | 633 (42.0%)* | 1,743 (24.7%)* |
| Norepinephrine | 6,958 (5.1%) | 13 (0.0%) | 5,411 (11.6%)* | 6,927 (26.8%)* | 456 (30.3%)* | 1,684 (23.9%)* |
| Phenylephrine | 7,599 (5.5%) | 43 (0.1%) | 4,937 (10.6%)* | 7,516 (29.1%)* | 446 (29.6%)* | 1,359 (19.3%)* |
| Propofol | 13,228 (9.7%) | 918 (1.1%) | 10,392 (22.2%)* | 7,538 (29.2%)* | 434 (28.8%)* | 1,330 (18.9%)* |
| Vasopressin | 2,835 (2.1%) | 0 (0.0%) | 2,320 (5.0%)* | 2,834 (11.0%)* | 302 (20.0%)* | 859 (12.2%)* |
| **Life-sustaining therapies** | | | | | | |
| CRRT | 1,507 (1.1%) | 0 (0.0%) | 1,311 (2.8%)* | 1,249 (4.8%)* | 1,507 (100.0%)* | 711 (10.1%)* |
| MV | 46,741 (34.1%) | 0 (0.0%) | 46,741 (100.0%)* | 17,706 (68.6%)* | 1,311 (87.0%)* | 5,916 (83.9%)* |
| VP | 25,806 (18.8%) | 0 (0.0%) | 17,706 (37.9%)* | 25,806 (100.0%)* | 1,249 (82.9%)* | 4,442 (63.0%)* |
| **Outcomes** | | | | | | |
| Deceased | 7,050 (5.1%) | 0 (0.0%) | 5,916 (12.7%)* | 4,442 (17.2%)* | 711 (47.2%)* | 7,050 (100.0%)* |

Statistics are presented as median (interquartile range) for continuous variables, and number (percentage) for categorical variables. P-values for continuous variables are based on pairwise Wilcoxon rank sum test. P-values for categorical variables are based on pairwise Pearson's chi-squared test for proportions.
*p-value < 0.05 compared to stable admissions.

**Supplementary Table S3. External cohort baseline characteristics.**

|  | Overall (*n* = 101,827) | Stable (*n* = 58,982) | MV (*n* = 35,241) | VP (*n* = 24,805) | CRRT (*n* = 1,684) | Deceased (*n* = 5,716) |
|---|---|---|---|---|---|---|
| **Number of patients** | 75,668 | 47,103 | 30,968 | 21,649 | 1,562 | 5,716 |
| **Number of hospital encounters** | 93,718 | 56,015 | 33,919 | 23,820 | 1,621 | 5,716 |
| **Basic information** | | | | | | |
| Age, years | 65.0 (53.0-76.0) | 65.0 (52.0-77.0) | 65.0 (54.0-75.0) | 67.0 (57.0-77.0)* | 62.0 (52.0-72.0)* | 71.0 (60.0-81.0)* |
| BMI, kg/m^2 | 27.5 (26.0-29.1) | 27.5 (26.9-27.8) | 27.5 (24.9-31.1) | 27.5 (25.4-30.4) | 27.6 (26.7-33.4) | 27.5 (24.7-29.1) |
| Female | 46,304 (45.5%) | 28,372 (48.1%) | 14,291 (40.6%) | 10,107 (40.7%) | 660 (39.2%) | 2,658 (46.5%) |
| ICU length of stay, days | 1.9 (1.1-3.6) | 1.5 (0.9-2.4) | 3.2 (1.7-6.6)* | 3.2 (1.7-6.4)* | 8.6 (4.4-14.4)* | 3.2 (1.5-7.1)* |
| **Race** | | | | | | |
| Black | 10,432 (10.2%) | 6,619 (11.2%) | 3,107 (8.8%)* | 1,955 (7.9%)* | 183 (10.9%) | 520 (9.1%)* |
| Other | 18,475 (18.1%) | 9,252 (15.7%) | 7,928 (22.5%)* | 5,364 (21.6%)* | 448 (26.6%)* | 1,434 (25.1%)* |
| White | 72,920 (71.6%) | 43,111 (73.1%) | 24,206 (68.7%)* | 17,486 (70.5%)* | 1,053 (62.5%)* | 3,762 (65.8%)* |
| **Comorbidity index** | | | | | | |
| CCI | 0.0 (0.0-0.0) | 0.0 (0.0-0.0) | 0.0 (0.0-1.0) | 0.0 (0.0-1.0) | 0.0 (0.0-3.0) | 0.0 (0.0-1.0) |
| **Comorbidities** | | | | | | |
| Acquired immunodeficiency syndrome | 865 (0.8%) | 407 (0.7%) | 350 (1.0%)* | 270 (1.1%)* | 24 (1.4%)* | 37 (0.6%) |
| Cancer | 4,867 (4.8%) | 2,253 (3.8%) | 2,141 (6.1%)* | 1,649 (6.6%)* | 142 (8.4%)* | 533 (9.3%)* |
| Cerebrovascular disease | 3,051 (3.0%) | 1,243 (2.1%) | 1,640 (4.7%)* | 1,085 (4.4%)* | 49 (2.9%)* | 268 (4.7%)* |
| Chronic obstructive pulmonary disease | 6,916 (6.8%) | 2,950 (5.0%) | 3,528 (10.0%)* | 2,509 (10.1%)* | 182 (10.8%)* | 496 (8.7%)* |
| Congestive heart failure | 7,015 (6.9%) | 2,948 (5.0%) | 3,404 (9.7%)* | 2,884 (11.6%)* | 244 (14.5%)* | 578 (10.1%)* |
| Dementia | 100 (0.1%) | 59 (0.1%) | 32 (0.1%) | 25 (0.1%) | 1 (0.1%) | 7 (0.1%) |
| Diabetes with complications | 2,670 (2.6%) | 1,090 (1.8%) | 1,302 (3.7%)* | 1,232 (5.0%)* | 160 (9.5%)* | 172 (3.0%)* |
| Diabetes without complications | 4,267 (4.2%) | 1,526 (2.6%) | 2,433 (6.9%)* | 1,990 (8.0%)* | 142 (8.4%)* | 307 (5.4%)* |
| Metastatic carcinoma | 6,166 (6.1%) | 3,143 (5.3%) | 2,233 (6.3%)* | 1,889 (7.6%)* | 96 (5.7%) | 890 (15.6%)* |
| Mild liver disease | 2,513 (2.5%) | 918 (1.6%) | 1,389 (3.9%)* | 1,100 (4.4%)* | 254 (15.1%)* | 352 (6.2%)* |
| Moderate severe liver disease | 2,859 (2.8%) | 948 (1.6%) | 1,680 (4.8%)* | 1,218 (4.9%)* | 384 (22.8%)* | 435 (7.6%)* |
| Myocardial infarction | 3,974 (3.9%) | 1,613 (2.7%) | 2,053 (5.8%)* | 1,881 (7.6%)* | 99 (5.9%)* | 296 (5.2%)* |
| Paraplegia hemiplegia | 1,412 (1.4%) | 449 (0.8%) | 859 (2.4%)* | 546 (2.2%)* | 30 (1.8%)* | 124 (2.2%)* |
| Peptic ulcer disease | 604 (0.6%) | 311 (0.5%) | 276 (0.8%)* | 189 (0.8%)* | 20 (1.2%)* | 39 (0.7%) |
| Peripheral vascular disease | 2,596 (2.5%) | 840 (1.4%) | 1,543 (4.4%)* | 1,353 (5.5%)* | 111 (6.6%)* | 196 (3.4%)* |
| Renal disease | 8,807 (8.6%) | 3,840 (6.5%) | 4,001 (11.4%)* | 3,623 (14.6%)* | 594 (35.3%)* | 787 (13.8%)* |
| Rheumatologic disease | 689 (0.7%) | 297 (0.5%) | 334 (0.9%)* | 299 (1.2%)* | 28 (1.7%)* | 55 (1.0%)* |
| **Vital signs** | | | | | | |
| Body temperature | 36.9 (36.6-37.2) | 36.8 (36.6-37.1) | 37.0 (36.6-37.4)* | 36.9 (36.6-37.3)* | 36.7 (36.3-37.1)* | 36.9 (36.4-37.4)* |
| DBP | 64.0 (55.0-74.0) | 66.0 (57.0-76.0) | 63.0 (54.0-73.0)* | 60.0 (52.0-70.0)* | 59.0 (51.0-69.0)* | 59.0 (50.0-68.0)* |
| EtCO2 | 34.0 (29.0-39.0) | 33.0 (27.0-38.0) | 34.0 (29.0-39.0)* | 33.0 (28.0-39.0) | 33.0 (28.0-37.0) | 32.0 (26.0-37.0)* |
| Heart rate | 85.0 (73.0-97.0) | 83.0 (71.0-96.0) | 86.0 (74.0-98.0)* | 86.0 (75.0-99.0)* | 89.0 (77.0-102.0)* | 90.0 (77.0-104.0)* |
| Inspired O2 Fraction | 40.0 (40.0-50.0) | 45.0 (35.0-60.0) | 40.0 (40.0-50.0)* | 40.0 (40.0-50.0)* | 40.0 (40.0-50.0)* | 50.0 (40.0-60.0)* |
| O2 flow | 4.0 (2.0-10.0) | 3.0 (2.0-5.0) | 4.0 (2.0-10.0)* | 4.0 (2.0-10.0)* | 4.0 (2.0-10.0)* | 6.0 (3.0-15.0)* |
| PEEP | 5.0 (5.0-8.5) | 5.0 (5.0-7.0) | 5.0 (5.0-8.5) | 6.0 (5.0-10.0)* | 9.0 (5.0-12.0)* | 7.0 (5.0-10.0)* |
| PIP | 20.0 (15.0-25.0) | 16.0 (13.0-21.0) | 20.0 (15.0-25.0)* | 20.0 (15.0-26.0)* | 22.0 (16.0-28.0)* | 22.0 (17.0-28.0)* |
| Respiratory rate | 19.0 (16.0-23.0) | 19.0 (16.0-22.0) | 20.0 (16.0-24.0)* | 20.0 (16.0-24.0)* | 20.0 (17.0-25.0)* | 21.0 (17.0-26.0)* |
| SBP | 118.0 (104.0-135.0) | 121.0 (107.0-138.0) | 118.0 (104.0-134.0)* | 112.0 (100.0-127.0)* | 110.0 (98.0-125.0)* | 110.0 (98.0-125.0)* |
| SPO2 | 97.0 (95.0-99.0) | 97.0 (95.0-98.0) | 98.0 (95.0-99.0)* | 97.0 (95.0-99.0) | 98.0 (96.0-100.0)* | 97.0 (95.0-99.0) |
| Tidal volume | 452.0 (383.0-528.0) | 453.0 (365.0-542.0) | 452.0 (383.0-528.0) | 451.0 (381.0-526.0)* | 445.0 (378.0-521.0)* | 438.0 (364.0-508.0)* |
| **Assessment scores** | | | | | | |
| CAM | 0.0 (0.0-0.0) | 0.0 (0.0-0.0) | 0.0 (0.0-1.0) | 0.0 (0.0-1.0) | 0.0 (0.0-1.0) | 0.0 (0.0-1.0) |
| GCS | 14.0 (10.0-15.0) | 15.0 (14.0-15.0) | 11.0 (8.0-15.0)* | 11.0 (8.0-15.0)* | 10.0 (7.0-14.0)* | 8.0 (5.0-11.0)* |
| RASS | 0.0 (-1.0-0.0) | 0.0 (0.0-0.0) | -1.0 (-2.0-0.0)* | -1.0 (-2.0-0.0)* | -1.0 (-3.0-0.0)* | -2.0 (-4.0-0.0)* |
| **Laboratory values** | | | | | | |
| ALT | 33.0 (18.0-80.0) | 28.0 (16.0-58.0) | 38.0 (19.0-100.0)* | 40.0 (20.0-112.0)* | 63.0 (25.0-225.0)* | 48.0 (22.0-150.0)* |
| AST | 42.0 (23.0-98.0) | 32.0 (20.0-68.0) | 50.0 (27.0-120.0)* | 57.0 (30.0-138.0)* | 96.0 (46.0-276.0)* | 78.0 (37.0-214.0)* |
| Albumin | 2.8 (2.3-3.2) | 2.9 (2.4-3.3) | 2.7 (2.3-3.1)* | 2.7 (2.3-3.1)* | 2.8 (2.3-3.2)* | 2.6 (2.2-3.1)* |
| Anion gap | 13.0 (10.0-16.0) | 13.0 (10.0-15.0) | 13.0 (10.0-16.0) | 14.0 (11.0-17.0)* | 17.0 (14.0-20.0)* | 15.0 (12.0-19.0)* |
| Arterial CO2 pressure | 40.0 (35.0-46.0) | 42.0 (35.0-52.0) | 40.0 (35.0-46.0)* | 40.0 (35.0-46.0)* | 39.0 (34.0-45.0)* | 40.0 (34.0-47.0)* |
| Arterial O2 Saturation | 97.0 (95.0-98.0) | 96.0 (93.0-98.0) | 97.0 (95.0-98.0)* | 97.0 (95.0-98.0)* | 96.0 (94.0-98.0) | 96.0 (93.0-98.0) |
| Arterial O2 pressure | 107.0 (83.0-149.0) | 89.0 (71.0-120.0) | 109.0 (84.0-152.0)* | 109.0 (85.0-151.0)* | 102.0 (83.0-130.0)* | 97.0 (78.0-128.0)* |
| Arterial PH | 7.4 (7.3-7.4) | 7.4 (7.3-7.4) | 7.4 (7.3-7.4) | 7.4 (7.3-7.4) | 7.4 (7.3-7.4) | 7.4 (7.3-7.4) |
| Arterial base excess | -0.0 (-3.0-2.0) | 0.7 (-2.0-4.4) | -0.0 (-3.0-2.0)* | -0.0 (-3.0-2.0)* | -1.0 (-5.0--0.0)* | -2.0 (-6.0-1.0)* |
| BNP | 881.0 (298.0-3220.0) | 807.0 (296.0-2780.0) | 842.0 (271.0-3080.0)* | 1991.5 (588.8-7811.5)* | 5521.0 (1495.0-18800.5)* | 2053.5 (662.2-7665.0)* |
| Basophils | 0.0 (0.0-0.4) | 0.1 (0.0-0.5) | 0.0 (0.0-0.3)* | 0.0 (0.0-0.2)* | 0.0 (0.0-0.1)* | 0.0 (0.0-0.1)* |
| Bilirubin direct | 0.8 (0.2-3.0) | 0.4 (0.2-1.5) | 1.2 (0.3-3.8)* | 1.7 (0.5-4.6)* | 3.5 (1.3-7.6)* | 2.0 (0.5-6.0)* |
| Bilirubin total | 0.8 (0.4-1.9) | 0.7 (0.4-1.2) | 0.9 (0.5-2.5)* | 1.1 (0.5-3.1)* | 2.8 (1.0-8.8)* | 1.4 (0.6-4.7)* |
| CRP | 17.8 (5.2-89.7) | 13.5 (4.4-56.0) | 22.1 (6.0-109.0)* | 61.1 (12.9-148.7)* | 42.3 (12.9-140.5)* | 49.0 (10.2-132.0)* |
| Calcium ionized | 1.1 (1.1-1.2) | 1.1 (1.1-1.4) | 1.1 (1.1-1.2) | 1.1 (1.1-1.2) | 1.1 (1.0-1.2) | 1.1 (1.0-1.2) |
| Calcium non-ionized | 8.3 (7.9-8.8) | 8.4 (8.0-8.9) | 8.3 (7.8-8.8)* | 8.2 (7.8-8.7)* | 8.6 (8.0-9.2)* | 8.3 (7.7-8.8)* |
| Chloride | 104.0 (99.0-108.0) | 103.0 (99.0-107.0) | 104.0 (100.0-108.0)* | 104.0 (99.0-108.0)* | 100.0 (96.0-104.0)* | 103.0 (98.0-108.0) |
| Creatinine | 1.0 (0.7-1.7) | 0.9 (0.7-1.4) | 1.0 (0.7-1.8)* | 1.2 (0.8-2.0)* | 2.1 (1.4-3.3)* | 1.5 (0.9-2.5)* |
| Eosinophils | 0.8 (0.0-2.0) | 1.0 (0.0-2.6) | 0.3 (0.0-2.0)* | 0.2 (0.0-1.0)* | 0.1 (0.0-0.7)* | 0.0 (0.0-0.4)* |
| Glucose | 133.0 (108.0-172.0) | 133.0 (106.0-178.0) | 134.0 (110.0-168.0)* | 133.0 (109.0-167.0) | 139.0 (112.0-177.0)* | 142.0 (114.0-180.0)* |
| Hematocrit | 30.0 (26.3-34.5) | 31.2 (27.2-36.0) | 29.2 (26.0-33.5)* | 28.9 (25.0-32.4)* | 27.2 (24.2-31.0)* | 28.7 (25.0-32.6)* |
| Hemoglobin | 9.7 (8.5-11.1) | 10.2 (8.8-11.8) | 9.5 (8.3-10.8)* | 9.3 (8.2-10.5)* | 8.9 (8.0-10.0)* | 9.1 (8.1-10.4)* |
| INR | 1.3 (1.1-1.6) | 1.3 (1.1-1.6) | 1.3 (1.2-1.6) | 1.4 (1.2-1.7)* | 1.5 (1.2-2.1)* | 1.5 (1.2-2.1)* |
| Lactate | 1.8 (1.2-2.8) | 1.6 (1.1-2.3) | 1.8 (1.2-2.9)* | 1.9 (1.3-3.0)* | 2.2 (1.5-3.9)* | 2.4 (1.5-4.4)* |
| Lymphocytes | 9.0 (4.0-16.0) | 11.8 (6.0-19.0) | 7.1 (2.7-13.0)* | 4.0 (1.3-10.4)* | 1.8 (0.8-6.0)* | 3.0 (1.0-7.3)* |
| Monocytes | 6.7 (3.0-9.3) | 7.5 (4.3-10.0) | 6.0 (2.0-9.0)* | 3.1 (0.8-7.4)* | 1.6 (0.6-5.0)* | 2.8 (1.0-6.0)* |
| Platelets | 182.0 (121.0-256.0) | 198.0 (143.0-264.0) | 171.0 (110.0-251.0)* | 158.0 (101.0-238.0)* | 103.0 (60.0-180.0)* | 138.0 (74.0-220.0)* |
| Potassium | 4.0 (3.7-4.4) | 4.0 (3.7-4.4) | 4.0 (3.7-4.4) | 4.1 (3.7-4.4) | 4.2 (3.8-4.6)* | 4.1 (3.8-4.6)* |
| Sodium | 139.0 (136.0-142.0) | 138.0 (135.0-141.0) | 139.0 (136.0-143.0)* | 139.0 (136.0-142.0) | 137.0 (134.0-140.0)* | 139.0 (135.0-143.0)* |
| Specific gravity urine | 1.0 (1.0-1.0) | 1.0 (1.0-1.0) | 1.0 (1.0-1.0) | 1.0 (1.0-1.0) | 1.0 (1.0-1.0) | 1.0 (1.0-1.0) |
| Troponin-T | 0.1 (0.0-0.4) | 0.1 (0.0-0.3) | 0.1 (0.0-0.4) | 0.1 (0.0-0.5) | 0.2 (0.1-0.8)* | 0.2 (0.0-0.6)* |
| WBC | 10.4 (7.6-14.3) | 9.3 (6.9-12.6) | 11.2 (8.2-15.3)* | 11.6 (8.4-16.0)* | 13.1 (8.8-18.7)* | 13.2 (9.1-18.6)* |
| **Medications** | | | | | | |
| Amiodarone | 3,635 (3.6%) | 684 (1.2%) | 2,592 (7.4%)* | 2,575 (10.4%)* | 340 (20.2%)* | 588 (10.3%)* |
| Dexmedetomidine | 6,433 (6.3%) | 365 (0.6%) | 5,899 (16.7%)* | 4,245 (17.1%)* | 410 (24.3%)* | 548 (9.6%)* |
| Digoxin | 1,187 (1.2%) | 572 (1.0%) | 499 (1.4%)* | 388 (1.6%)* | 52 (3.1%)* | 123 (2.2%)* |
| Dopamine | 2,067 (2.0%) | 0 (0.0%) | 1,126 (3.2%)* | 2,067 (8.3%)* | 195 (11.6%)* | 557 (9.7%)* |
| Epinephrine | 2,746 (2.7%) | 0 (0.0%) | 2,598 (7.4%)* | 2,746 (11.1%)* | 359 (21.3%)* | 588 (10.3%)* |
| Fentanyl | 18,920 (18.6%) | 2,038 (3.5%) | 16,247 (46.1%)* | 11,080 (44.7%)* | 1,132 (67.2%)* | 2,439 (42.7%)* |
| Folic acid | 3,177 (3.1%) | 1,551 (2.6%) | 1,472 (4.2%)* | 702 (2.8%) | 95 (5.6%)* | 152 (2.7%) |
| Heparin sodium | 9,930 (9.8%) | 4,748 (8.0%) | 4,055 (11.5%)* | 4,015 (16.2%)* | 516 (30.6%)* | 1,010 (17.7%)* |
| Norepinephrine | 13,318 (13.1%) | 1 (0.0%) | 9,330 (26.5%)* | 13,315 (53.7%)* | 1,256 (74.6%)* | 3,102 (54.3%)* |
| Phenylephrine | 12,740 (12.5%) | 0 (0.0%) | 10,789 (30.6%)* | 12,738 (51.4%)* | 771 (45.8%)* | 1,753 (30.7%)* |
| Propofol | 23,928 (23.5%) | 580 (1.0%) | 22,914 (65.0%)* | 14,586 (58.8%)* | 1,079 (64.1%)* | 2,255 (39.5%)* |
| Vasopressin | 4,003 (3.9%) | 0 (0.0%) | 3,481 (9.9%)* | 4,003 (16.1%)* | 886 (52.6%)* | 1,778 (31.1%)* |
| **Life-sustaining therapies** | | | | | | |
| CRRT | 1,684 (1.7%) | 0 (0.0%) | 1,378 (3.9%)* | 1,467 (5.9%)* | 1,684 (100.0%)* | 724 (12.7%)* |
| MV | 35,241 (34.6%) | 0 (0.0%) | 35,241 (100.0%)* | 18,167 (73.2%)* | 1,378 (81.8%)* | 4,195 (73.4%)* |
| VP | 24,805 (24.4%) | 0 (0.0%) | 18,167 (51.6%)* | 24,805 (100.0%)* | 1,467 (87.1%)* | 3,839 (67.2%)* |
| **Outcomes** | | | | | | |
| Deceased | 5,716 (5.6%) | 0 (0.0%) | 4,195 (11.9%)* | 3,839 (15.5%)* | 724 (43.0%)* | 5,716 (100.0%)* |

Statistics are presented as median (interquartile range) for continuous variables, and number (percentage) for categorical variables. P-values for continuous variables are based on pairwise Wilcoxon rank sum test. P-values for categorical variables are based on pairwise Pearson's chi-squared test for proportions.
*p-value < 0.05 compared to stable admissions.

**Supplementary Table S4. Temporal cohort baseline characteristics.**

|  |  | Overall (*n* = 15,940) | Stable (*n* = 9,477) | MV (*n* = 4,595) | VP (*n* = 4,586) | CRRT (*n* = 392) | Deceased (*n* = 942) |
|---|---|---|---|---|---|---|---|
| **Number of patients** | | 12,927 | 8,083 | 4,257 | 4,269 | 383 | 942 |
| **Number of hospital encounters** | | 15,940 | 9,477 | 4,595 | 4,586 | 392 | 942 |
| **Basic information** | | | | | | | |
| | Age, years | 62.0 (49.0-72.0) | 62.0 (49.0-72.0) | 61.0 (49.0-71.0)* | 63.0 (51.0-72.0)* | 61.0 (52.0-68.0) | 65.0 (56.2-75.0)* |
| | BMI, kg/m^2 | 27.5 (23.3-32.3) | 27.5 (23.3-32.2) | 27.5 (23.4-32.6) | 27.4 (23.2-32.3) | 28.7 (24.3-34.4)* | 27.3 (22.9-32.1) |
| | Female | 7,158 (44.9%) | 4,349 (45.9%) | 1,953 (42.5%)* | 1,958 (42.7%)* | 157 (40.1%)* | 419 (44.5%) |
| | ICU length of stay, days | 3.1 (1.7-6.2) | 2.2 (1.3-3.9) | 5.9 (3.1-11.2)* | 6.2 (3.4-11.7)* | 9.1 (4.2-16.3)* | 4.2 (2.0-9.2)* |
| **Race** | | | | | | | |
| | Black | 3,046 (19.1%) | 1,831 (19.3%) | 911 (19.8%) | 801 (17.5%)* | 93 (23.7%)* | 193 (20.5%) |
| | Other | 1,013 (6.4%) | 591 (6.2%) | 295 (6.4%) | 310 (6.8%) | 31 (7.9%) | 71 (7.5%) |
| | White | 11,881 (74.5%) | 7,055 (74.4%) | 3,389 (73.8%) | 3,475 (75.8%) | 268 (68.4%)* | 678 (72.0%) |
| **Comorbidity index** | | | | | | | |
| | CCI | 2.0 (1.0-4.0) | 2.0 (0.0-4.0) | 2.0 (0.0-4.0) | 2.0 (1.0-4.0) | 4.0 (2.0-6.0)* | 3.0 (2.0-6.0)* |
| **Comorbidities** | | | | | | | |
| | Acquired immunodeficiency syndrome | 0 (0.0%) | 0 (0.0%) | 0 (0.0%) | 0 (0.0%) | 0 (0.0%) | 0 (0.0%) |
| | Cancer | 2,043 (12.8%) | 1,268 (13.4%) | 473 (10.3%)* | 564 (12.3%) | 39 (9.9%)* | 173 (18.4%)* |
| | Cerebrovascular disease | 2,211 (13.9%) | 1,272 (13.4%) | 723 (15.7%)* | 635 (13.8%) | 40 (10.2%) | 221 (23.5%)* |
| | Chronic obstructive pulmonary disease | 4,761 (29.9%) | 2,810 (29.7%) | 1,411 (30.7%) | 1,323 (28.8%) | 108 (27.6%) | 283 (30.0%) |
| | Congestive heart failure | 4,806 (30.2%) | 2,750 (29.0%) | 1,444 (31.4%)* | 1,525 (33.3%)* | 207 (52.8%)* | 382 (40.6%)* |
| | Dementia | 715 (4.5%) | 407 (4.3%) | 204 (4.4%) | 203 (4.4%) | 10 (2.6%) | 66 (7.0%)* |
| | Diabetes with complications | 2,544 (16.0%) | 1,585 (16.7%) | 633 (13.8%)* | 711 (15.5%) | 126 (32.1%)* | 161 (17.1%) |
| | Diabetes without complications | 3,128 (19.6%) | 2,011 (21.2%) | 795 (17.3%)* | 767 (16.7%)* | 74 (18.9%) | 184 (19.5%) |
| | Metastatic carcinoma | 916 (5.7%) | 579 (6.1%) | 184 (4.0%)* | 250 (5.5%) | 12 (3.1%)* | 97 (10.3%)* |
| | Mild liver disease | 1,246 (7.8%) | 640 (6.8%) | 436 (9.5%)* | 464 (10.1%)* | 80 (20.4%)* | 140 (14.9%)* |
| | Moderate severe liver disease | 465 (2.9%) | 189 (2.0%) | 201 (4.4%)* | 222 (4.8%)* | 63 (16.1%)* | 88 (9.3%)* |
| | Myocardial infarction | 2,741 (17.2%) | 1,682 (17.7%) | 770 (16.8%) | 791 (17.2%) | 91 (23.2%)* | 196 (20.8%)* |
| | Paraplegia hemiplegia | 970 (6.1%) | 522 (5.5%) | 331 (7.2%)* | 298 (6.5%)* | 9 (2.3%)* | 103 (10.9%)* |
| | Peptic ulcer disease | 253 (1.6%) | 118 (1.2%) | 89 (1.9%)* | 107 (2.3%)* | 14 (3.6%)* | 20 (2.1%)* |
| | Peripheral vascular disease | 2,932 (18.4%) | 1,560 (16.5%) | 955 (20.8%)* | 1,137 (24.8%)* | 125 (31.9%)* | 187 (19.9%)* |
| | Renal disease | 3,577 (22.4%) | 2,123 (22.4%) | 972 (21.2%) | 1,084 (23.6%) | 237 (60.5%)* | 298 (31.6%)* |
| | Rheumatologic disease | 485 (3.0%) | 299 (3.2%) | 135 (2.9%) | 126 (2.7%) | 13 (3.3%) | 34 (3.6%) |
| **Vital signs** | | | | | | | |
| | Body temperature | 37.3 (36.8-37.7) | 37.3 (37.1-37.6) | 37.2 (36.7-37.7)* | 37.2 (36.6-37.6)* | 37.0 (36.4-37.5)* | 37.1 (36.4-37.6)* |
| | DBP | 63.0 (54.0-73.0) | 67.0 (58.0-78.0) | 62.0 (53.0-72.0)* | 61.0 (53.0-71.0)* | 57.0 (49.0-66.0)* | 58.0 (50.0-67.0)* |
| | EtCO2 | 34.0 (30.0-38.0) | 34.0 (29.0-37.0) | 34.0 (29.0-38.0) | 34.0 (30.0-38.0) | 31.0 (26.0-36.0)* | 30.0 (25.0-36.0)* |
| | Heart rate | 85.0 (73.0-98.5) | 82.0 (71.0-95.0) | 87.5 (75.0-100.5)* | 87.0 (74.5-100.0)* | 90.0 (77.0-102.5)* | 92.5 (79.0-106.5)* |
| | Inspired O2 Fraction | 40.0 (29.7-40.0) | 29.7 (24.9-40.0) | 40.0 (30.0-40.0)* | 40.0 (30.0-40.0)* | 40.0 (34.6-45.0)* | 40.0 (35.0-50.0)* |
| | O2 flow | 2.0 (2.0-4.0) | 2.0 (2.0-3.0) | 2.0 (2.0-4.0) | 2.0 (2.0-4.0) | 3.0 (2.0-4.0)* | 3.0 (2.0-4.0)* |
| | PEEP | 5.0 (5.0-9.0) | | 5.0 (5.0-9.0) | 5.0 (5.0-10.0) | 8.0 (5.0-10.0) | 8.0 (5.0-10.0) |
| | PIP | 18.0 (3.0-24.0) | 3.0 (0.0-19.0) | 20.0 (9.0-25.0)* | 18.0 (8.0-24.0)* | 21.0 (2.0-26.0)* | 22.0 (13.0-27.0)* |
| | Respiratory rate | 17.0 (13.0-21.0) | 18.0 (15.0-21.0) | 16.0 (13.0-21.0)* | 16.0 (12.0-21.0)* | 17.0 (13.0-22.0)* | 18.0 (14.0-23.0) |
| | SBP | 117.0 (102.0-134.0) | 121.0 (107.0-137.0) | 116.0 (101.0-134.0)* | 114.0 (100.0-131.0)* | 110.0 (95.0-128.0)* | 108.0 (94.0-126.0)* |
| | SPO2 | 97.3 (95.0-100.0) | 96.0 (94.0-98.0) | 98.0 (95.0-100.0)* | 98.0 (95.0-100.0)* | 98.0 (95.0-100.0)* | 97.0 (94.0-100.0)* |
| | Tidal volume | 450.0 (400.0-500.0) | | 450.0 (400.0-500.0) | 450.0 (400.0-500.0) | 450.0 (400.0-490.0) | 440.0 (400.0-500.0) |
| **Assessment scores** | | | | | | | |

|  | | | | | | |
|---|---|---|---|---|---|---|
| CAM | 0.0 (0.0-0.0) | 0.0 (0.0-0.0) | 0.0 (0.0-0.0) | 0.0 (0.0-0.0) | 0.0 (0.0-0.0) | 0.0 (0.0-0.0) |
| GCS | 15.0 (11.0-15.0) | 15.0 (15.0-15.0) | 11.0 (9.0-15.0)* | 14.0 (10.0-15.0)* | 11.0 (9.0-15.0)* | 9.0 (6.0-13.0)* |
| RASS | 0.0 (-1.0-0.0) | 0.0 (0.0-0.0) | 0.0 (-2.0-0.0) | 0.0 (-1.0-0.0) | -1.0 (-2.0-0.0)* | -2.0 (-4.0-0.0)* |
| **Laboratory values** | | | | | | |
| ALT | 25.0 (13.0-64.0) | 20.0 (11.0-42.0) | 31.0 (15.0-83.0)* | 30.0 (15.0-80.0)* | 58.0 (20.0-239.0)* | 42.0 (17.0-151.0)* |
| AST | 34.0 (20.0-72.0) | 27.0 (18.0-48.0) | 40.0 (23.0-92.0)* | 39.0 (22.0-89.0)* | 78.0 (34.0-271.0)* | 63.0 (30.0-201.2)* |
| Albumin | 2.9 (2.5-3.3) | 3.2 (2.7-3.6) | 2.8 (2.4-3.2)* | 2.8 (2.4-3.2)* | 2.6 (2.2-3.0)* | 2.6 (2.2-3.0)* |
| Anion gap | 18.0 (16.0-21.0) | 18.0 (16.0-19.0) | 18.0 (16.0-21.0) | 18.0 (16.0-21.0) | 19.0 (16.0-22.0)* | 19.0 (16.0-22.0)* |
| Arterial $CO_2$ pressure | 39.7 (35.0-45.0) | 39.9 (34.6-46.4) | 39.8 (35.1-45.0) | 39.8 (35.0-45.0) | 39.9 (34.8-45.7) | 39.3 (34.0-45.9)* |
| Arterial $O_2$ Saturation | 93.3 (13.0-98.8) | 89.4 (14.3-97.9) | 93.4 (12.9-98.8)* | 93.9 (12.7-98.8)* | 92.2 (11.8-98.7) | 90.5 (12.3-98.5) |
| Arterial $O_2$ pressure | 124.0 (91.4-160.0) | 99.5 (78.6-131.0) | 126.0 (93.2-161.0)* | 125.0 (92.7-161.0)* | 120.0 (87.6-158.0)* | 118.0 (87.1-155.0)* |
| Arterial PH | 7.4 (7.4-7.4) | 7.4 (7.4-7.4) | 7.4 (7.4-7.4) | 7.4 (7.3-7.4) | 7.4 (7.3-7.4) | 7.4 (7.3-7.4) |
| Arterial base excess | -0.0 (-3.0-2.8) | 0.2 (-2.4-3.0) | -0.0 (-3.1-2.8)* | -0.1 (-3.3-2.7)* | -1.3 (-5.4-1.8)* | -1.5 (-5.5-1.7)* |
| BNP | 264.0 (105.0-645.0) | 196.0 (79.0-547.0) | 279.0 (113.0-643.0)* | 308.0 (126.3-692.8)* | 606.0 (304.0-1195.0)* | 489.0 (209.0-1202.0)* |
| Basophils | 0.0 (0.0-0.1) | 0.0 (0.0-0.1) | 0.0 (0.0-0.1) | 0.0 (0.0-0.1) | 0.0 (0.0-0.1) | 0.0 (0.0-0.1) |
| Bilirubin direct | 0.2 (0.1-0.6) | 0.2 (0.1-0.4) | 0.3 (0.1-0.8)* | 0.3 (0.1-0.9)* | 0.9 (0.3-3.0)* | 0.6 (0.2-2.3)* |
| Bilirubin total | 0.7 (0.5-1.4) | 0.6 (0.4-1.0) | 0.8 (0.5-1.6)* | 0.8 (0.5-1.7)* | 1.7 (0.7-4.7)* | 1.3 (0.6-3.5)* |
| CRP | 106.3 (42.7-184.5) | 79.6 (27.6-169.4) | 115.5 (50.5-194.8)* | 113.2 (50.3-193.4)* | 122.1 (56.9-208.5)* | 119.1 (57.5-200.6)* |
| Calcium ionized | 4.6 (4.3-4.8) | 4.6 (4.4-4.8) | 4.6 (4.3-4.8) | 4.6 (4.3-4.8) | 4.6 (4.4-4.8) | 4.5 (4.3-4.8)* |
| Calcium non-ionized | 8.4 (7.9-8.9) | 8.6 (8.2-9.1) | 8.3 (7.8-8.7)* | 8.3 (7.8-8.7)* | 8.2 (7.8-8.7)* | 8.2 (7.7-8.7)* |
| Chloride | 105.0 (101.0-108.0) | 104.0 (100.0-107.0) | 106.0 (102.0-110.0)* | 105.0 (101.0-109.0)* | 104.0 (101.0-107.0) | 105.0 (101.0-110.0)* |
| Creatinine | 0.9 (0.7-1.5) | 0.9 (0.7-1.3) | 1.0 (0.6-1.6)* | 1.0 (0.7-1.7)* | 1.8 (1.1-2.8)* | 1.3 (0.9-2.2)* |
| Eosinophils | 0.1 (0.0-0.2) | 0.1 (0.0-0.2) | 0.1 (0.0-0.2) | 0.1 (0.0-0.2) | 0.0 (0.0-0.2)* | 0.0 (0.0-0.1)* |
| Glucose | 130.0 (108.0-161.0) | 122.0 (103.0-156.0) | 135.0 (113.0-164.0)* | 133.0 (111.0-162.0)* | 136.0 (113.0-164.0)* | 140.0 (114.0-173.0)* |
| Hematocrit | 28.4 (24.8-33.6) | 31.6 (26.7-36.7) | 27.0 (24.1-31.4)* | 27.0 (24.1-31.2)* | 25.1 (23.1-28.1)* | 26.7 (23.9-30.9)* |
| Hemoglobin | 9.5 (8.3-11.3) | 10.6 (8.9-12.3) | 9.1 (8.1-10.5)* | 9.1 (8.1-10.5)* | 8.4 (7.7-9.4)* | 8.9 (8.0-10.3)* |
| INR | 1.5 (1.2-2.1) | 1.1 (1.1-1.3) | 1.5 (1.2-2.0)* | 1.6 (1.3-2.0)* | 1.9 (1.8-2.6)* | 1.9 (1.7-2.7)* |
| Lactate | 1.5 (1.0-2.5) | 1.4 (1.0-2.1) | 1.6 (1.1-2.7)* | 1.6 (1.0-2.7)* | 2.1 (1.2-4.7)* | 2.3 (1.4-4.9)* |
| Lymphocytes | 1.0 (0.6-1.5) | 1.1 (0.7-1.7) | 0.9 (0.5-1.4)* | 0.9 (0.5-1.4)* | 0.7 (0.4-1.2)* | 0.7 (0.4-1.2)* |
| Monocytes | 0.7 (0.4-1.0) | 0.7 (0.5-0.9) | 0.7 (0.4-1.0) | 0.7 (0.4-1.0) | 0.6 (0.3-1.1)* | 0.6 (0.3-1.0)* |
| Platelets | 185.0 (125.0-259.0) | 200.0 (150.0-265.0) | 171.0 (109.0-253.0)* | 171.0 (109.0-252.0)* | 99.0 (61.8-170.0)* | 123.0 (72.0-203.0)* |
| Potassium | 4.0 (3.7-4.3) | 4.0 (3.7-4.3) | 4.0 (3.7-4.3) | 4.0 (3.7-4.3) | 4.0 (3.8-4.4) | 4.1 (3.8-4.5)* |
| Sodium | 139.0 (136.0-142.0) | 138.0 (135.0-140.0) | 140.0 (137.0-144.0)* | 139.0 (136.0-143.0)* | 138.0 (136.0-141.0) | 140.0 (136.0-146.0)* |
| WBC | 10.1 (7.3-13.9) | 8.7 (6.5-11.8) | 11.3 (8.2-15.4)* | 10.9 (7.9-15.1)* | 12.5 (8.5-17.7)* | 12.6 (8.5-18.0)* |
| **Medications** | | | | | | |
| Amiodarone | 1,550 (9.7%) | 477 (5.0%) | 887 (19.3%)* | 919 (20.0%)* | 121 (30.9%)* | 181 (19.2%)* |
| Dexmedetomidine | 2,302 (14.4%) | 152 (1.6%) | 2,038 (44.4%)* | 1,590 (34.7%)* | 157 (40.1%)* | 232 (24.6%)* |
| Digoxin | 335 (2.1%) | 175 (1.8%) | 102 (2.2%) | 139 (3.0%)* | 14 (3.6%)* | 25 (2.7%) |
| Dopamine | 269 (1.7%) | 3 (0.0%) | 140 (3.0%)* | 266 (5.8%)* | 44 (11.2%)* | 78 (8.3%)* |
| Epinephrine | 1,004 (6.3%) | 37 (0.4%) | 831 (18.1%)* | 943 (20.6%)* | 173 (44.1%)* | 312 (33.1%)* |
| Fentanyl | 192 (1.2%) | 90 (0.9%) | 73 (1.6%)* | 74 (1.6%)* | 9 (2.3%)* | 7 (0.7%) |
| Folic acid | 1,493 (9.4%) | 817 (8.6%) | 518 (11.3%)* | 448 (9.8%)* | 48 (12.2%)* | 73 (7.7%) |
| Heparin sodium | 11,552 (72.5%) | 6,339 (66.9%) | 3,745 (81.5%)* | 3,793 (82.7%)* | 365 (93.1%)* | 702 (74.5%)* |
| Norepinephrine | 1,069 (6.7%) | 6 (0.1%) | 821 (17.9%)* | 1,056 (23.0%)* | 161 (41.1%)* | 317 (33.7%)* |
| Phenylephrine | 3,234 (20.3%) | 37 (0.4%) | 1,960 (42.7%)* | 3,108 (67.8%)* | 242 (61.7%)* | 440 (46.7%)* |
| Propofol | 4,582 (28.7%) | 456 (4.8%) | 3,193 (69.5%)* | 2,988 (65.2%)* | 231 (58.9%)* | 439 (46.6%)* |
| Vasopressin | 688 (4.3%) | 1 (0.0%) | 566 (12.3%)* | 684 (14.9%)* | 97 (24.7%)* | 121 (12.8%)* |
| **Life-sustaining therapies** | | | | | | |
| CRRT | 392 (2.5%) | 0 (0.0%) | 345 (7.5%)* | 371 (8.1%)* | 392 (100.0%)* | 215 (22.8%)* |
| MV | 4,595 (28.8%) | 0 (0.0%) | 4,595 (100.0%)* | 2,816 (61.4%)* | 345 (88.0%)* | 759 (80.6%)* |
| VP | 4,586 (28.8%) | 0 (0.0%) | 2,816 (61.3%)* | 4,586 (100.0%)* | 371 (94.6%)* | 735 (78.0%)* |
| **Outcomes** | | | | | | |
| Deceased | 942 (5.9%) | 0 (0.0%) | 759 (16.5%)* | 735 (16.0%)* | 215 (54.8%)* | 942 (100.0%)* |

Statistics are presented as median (interquartile range) for continuous variables, and number (percentage) for categorical variables. P-values for continuous variables are based on pairwise Wilcoxon rank sum test. P-values for categorical variables are based on pairwise Pearson's chi-squared test for proportions.
*p-value < 0.05 compared to stable admissions.

**Supplementary Table S5. Prospective cohort baseline characteristics.**

|  |  | Overall (*n* = 368) | Stable (*n* = 150) | MV (*n* = 153) | VP (*n* = 186) | CRRT (*n* = 4) | Deceased (*n* = 10) |
|---|---|---|---|---|---|---|---|
| **Number of patients** | | 215 | 107 | 130 | 144 | 4 | 10 |
| **Number of hospital encounters** | | 325 | 142 | 145 | 169 | 4 | 10 |
| **Basic information** | | | | | | | |
| | Age, years | 60.0 (48.0-68.0) | 61.0 (51.0-70.0) | 59.0 (45.0-68.0) | 60.0 (46.0-68.0) | 62.5 (58.0-64.8) | 56.0 (45.8-65.8) |
| | BMI, kg/m^2 | 26.6 (23.0-30.5) | 26.3 (23.0-30.3) | 27.4 (23.5-32.5) | 26.6 (23.4-30.6) | 29.9 (28.6-31.7)* | 25.4 (22.9-29.5) |
| | Female | 147 (39.9%) | 60 (40.0%) | 61 (39.9%) | 74 (39.8%) | 4 (100.0%)* | 5 (50.0%) |
| | ICU length of stay, days | 6.2 (3.1-12.1) | 3.5 (2.0-6.3) | 10.6 (5.7-17.2)* | 10.3 (5.7-16.7)* | 16.9 (15.9-19.8)* | 17.2 (14.0-21.6)* |
| **Race** | | | | | | | |
| | Black | 53 (14.4%) | 26 (17.3%) | 22 (14.4%) | 22 (11.8%) | 1 (25.0%) | 2 (20.0%) |
| | Other | 30 (8.2%) | 9 (6.0%) | 13 (8.5%) | 18 (9.7%) | 0 (0.0%) | 2 (20.0%) |
| | White | 285 (77.4%) | 115 (76.7%) | 118 (77.1%) | 146 (78.5%) | 3 (75.0%) | 6 (60.0%) |
| **Comorbidity index** | | | | | | | |
| | CCI | 2.0 (1.0-4.0) | 2.0 (1.0-3.0) | 3.0 (1.0-6.0)* | 2.0 (0.2-5.8) | 5.5 (3.2-7.2) | 7.0 (4.8-8.8)* |
| **Comorbidities** | | | | | | | |
| | Acquired immunodeficiency syndrome | 5 (1.4%) | 0 (0.0%) | 2 (1.3%) | 5 (2.7%)* | 0 (0.0%) | 0 (0.0%) |
| | Cancer | 27 (7.3%) | 8 (5.3%) | 15 (9.8%) | 15 (8.1%) | 0 (0.0%) | 1 (10.0%) |
| | Cerebrovascular disease | 45 (12.2%) | 16 (10.7%) | 18 (11.8%) | 29 (15.6%) | 1 (25.0%) | 0 (0.0%) |
| | Chronic obstructive pulmonary disease | 73 (19.8%) | 35 (23.3%) | 24 (15.7%) | 31 (16.7%) | 0 (0.0%) | 1 (10.0%) |
| | Congestive heart failure | 88 (23.9%) | 37 (24.7%) | 37 (24.2%) | 43 (23.1%) | 1 (25.0%) | 4 (40.0%) |
| | Dementia | 16 (4.3%) | 7 (4.7%) | 7 (4.6%) | 8 (4.3%) | 0 (0.0%) | 0 (0.0%) |
| | Diabetes with complications | 62 (16.8%) | 29 (19.3%) | 21 (13.7%) | 28 (15.1%) | 2 (50.0%) | 2 (20.0%) |
| | Diabetes without complications | 68 (18.5%) | 33 (22.0%) | 24 (15.7%) | 32 (17.2%) | 1 (25.0%) | 1 (10.0%) |
| | Metastatic carcinoma | 10 (2.7%) | 2 (1.3%) | 6 (3.9%) | 6 (3.2%) | 0 (0.0%) | 1 (10.0%) |
| | Mild liver disease | 89 (24.2%) | 37 (24.7%) | 36 (23.5%) | 46 (24.7%) | 1 (25.0%) | 4 (40.0%) |
| | Moderate severe liver disease | 48 (13.0%) | 11 (7.3%) | 28 (18.3%)* | 31 (16.7%)* | 2 (50.0%)* | 5 (50.0%)* |
| | Myocardial infarction | 45 (12.2%) | 22 (14.7%) | 16 (10.5%) | 18 (9.7%) | 0 (0.0%) | 1 (10.0%) |
| | Paraplegia hemiplegia | 37 (10.1%) | 11 (7.3%) | 19 (12.4%) | 23 (12.4%) | 0 (0.0%) | 1 (10.0%) |
| | Peptic ulcer disease | 9 (2.4%) | 1 (0.7%) | 6 (3.9%) | 8 (4.3%)* | 0 (0.0%) | 0 (0.0%) |
| | Peripheral vascular disease | 84 (22.8%) | 27 (18.0%) | 42 (27.5%)* | 49 (26.3%) | 1 (25.0%) | 5 (50.0%)* |
| | Renal disease | 109 (29.6%) | 42 (28.0%) | 47 (30.7%) | 56 (30.1%) | 4 (100.0%)* | 6 (60.0%)* |
| | Rheumatologic disease | 15 (4.1%) | 9 (6.0%) | 4 (2.6%) | 4 (2.2%) | 0 (0.0%) | 1 (10.0%) |
| **Vital signs** | | | | | | | |
| | Body temperature | 37.2 (36.8-37.6) | 37.2 (37.0-37.5) | 37.2 (36.7-37.6) | 37.2 (36.7-37.6) | 36.9 (36.4-37.3) | 37.2 (36.8-37.5) |
| | DBP | 69.0 (59.6-79.0) | 72.4 (64.0-81.0) | 66.8 (58.0-76.0)* | 67.0 (58.0-77.0)* | 61.0 (52.0-75.0) | 62.0 (54.0-69.1)* |
| | EtCO2 | 34.0 (30.0-38.0) | 34.0 (32.0-35.0) | 34.0 (30.0-38.0) | 34.0 (30.0-38.0) | 30.0 (26.0-33.0) | 33.0 (28.0-36.0) |
| | Heart rate | 84.0 (73.0-96.0) | 84.0 (74.0-93.0) | 84.0 (73.0-96.0) | 84.0 (73.0-96.0) | 78.0 (70.0-88.0) | 85.0 (72.0-98.0) |
| | Inspired O2 Fraction | 28.0 (21.0-40.0) | 21.0 (21.0-28.0) | 35.0 (28.0-40.0)* | 30.0 (21.0-40.0)* | 40.0 (28.0-40.0)* | 40.0 (40.0-50.0)* |
| | O2 flow | 3.0 (2.0-6.0) | 2.0 (2.0-4.0) | 3.0 (2.0-8.0)* | 3.0 (2.0-8.0)* | 2.0 (2.0-3.0) | 25.0 (2.0-35.0)* |
| | PEEP | 5.0 (5.0-10.0) | | 5.0 (5.0-10.0) | 5.0 (5.0-10.0) | 10.0 (5.0-10.0) | 8.0 (5.0-10.0) |
| | PIP | 18.0 (13.0-23.0) | 20.0 (1.0-23.0) | 19.0 (14.0-24.0) | 18.0 (14.0-23.0) | 25.0 (20.0-27.0) | 23.0 (17.0-27.0) |
| | Respiratory rate | 16.0 (13.0-20.0) | 18.0 (15.0-21.0) | 16.0 (13.0-20.0)* | 16.0 (12.0-20.0)* | 16.0 (13.0-19.0) | 18.0 (15.0-22.0) |
| | SBP | 118.0 (105.0-133.0) | 120.0 (107.0-134.0) | 118.0 (104.0-134.0) | 117.0 (104.0-133.0) | 122.0 (105.0-139.0) | 108.0 (96.9-126.0) |
| | SPO2 | 98.0 (95.2-100.0) | 97.0 (95.0-99.0) | 98.0 (96.0-100.0)* | 98.0 (96.0-100.0)* | 97.0 (94.0-99.0) | 97.0 (95.0-99.0) |
| | Tidal volume | 483.0 (404.0-533.0) | 474.5 (354.0-502.8) | 489.0 (413.0-535.0) | 485.0 (407.0-535.0) | 393.5 (330.0-486.0) | 502.0 (381.0-552.0) |
| **Assessment scores** | | | | | | | |

| | | | | | | |
|---|---|---|---|---|---|---|
| CAM | 0.0 (0.0-0.0) | 0.0 (0.0-0.0) | 0.0 (0.0-0.0) | 0.0 (0.0-0.0) | 1.0 (0.8-1.0)* | 0.0 (0.0-1.0) |
| GCS | 15.0 (11.0-15.0) | 15.0 (15.0-15.0) | 14.0 (10.0-15.0)* | 15.0 (11.0-15.0) | 9.0 (8.0-14.0) | 10.0 (9.0-14.0)* |
| RASS | 0.0 (-1.0-0.0) | 0.0 (0.0-0.0) | 0.0 (-1.0-0.0) | 0.0 (-1.0-0.0) | -2.0 (-3.0-0.0) | -1.0 (-2.0-0.0) |
| **Laboratory values** | | | | | | |
| ALT | 30.0 (15.0-68.0) | 22.0 (11.0-43.0) | 36.0 (17.0-84.0)* | 34.0 (16.0-73.0)* | 19.0 (14.0-24.5) | 43.0 (19.0-77.0) |
| AST | 34.0 (21.0-72.2) | 25.0 (19.0-41.0) | 43.0 (24.0-105.0)* | 38.0 (22.0-91.0)* | 37.0 (28.0-50.0) | 60.0 (31.0-150.0) |
| Albumin | 3.1 (2.8-3.6) | 3.4 (2.9-3.7) | 3.1 (2.7-3.5)* | 3.1 (2.7-3.4)* | 3.1 (2.8-4.1) | 3.3 (2.9-4.0) |
| Anion gap | 23.0 (23.0-23.0) | | 23.0 (23.0-23.0) | 23.0 (23.0-23.0) | 23.0 (23.0-23.0) | 23.0 (23.0-23.0) |
| Arterial $CO_2$ pressure | 39.0 (34.9-44.8) | 36.0 (32.7-39.5) | 39.1 (35.0-44.9)* | 39.0 (34.9-44.9)* | 33.4 (31.3-35.8) | 38.4 (34.8-44.7) |
| Arterial $O_2$ Saturation | 87.8 (11.9-98.6) | 86.6 (13.3-98.6) | 87.9 (11.8-98.6) | 87.7 (11.8-98.6) | 89.8 (10.3-98.7) | 87.6 (11.0-98.1) |
| Arterial $O_2$ pressure | 116.0 (86.3-152.0) | 103.0 (87.4-144.0) | 116.0 (86.0-152.0)* | 116.0 (86.1-152.0)* | 107.0 (83.8-152.0) | 103.0 (80.2-145.0) |
| Arterial PH | 7.4 (7.4-7.4) | 7.4 (7.4-7.4) | 7.4 (7.4-7.4) | 7.4 (7.4-7.4) | 7.4 (7.4-7.5) | 7.4 (7.3-7.4) |
| Arterial base excess | -0.1 (-2.6-2.1) | -2.0 (-4.3--0.9) | 0.0 (-2.7-2.2)* | -0.1 (-2.6-2.1)* | -3.4 (-6.1-0.8) | -1.1 (-3.5-1.1) |
| BNP | 335.5 (175.8-776.0) | 454.0 (199.0-690.0) | 367.0 (218.0-944.8) | 297.0 (167.2-776.0)* | 408.0 (380.0-718.0) | 928.0 (416.8-1664.8) |
| Basophils | 0.0 (0.0-0.1) | 0.0 (0.0-0.1) | 0.0 (0.0-0.1) | 0.0 (0.0-0.1) | 0.0 (0.0-0.1) | 0.0 (0.0-0.1) |
| Bilirubin direct | 0.5 (0.2-2.4) | 0.2 (0.1-0.3) | 0.8 (0.2-3.6)* | 0.7 (0.2-3.3)* | 1.9 (1.5-2.8)* | 1.9 (1.0-6.6) |
| Bilirubin total | 1.0 (0.5-3.5) | 0.5 (0.4-0.8) | 1.5 (0.7-6.6)* | 1.3 (0.6-5.7)* | 1.1 (0.6-5.4) | 4.0 (2.0-9.6) |
| CRP | 70.0 (27.0-166.8) | 74.2 (29.5-201.1) | 76.3 (29.8-162.5) | 71.8 (29.8-165.9) | 39.0 (16.2-126.0) | 39.0 (11.4-104.8) |
| Calcium ionized | 4.6 (4.2-4.8) | 4.5 (4.2-4.8) | 4.6 (4.2-4.8) | 4.6 (4.2-4.8)* | 4.8 (4.5-5.0) | 4.7 (4.5-4.9) |
| Calcium non-ionized | 8.5 (8.1-9.1) | 8.6 (8.1-9.1) | 8.5 (8.0-9.1) | 8.5 (8.0-9.1) | 8.8 (7.7-9.6) | 8.9 (8.3-9.5) |
| Chloride | 104.0 (101.0-108.0) | 103.0 (99.0-106.0) | 105.0 (101.0-108.0)* | 104.0 (101.0-108.0) | 106.0 (102.0-109.0) | 104.0 (102.0-107.0) |
| Creatinine | 1.0 (0.7-1.4) | 0.9 (0.7-1.3) | 1.0 (0.7-1.6) | 1.0 (0.7-1.5) | 1.1 (1.0-1.5) | 1.0 (0.7-1.4) |
| Eosinophils | 0.1 (0.0-0.2) | 0.1 (0.0-0.3) | 0.1 (0.0-0.2) | 0.1 (0.0-0.2) | 0.1 (0.0-0.2) | 0.1 (0.0-0.2) |
| Glucose | 131.0 (110.0-157.0) | 121.0 (103.0-146.0) | 136.0 (114.0-161.0)* | 133.0 (111.0-159.0)* | 152.0 (126.0-171.0) | 145.0 (118.0-169.0) |
| Hematocrit | 26.0 (23.5-30.1) | 29.5 (25.5-33.6) | 25.2 (23.1-28.7)* | 25.3 (23.2-28.8)* | 22.4 (21.2-23.7)* | 24.6 (22.6-27.0)* |
| Hemoglobin | 8.6 (7.9-10.0) | 9.8 (8.4-11.3) | 8.4 (7.8-9.5)* | 8.4 (7.8-9.5)* | 7.6 (7.2-8.1)* | 8.1 (7.6-8.8)* |
| INR | 1.6 (1.4-2.0) | 1.3 (1.3-1.3) | 1.9 (1.5-2.0)* | 1.8 (1.5-2.0)* | | |
| Lactate | 1.5 (1.0-2.1) | 1.4 (1.0-2.1) | 1.5 (1.0-2.2) | 1.5 (1.0-2.2) | 2.2 (1.8-2.6) | 2.0 (1.5-2.6)* |
| Lymphocytes | 0.9 (0.5-1.3) | 1.0 (0.5-1.4) | 0.9 (0.5-1.3) | 0.9 (0.5-1.3) | 0.6 (0.3-1.2) | 0.5 (0.1-1.2) |
| Monocytes | 0.7 (0.4-1.0) | 0.6 (0.4-0.9) | 0.8 (0.5-1.2)* | 0.7 (0.4-1.1)* | 0.9 (0.3-1.1) | 1.0 (0.5-1.3) |
| Platelets | 172.0 (97.0-265.0) | 207.0 (156.0-281.0) | 155.0 (85.0-249.0)* | 160.0 (88.0-265.0)* | 71.0 (52.0-91.0)* | 78.0 (58.0-117.0)* |
| Potassium | 4.1 (3.8-4.4) | 4.1 (3.8-4.4) | 4.1 (3.8-4.4) | 4.1 (3.8-4.4) | 4.2 (4.0-4.4) | 4.2 (4.0-4.5) |
| Sodium | 138.0 (136.0-142.0) | 138.0 (135.0-140.0) | 139.0 (136.0-143.0) | 138.0 (136.0-142.0) | 139.0 (137.0-143.0) | 138.0 (136.0-142.0) |
| WBC | 10.5 (7.3-14.5) | 8.1 (5.9-11.5) | 11.4 (8.3-15.5)* | 11.0 (7.9-15.1)* | 10.8 (9.0-13.6) | 14.2 (9.7-18.5)* |
| **Medications** | | | | | | |
| Amiodarone | 61 (16.6%) | 14 (9.3%) | 35 (22.9%)* | 41 (22.0%)* | 1 (25.0%) | 6 (60.0%)* |
| Dexmedetomidine | 99 (26.9%) | 8 (5.3%) | 72 (47.1%)* | 82 (44.1%)* | 1 (25.0%) | 5 (50.0%)* |
| Digoxin | 9 (2.4%) | 3 (2.0%) | 6 (3.9%) | 5 (2.7%) | 0 (0.0%)* | 1 (10.0%) |
| Dopamine | 43 (11.7%) | 0 (0.0%) | 35 (22.9%)* | 43 (23.1%) | 1 (25.0%)* | 5 (50.0%)* |
| Epinephrine | 204 (55.4%) | 38 (25.3%) | 121 (79.1%)* | 151 (81.2%)* | 4 (100.0%)* | 10 (100.0%)* |
| Fentanyl | 46 (12.5%) | 15 (10.0%) | 23 (15.0%) | 30 (16.1%) | 3 (75.0%)* | 1 (10.0%) |
| Folic acid | 309 (84.0%) | 116 (77.3%) | 138 (90.2%)* | 165 (88.7%)* | 4 (100.0%) | 10 (100.0%) |
| Heparin sodium | 99 (26.9%) | 0 (0.0%) | 77 (50.3%)* | 99 (53.2%)* | 3 (75.0%)* | 9 (90.0%)* |
| Norepinephrine | 130 (35.3%) | 0 (0.0%) | 89 (58.2%)* | 130 (69.9%)* | 3 (75.0%)* | 7 (70.0%)* |
| Phenylephrine | 161 (43.8%) | 13 (8.7%) | 112 (73.2%)* | 137 (73.7%)* | 3 (75.0%)* | 8 (80.0%)* |
| Propofol | 72 (19.6%) | 0 (0.0%) | 61 (39.9%)* | 72 (38.7%)* | 2 (50.0%)* | 8 (80.0%)* |
| Vasopressin | 61 (16.6%) | 14 (9.3%) | 35 (22.9%)* | 41 (22.0%)* | 1 (25.0%) | 6 (60.0%)* |
| **Life-sustaining therapies** | | | | | | |
| CRRT | 4 (1.1%) | 0 (0.0%) | 4 (2.6%)* | 4 (2.2%) | 4 (100.0%)* | 2 (20.0%)* |
| MV | 153 (41.6%) | 0 (0.0%) | 153 (100.0%)* | 122 (65.6%)* | 4 (100.0%)* | 8 (80.0%)* |
| VP | 186 (50.5%) | 0 (0.0%) | 122 (79.7%)* | 186 (100.0%)* | 4 (100.0%)* | 9 (90.0%)* |
| **Outcomes** | | | | | | |
| Deceased | 10 (2.7%) | 0 (0.0%) | 8 (5.2%)* | 9 (4.8%)* | 2 (50.0%)* | 10 (100.0%)* |

Statistics are presented as median (interquartile range) for continuous variables, and number (percentage) for categorical variables. P-values for continuous variables are based on pairwise Wilcoxon rank sum test. P-values for categorical variables are based on pairwise Pearson's chi-squared test for proportions.
*p-value < 0.05 compared to stable admissions.

**Supplementary Table S6. Classification performance metrics for all models and prediction tasks in validation cohorts.**

| Task/Validation cohort | AUROC (95% CI) | AUPRC (95% CI) | Sensitivity (95% CI) | Specificity (95% CI) | PPV (95% CI) | NPV (95% CI) |
|---|---|---|---|---|---|---|
| **Discharge** | | | | | | |
|   **Development** | | | | | | |
|     CatBoost | 0.84 (0.83-0.85) | 0.03 (0.02-0.03) | 0.81 (0.76-0.86) | 0.74 (0.69-0.78) | 0.01 (0.01-0.02) | 1.00 (1.00-1.00) |
|     GRU | 0.81 (0.80-0.82) | 0.02 (0.02-0.02) | 0.83 (0.79-0.86) | 0.67 (0.62-0.70) | 0.01 (0.01-0.01) | 1.00 (1.00-1.00) |
|     Transformer | 0.85 (0.84-0.85) | 0.03 (0.02-0.03) | 0.81 (0.79-0.83) | 0.74 (0.72-0.75) | 0.01 (0.01-0.02) | 1.00 (1.00-1.00) |
|     APRICOT-T | 0.86 (0.85-0.87) | 0.03 (0.03-0.03) | 0.84 (0.82-0.87) | 0.74 (0.72-0.76) | 0.02 (0.01-0.02) | 1.00 (1.00-1.00) |
|     APRICOT-M | 0.87 (0.86-0.87) [a,b,c] | 0.03 (0.03-0.03) [b] | 0.84 (0.81-0.88) | 0.74 (0.70-0.77) [b] | 0.02 (0.01-0.02) [a,b,c] | 1.00 (1.00-1.00) |
|   **External** | | | | | | |
|     CatBoost | 0.85 (0.84-0.85) | 0.03 (0.03-0.03) | 0.82 (0.80-0.84) | 0.72 (0.70-0.73) | 0.01 (0.01-0.01) | 1.00 (1.00-1.00) |
|     GRU | 0.82 (0.82-0.83) | 0.02 (0.02-0.02) | 0.84 (0.80-0.86) | 0.66 (0.64-0.70) | 0.01 (0.01-0.01) | 1.00 (1.00-1.00) |
|     Transformer | 0.84 (0.83-0.84) | 0.03 (0.03-0.03) | 0.82 (0.79-0.83) | 0.70 (0.68-0.73) | 0.01 (0.01-0.01) | 1.00 (1.00-1.00) |
|     APRICOT-T | 0.87 (0.87-0.87) | 0.03 (0.03-0.04) | 0.86 (0.82-0.87) | 0.74 (0.73-0.77) | 0.02 (0.01-0.02) | 1.00 (1.00-1.00) |
|     APRICOT-M | 0.87 (0.87-0.88) [a,b,c] | 0.04 (0.03-0.04) [a,b,c,d] | 0.86 (0.84-0.87) [a,c] | 0.74 (0.73-0.75) [a,b,c] | 0.02 (0.02-0.02) [a,b,c] | 1.00 (1.00-1.00) |
|   **Temporal** | | | | | | |
|     CatBoost | 0.67 (0.67-0.68) | 0.02 (0.02-0.03) | 0.55 (0.50-0.58) | 0.71 (0.68-0.75) | 0.02 (0.02-0.02) | 0.99 (0.99-0.99) |
|     GRU | 0.72 (0.71-0.72) | 0.03 (0.02-0.03) | 0.75 (0.68-0.80) | 0.58 (0.52-0.64) | 0.02 (0.02-0.02) | 1.00 (0.99-1.00) |
|     Transformer | 0.78 (0.77-0.79) | 0.04 (0.04-0.05) | 0.73 (0.70-0.76) | 0.68 (0.65-0.71) | 0.03 (0.02-0.03) | 1.00 (1.00-1.00) |
|     APRICOT-T | 0.80 (0.80-0.81) | 0.05 (0.05-0.05) | 0.78 (0.74-0.81) | 0.69 (0.66-0.72) | 0.03 (0.03-0.03) | 1.00 (1.00-1.00) |
|     APRICOT-M | 0.80 (0.80-0.81) [a,b,c] | 0.05 (0.05-0.05) [a,b,c] | 0.76 (0.72-0.82) [a] | 0.70 (0.64-0.74) [b] | 0.03 (0.03-0.03) [a,b] | 1.00 (1.00-1.00) [a] |
|   **Prospective** | | | | | | |
|     CatBoost | 0.79 (0.74-0.83) | 0.03 (0.01-0.08) | 0.84 (0.73-0.93) | 0.65 (0.58-0.73) | 0.01 (0.01-0.01) | 1.00 (1.00-1.00) |
|     GRU | 0.74 (0.69-0.78) | 0.01 (0.01-0.03) | 0.91 (0.68-1.00) | 0.52 (0.46-0.74) | 0.01 (0.01-0.01) | 1.00 (1.00-1.00) |
|     Transformer | 0.82 (0.78-0.86) | 0.03 (0.01-0.06) | 0.84 (0.66-0.97) | 0.69 (0.58-0.86) | 0.01 (0.01-0.02) | 1.00 (1.00-1.00) |
|     APRICOT-T | 0.81 (0.76-0.86) | 0.04 (0.02-0.08) | 0.79 (0.65-0.93) | 0.71 (0.51-0.86) | 0.01 (0.01-0.02) | 1.00 (1.00-1.00) |
|     APRICOT-M | 0.84 (0.79-0.88) [b] | 0.03 (0.02-0.06) | 0.81 (0.71-0.93) | 0.81 (0.65-0.84) [a,b] | 0.01 (0.01-0.02) | 1.00 (1.00-1.00) |
| **Stable** | | | | | | |
|   **Development** | | | | | | |
|     CatBoost | 0.87 (0.86-0.87) | 0.90 (0.90-0.90) | 0.87 (0.87-0.88) | 0.75 (0.75-0.76) | 0.86 (0.86-0.86) | 0.77 (0.76-0.78) |
|     GRU | 0.93 (0.93-0.93) | 0.94 (0.94-0.94) | 0.96 (0.95-0.96) | 0.80 (0.80-0.81) | 0.90 (0.90-0.90) | 0.91 (0.91-0.92) |

|  | | | | | | | |
|---|---|---|---|---|---|---|---|
| | Transformer | 0.93 (0.93-0.94) | 0.95 (0.94-0.95) | 0.96 (0.95-0.96) | 0.81 (0.80-0.81) | 0.90 (0.90-0.90) | 0.92 (0.91-0.92) |
| | APRICOT-T | 0.95 (0.94-0.95) | 0.95 (0.95-0.95) | 0.97 (0.96-0.97) | 0.83 (0.83-0.83) | 0.91 (0.91-0.91) | 0.93 (0.93-0.94) |
| | APRICOT-M | 0.94 (0.94-0.95) [a,b,c,d] | 0.95 (0.95-0.95) [a,b] | 0.96 (0.95-0.96) [a,d] | 0.84 (0.83-0.84) [a,b,c,d] | 0.91 (0.91-0.91) [a,b,c] | 0.91 (0.91-0.92) [a,c,d] |
| **External** | | | | | | | |
| | CatBoost | 0.84 (0.84-0.84) | 0.87 (0.87-0.87) | 0.82 (0.82-0.83) | 0.75 (0.74-0.75) | 0.84 (0.84-0.84) | 0.72 (0.72-0.73) |
| | GRU | 0.90 (0.90-0.91) | 0.91 (0.91-0.91) | 0.92 (0.92-0.92) | 0.78 (0.77-0.78) | 0.87 (0.87-0.87) | 0.86 (0.85-0.86) |
| | Transformer | 0.91 (0.91-0.91) | 0.92 (0.92-0.92) | 0.92 (0.91-0.92) | 0.78 (0.78-0.79) | 0.87 (0.87-0.87) | 0.85 (0.85-0.86) |
| | APRICOT-T | 0.95 (0.94-0.95) | 0.95 (0.95-0.95) | 0.95 (0.95-0.95) | 0.86 (0.85-0.86) | 0.91 (0.91-0.91) | 0.91 (0.91-0.91) |
| | APRICOT-M | 0.95 (0.94-0.95) [a,b,c] | 0.95 (0.95-0.95) [a,b,c] | 0.95 (0.95-0.95) [a,b,c] | 0.86 (0.85-0.86) [a,b,c] | 0.91 (0.91-0.91) [a,b,c] | 0.91 (0.91-0.91) [a,b,c] |
| **Temporal** | | | | | | | |
| | CatBoost | 0.84 (0.84-0.84) | 0.92 (0.92-0.92) | 0.87 (0.86-0.87) | 0.70 (0.69-0.71) | 0.90 (0.89-0.90) | 0.64 (0.64-0.65) |
| | GRU | 0.93 (0.93-0.93) | 0.96 (0.96-0.96) | 0.96 (0.95-0.96) | 0.82 (0.82-0.83) | 0.94 (0.94-0.94) | 0.86 (0.85-0.87) |
| | Transformer | 0.94 (0.93-0.94) | 0.96 (0.96-0.97) | 0.95 (0.95-0.96) | 0.82 (0.82-0.83) | 0.94 (0.94-0.94) | 0.86 (0.85-0.87) |
| | APRICOT-T | 0.95 (0.95-0.95) | 0.97 (0.97-0.97) | 0.97 (0.96-0.97) | 0.85 (0.85-0.85) | 0.95 (0.95-0.95) | 0.90 (0.89-0.90) |
| | APRICOT-M | 0.95 (0.95-0.95) [a,b,c] | 0.97 (0.97-0.97) [a,b,c] | 0.97 (0.96-0.97) [a,b,c] | 0.85 (0.85-0.86) [a,b,c] | 0.95 (0.95-0.95) [a,b,c] | 0.90 (0.88-0.91) [a,b,c] |
| **Prospective** | | | | | | | |
| | CatBoost | 0.93 (0.93-0.94) | 0.96 (0.96-0.97) | 0.93 (0.92-0.94) | 0.85 (0.84-0.87) | 0.95 (0.94-0.95) | 0.81 (0.78-0.84) |
| | GRU | 0.91 (0.91-0.92) | 0.95 (0.95-0.96) | 0.95 (0.94-0.96) | 0.77 (0.76-0.79) | 0.92 (0.92-0.92) | 0.86 (0.83-0.88) |
| | Transformer | 0.90 (0.90-0.91) | 0.95 (0.94-0.95) | 0.92 (0.90-0.96) | 0.77 (0.74-0.79) | 0.92 (0.91-0.92) | 0.79 (0.75-0.86) |
| | APRICOT-T | 0.96 (0.95-0.96) | 0.97 (0.97-0.98) | 0.97 (0.95-0.97) | 0.88 (0.87-0.89) | 0.96 (0.95-0.96) | 0.90 (0.87-0.92) |
| | APRICOT-M | 0.95 (0.95-0.96) [a,b,c,d] | 0.98 (0.97-0.98) [a,b,c,d] | 0.94 (0.93-0.95) [d] | 0.88 (0.87-0.90) [a,b,c] | 0.96 (0.95-0.96) [a,b,c] | 0.84 (0.81-0.86) [d] |
| **Unstable** | | | | | | | |
| **Development** | | | | | | | |
| | CatBoost | 0.87 (0.87-0.87) | 0.80 (0.80-0.80) | 0.76 (0.76-0.77) | 0.87 (0.86-0.87) | 0.76 (0.75-0.77) | 0.87 (0.87-0.87) |
| | GRU | 0.94 (0.94-0.94) | 0.93 (0.92-0.93) | 0.81 (0.81-0.82) | 0.96 (0.95-0.96) | 0.91 (0.90-0.92) | 0.90 (0.90-0.91) |
| | Transformer | 0.94 (0.94-0.94) | 0.93 (0.93-0.93) | 0.82 (0.81-0.82) | 0.96 (0.95-0.96) | 0.91 (0.90-0.92) | 0.91 (0.90-0.91) |
| | APRICOT-T | 0.95 (0.95-0.95) | 0.94 (0.94-0.94) | 0.84 (0.84-0.85) | 0.96 (0.96-0.97) | 0.92 (0.91-0.93) | 0.92 (0.92-0.92) |
| | APRICOT-M | 0.95 (0.95-0.95) [a,b,c] | 0.94 (0.94-0.94) [a,b,c] | 0.85 (0.84-0.85) [a,b,c,d] | 0.95 (0.95-0.96) [a,b,c,d] | 0.91 (0.90-0.92) [a] | 0.92 (0.92-0.92) [a,b,c] |
| **External** | | | | | | | |
| | CatBoost | 0.85 (0.85-0.85) | 0.76 (0.76-0.76) | 0.76 (0.75-0.76) | 0.82 (0.82-0.83) | 0.72 (0.71-0.72) | 0.85 (0.85-0.85) |
| | GRU | 0.91 (0.91-0.91) | 0.89 (0.89-0.89) | 0.79 (0.79-0.80) | 0.92 (0.91-0.92) | 0.85 (0.84-0.85) | 0.88 (0.88-0.88) |
| | Transformer | 0.91 (0.91-0.91) | 0.90 (0.90-0.90) | 0.79 (0.79-0.79) | 0.92 (0.91-0.92) | 0.85 (0.85-0.85) | 0.88 (0.88-0.88) |
| | APRICOT-T | 0.95 (0.95-0.95) | 0.94 (0.94-0.94) | 0.87 (0.87-0.87) | 0.95 (0.95-0.95) | 0.91 (0.91-0.91) | 0.92 (0.92-0.92) |
| | APRICOT-M | 0.95 (0.95-0.95) a,b,c | 0.94 (0.94-0.94) a,b,c | 0.87 (0.87-0.87) a,b,c | 0.95 (0.94-0.95) a,b,c | 0.91 (0.90-0.91) a,b,c | 0.92 (0.92-0.92) a,b,c |
| **Temporal** | | | | | | | |
| | CatBoost | 0.86 (0.86-0.86) | 0.69 (0.69-0.69) | 0.72 (0.72-0.74) | 0.87 (0.86-0.88) | 0.64 (0.62-0.64) | 0.91 (0.91-0.91) |
| | GRU | 0.95 (0.95-0.96) | 0.92 (0.92-0.92) | 0.86 (0.86-0.87) | 0.95 (0.95-0.96) | 0.85 (0.84-0.86) | 0.96 (0.96-0.96) |
| | Transformer | 0.96 (0.96-0.96) | 0.93 (0.93-0.93) | 0.87 (0.86-0.87) | 0.95 (0.95-0.96) | 0.85 (0.83-0.86) | 0.96 (0.96-0.96) |
| | APRICOT-T | 0.97 (0.97-0.97) | 0.95 (0.94-0.95) | 0.89 (0.88-0.89) | 0.97 (0.96-0.97) | 0.89 (0.88-0.90) | 0.96 (0.96-0.97) |
| | APRICOT-M | 0.97 (0.97-0.97) a,b,c | 0.95 (0.95-0.95) a,b,c | 0.89 (0.89-0.89) a,b,c | 0.97 (0.96-0.97) a,b,c | 0.89 (0.88-0.90) a,b,c | 0.97 (0.96-0.97) a,b,c,d |
| **Prospective** | | | | | | | |
| | CatBoost | 0.94 (0.94-0.95) | 0.91 (0.90-0.91) | 0.86 (0.85-0.88) | 0.92 (0.92-0.93) | 0.80 (0.78-0.82) | 0.95 (0.95-0.95) |
| | GRU | 0.92 (0.92-0.93) | 0.89 (0.88-0.89) | 0.79 (0.77-0.81) | 0.95 (0.93-0.96) | 0.84 (0.81-0.88) | 0.93 (0.92-0.93) |
| | Transformer | 0.91 (0.90-0.92) | 0.86 (0.86-0.87) | 0.77 (0.74-0.80) | 0.93 (0.90-0.96) | 0.79 (0.74-0.86) | 0.92 (0.91-0.93) |
| | APRICOT-T | 0.96 (0.96-0.96) | 0.94 (0.93-0.94) | 0.88 (0.87-0.89) | 0.97 (0.96-0.97) | 0.90 (0.87-0.92) | 0.96 (0.95-0.96) |
| | APRICOT-M | 0.96 (0.96-0.96) a,b,c | 0.92 (0.91-0.92) a,b,c,d | 0.90 (0.88-0.91) a,b,c,d | 0.93 (0.92-0.95) d | 0.82 (0.80-0.85) d | 0.96 (0.96-0.97) a,b,c |
| **Deceased** | | | | | | | |
| **Development** | | | | | | | |
| | CatBoost | 0.85 (0.84-0.86) | 0.04 (0.03-0.05) | 0.74 (0.69-0.80) | 0.81 (0.76-0.86) | 0.01 (0.01-0.01) | 1.00 (1.00-1.00) |
| | GRU | 0.86 (0.85-0.87) | 0.03 (0.02-0.03) | 0.80 (0.76-0.83) | 0.76 (0.75-0.81) | 0.01 (0.01-0.01) | 1.00 (1.00-1.00) |
| | Transformer | 0.90 (0.89-0.91) | 0.05 (0.04-0.06) | 0.82 (0.76-0.86) | 0.84 (0.79-0.89) | 0.01 (0.01-0.02) | 1.00 (1.00-1.00) |
| | APRICOT-T | 0.93 (0.93-0.94) | 0.12 (0.10-0.14) | 0.87 (0.82-0.90) | 0.86 (0.83-0.90) | 0.01 (0.01-0.02) | 1.00 (1.00-1.00) |
| | APRICOT-M | 0.93 (0.92-0.94) a,b,c | 0.14 (0.12-0.16) a,b,c | 0.86 (0.82-0.88) a,b | 0.88 (0.86-0.91) a,b | 0.02 (0.01-0.02) a,b,c,d | 1.00 (1.00-1.00) |
| **External** | | | | | | | |
| | CatBoost | 0.84 (0.83-0.84) | 0.02 (0.02-0.03) | 0.77 (0.72-0.79) | 0.76 (0.74-0.80) | 0.01 (0.01-0.01) | 1.00 (1.00-1.00) |
| | GRU | 0.79 (0.78-0.80) | 0.02 (0.01-0.02) | 0.78 (0.76-0.80) | 0.64 (0.61-0.66) | 0.01 (0.01-0.01) | 1.00 (1.00-1.00) |
| | Transformer | 0.87 (0.87-0.88) | 0.02 (0.01-0.02) | 0.85 (0.83-0.87) | 0.77 (0.74-0.78) | 0.01 (0.01-0.01) | 1.00 (1.00-1.00) |
| | APRICOT-T | 0.95 (0.94-0.95) | 0.17 (0.16-0.19) | 0.86 (0.83-0.89) | 0.88 (0.86-0.91) | 0.02 (0.02-0.03) | 1.00 (1.00-1.00) |
| | APRICOT-M | 0.95 (0.94-0.95) a,b,c | 0.18 (0.16-0.19) a,b,c | 0.86 (0.84-0.88) a,b | 0.89 (0.87-0.91) a,b,c | 0.02 (0.02-0.03) a,b,c | 1.00 (1.00-1.00) |
| **Temporal** | | | | | | | |
| | CatBoost | 0.87 (0.86-0.88) | 0.04 (0.03-0.05) | 0.77 (0.71-0.82) | 0.83 (0.78-0.88) | 0.01 (0.01-0.01) | 1.00 (1.00-1.00) |
| | GRU | 0.92 (0.91-0.93) | 0.09 (0.07-0.11) | 0.82 (0.76-0.87) | 0.85 (0.82-0.89) | 0.01 (0.01-0.01) | 1.00 (1.00-1.00) |
| | Transformer | 0.95 (0.95-0.96) | 0.25 (0.21-0.28) | 0.90 (0.87-0.92) | 0.88 (0.87-0.91) | 0.01 (0.01-0.02) | 1.00 (1.00-1.00) |
| | APRICOT-T | 0.98 (0.97-0.98) | 0.36 (0.33-0.39) | 0.92 (0.90-0.94) | 0.92 (0.90-0.94) | 0.02 (0.02-0.03) | 1.00 (1.00-1.00) |
| | APRICOT-M | 0.98 (0.97-0.98) a,b,c | 0.42 (0.39-0.45) a,b,c,d | 0.91 (0.88-0.94) a,b | 0.94 (0.92-0.96) a,b,c | 0.03 (0.02-0.04) a,b,c | 1.00 (1.00-1.00) |

| | | | | | | |
|---|---|---|---|---|---|---|
| **Prospective** | | | | | | |
| CatBoost | 0.94 (0.90-0.98) | 0.01 (0.00-0.03) | 0.95 (0.83-1.00) | 0.87 (0.76-0.96) | 0.01 (0.00-0.01) | 1.00 (1.00-1.00) |
| GRU | 0.88 (0.79-0.96) | 0.00 (0.00-0.01) | 0.91 (0.71-1.00) | 0.80 (0.60-0.96) | 0.00 (0.00-0.01) | 1.00 (1.00-1.00) |
| Transformer | 0.88 (0.77-0.96) | 0.04 (0.00-0.11) | 0.92 (0.71-1.00) | 0.77 (0.64-0.99) | 0.01 (0.00-0.08) | 1.00 (1.00-1.00) |
| APRICOT-T | 0.98 (0.95-1.00) | 0.04 (0.01-0.13) | 1.00 (1.00-1.00) | 0.90 (0.90-1.00) | 0.01 (0.00-0.06) | 1.00 (1.00-1.00) |
| APRICOT-M | 0.98 (0.96-1.00) [b,c] | 0.05 (0.01-0.15) | 1.00 (1.00-1.00) | 0.92 (0.91-0.99) | 0.01 (0.00-0.07) | 1.00 (1.00-1.00) |
| **Unstable-Stable** | | | | | | |
| **Development** | | | | | | |
| CatBoost | 0.63 (0.62-0.64) | 0.04 (0.04-0.04) | 0.53 (0.48-0.60) | 0.66 (0.59-0.71) | 0.03 (0.03-0.03) | 0.99 (0.99-0.99) |
| GRU | 0.69 (0.68-0.69) | 0.07 (0.07-0.08) | 0.58 (0.54-0.64) | 0.68 (0.62-0.72) | 0.04 (0.03-0.04) | 0.99 (0.99-0.99) |
| Transformer | 0.73 (0.72-0.73) | 0.08 (0.07-0.08) | 0.64 (0.59-0.68) | 0.69 (0.65-0.74) | 0.04 (0.04-0.04) | 0.99 (0.99-0.99) |
| APRICOT-T | 0.75 (0.75-0.76) | 0.10 (0.10-0.11) | 0.68 (0.62-0.74) | 0.68 (0.63-0.75) | 0.04 (0.04-0.05) | 0.99 (0.99-0.99) |
| APRICOT-M | 0.74 (0.73-0.74) [a,b,c,d] | 0.10 (0.09-0.10) [a,b,c] | 0.65 (0.63-0.67) [a,b] | 0.70 (0.68-0.72) | 0.04 (0.04-0.04) [a] | 0.99 (0.99-0.99) |
| **External** | | | | | | |
| CatBoost | 0.65 (0.64-0.65) | 0.07 (0.06-0.07) | 0.46 (0.40-0.51) | 0.74 (0.70-0.80) | 0.04 (0.04-0.04) | 0.98 (0.98-0.98) |
| GRU | 0.68 (0.68-0.69) | 0.06 (0.05-0.06) | 0.58 (0.53-0.61) | 0.68 (0.65-0.73) | 0.04 (0.04-0.04) | 0.99 (0.99-0.99) |
| Transformer | 0.73 (0.72-0.73) | 0.10 (0.10-0.11) | 0.60 (0.57-0.63) | 0.73 (0.71-0.76) | 0.05 (0.05-0.05) | 0.99 (0.99-0.99) |
| APRICOT-T | 0.80 (0.79-0.80) | 0.14 (0.14-0.14) | 0.68 (0.66-0.69) | 0.79 (0.77-0.80) | 0.07 (0.06-0.07) | 0.99 (0.99-0.99) |
| APRICOT-M | 0.79 (0.79-0.79) [a,b,c,d] | 0.14 (0.13-0.14) [a,b,c] | 0.65 (0.63-0.67) [a,b,c,d] | 0.80 (0.77-0.82) [a,b,c] | 0.07 (0.06-0.07) [a,b,c] | 0.99 (0.99-0.99) [a] |
| **Temporal** | | | | | | |
| CatBoost | 0.61 (0.61-0.62) | 0.04 (0.04-0.04) | 0.58 (0.46-0.64) | 0.58 (0.53-0.70) | 0.02 (0.02-0.03) | 0.99 (0.99-0.99) |
| GRU | 0.82 (0.81-0.82) | 0.23 (0.22-0.24) | 0.65 (0.64-0.68) | 0.84 (0.82-0.85) | 0.07 (0.06-0.07) | 0.99 (0.99-0.99) |
| Transformer | 0.87 (0.87-0.88) | 0.31 (0.30-0.32) | 0.76 (0.72-0.78) | 0.83 (0.81-0.86) | 0.07 (0.07-0.09) | 0.99 (0.99-1.00) |
| APRICOT-T | 0.89 (0.89-0.90) | 0.33 (0.32-0.34) | 0.76 (0.74-0.77) | 0.86 (0.85-0.88) | 0.09 (0.09-0.10) | 1.00 (0.99-1.00) |
| APRICOT-M | 0.91 (0.91-0.92) [a,b,c,d] | 0.44 (0.42-0.45) [a,b,c,d] | 0.79 (0.77-0.81) [a,b,d] | 0.88 (0.87-0.89) [a,b,c,d] | 0.11 (0.10-0.12) [a,b,c,d] | 1.00 (1.00-1.00) [a,b,c] |
| **Prospective** | | | | | | |
| CatBoost | 0.85 (0.82-0.88) | 0.17 (0.14-0.21) | 0.75 (0.67-0.83) | 0.86 (0.78-0.90) | 0.10 (0.06-0.13) | 0.99 (0.99-1.00) |
| GRU | 0.79 (0.76-0.82) | 0.18 (0.14-0.22) | 0.66 (0.57-0.74) | 0.81 (0.75-0.86) | 0.06 (0.05-0.08) | 0.99 (0.99-0.99) |
| Transformer | 0.81 (0.78-0.84) | 0.19 (0.15-0.24) | 0.69 (0.58-0.77) | 0.81 (0.68-0.90) | 0.07 (0.05-0.10) | 0.99 (0.99-0.99) |
| APRICOT-T | 0.85 (0.83-0.88) | 0.24 (0.20-0.28) | 0.77 (0.70-0.82) | 0.83 (0.79-0.89) | 0.08 (0.07-0.11) | 0.99 (0.99-1.00) |
| APRICOT-M | 0.84 (0.80-0.87) [b] | 0.20 (0.16-0.25) | 0.71 (0.62-0.81) | 0.85 (0.74-0.90) | 0.08 (0.05-0.12) | 0.99 (0.99-1.00) |
| **Stable-Unstable** | | | | | | |
| **Development** | | | | | | |

|  |  |  |  |  |  |  |  |
|---|---|---|---|---|---|---|---|
|  | CatBoost | 0.65 (0.64-0.66) | 0.02 (0.02-0.02) | 0.59 (0.44-0.67) | 0.63 (0.56-0.78) | 0.01 (0.01-0.01) | 1.00 (1.00-1.00) |
|  | GRU | 0.79 (0.78-0.79) | 0.03 (0.02-0.03) | 0.71 (0.63-0.78) | 0.72 (0.65-0.79) | 0.01 (0.01-0.02) | 1.00 (1.00-1.00) |
|  | Transformer | 0.82 (0.82-0.83) | 0.11 (0.10-0.13) | 0.70 (0.65-0.75) | 0.77 (0.73-0.83) | 0.02 (0.02-0.02) | 1.00 (1.00-1.00) |
|  | APRICOT-T | 0.84 (0.83-0.84) | 0.12 (0.11-0.14) | 0.70 (0.66-0.77) | 0.80 (0.71-0.82) | 0.02 (0.02-0.02) | 1.00 (1.00-1.00) |
|  | APRICOT-M | 0.84 (0.83-0.85) [a,b,c] | 0.12 (0.11-0.14) [a,b] | 0.75 (0.68-0.81) [a] | 0.75 (0.69-0.81) | 0.02 (0.01-0.02) [a,b] | 1.00 (1.00-1.00) |
| **External** |  |  |  |  |  |  |  |
|  | CatBoost | 0.69 (0.69-0.70) | 0.10 (0.09-0.10) | 0.46 (0.40-0.51) | 0.81 (0.76-0.87) | 0.02 (0.02-0.02) | 0.99 (0.99-0.99) |
|  | GRU | 0.74 (0.73-0.74) | 0.02 (0.02-0.02) | 0.63 (0.60-0.67) | 0.72 (0.69-0.75) | 0.02 (0.02-0.02) | 1.00 (1.00-1.00) |
|  | Transformer | 0.78 (0.78-0.79) | 0.13 (0.13-0.14) | 0.68 (0.65-0.70) | 0.73 (0.72-0.77) | 0.02 (0.02-0.02) | 1.00 (1.00-1.00) |
|  | APRICOT-T | 0.82 (0.81-0.82) | 0.22 (0.21-0.23) | 0.68 (0.65-0.70) | 0.79 (0.78-0.81) | 0.03 (0.03-0.03) | 1.00 (1.00-1.00) |
|  | APRICOT-M | 0.82 (0.81-0.82) [a,b,c] | 0.22 (0.21-0.22) [a,b,c] | 0.66 (0.64-0.69) [a] | 0.81 (0.78-0.83) [b,c] | 0.03 (0.03-0.03) [a,b,c] | 1.00 (1.00-1.00) [a] |
| **Temporal** |  |  |  |  |  |  |  |
|  | CatBoost | 0.63 (0.62-0.63) | 0.02 (0.02-0.02) | 0.60 (0.51-0.67) | 0.59 (0.52-0.67) | 0.01 (0.01-0.02) | 0.99 (0.99-0.99) |
|  | GRU | 0.73 (0.72-0.73) | 0.06 (0.06-0.07) | 0.60 (0.52-0.67) | 0.72 (0.65-0.78) | 0.02 (0.02-0.03) | 0.99 (0.99-0.99) |
|  | Transformer | 0.75 (0.74-0.75) | 0.05 (0.05-0.05) | 0.60 (0.55-0.68) | 0.74 (0.67-0.79) | 0.02 (0.02-0.03) | 0.99 (0.99-1.00) |
|  | APRICOT-T | 0.76 (0.76-0.77) | 0.06 (0.06-0.07) | 0.63 (0.58-0.68) | 0.74 (0.70-0.79) | 0.03 (0.02-0.03) | 0.99 (0.99-1.00) |
|  | APRICOT-M | 0.78 (0.77-0.78) [a,b,c,d] | 0.08 (0.07-0.09) [a,b,c,d] | 0.67 (0.64-0.72) | 0.72 (0.68-0.76) [a] | 0.02 (0.02-0.03) [a,d] | 1.00 (0.99-1.00) [a,b,c,d] |
| **Prospective** |  |  |  |  |  |  |  |
|  | CatBoost | 0.57 (0.53-0.61) | 0.02 (0.01-0.03) | 0.57 (0.24-0.85) | 0.57 (0.27-0.87) | 0.02 (0.01-0.02) | 0.99 (0.99-0.99) |
|  | GRU | 0.69 (0.66-0.73) | 0.04 (0.02-0.07) | 0.69 (0.50-0.81) | 0.60 (0.49-0.80) | 0.02 (0.02-0.03) | 0.99 (0.99-1.00) |
|  | Transformer | 0.70 (0.67-0.74) | 0.03 (0.02-0.04) | 0.72 (0.61-0.79) | 0.63 (0.59-0.75) | 0.02 (0.02-0.03) | 0.99 (0.99-1.00) |
|  | APRICOT-T | 0.72 (0.69-0.75) | 0.04 (0.03-0.07) | 0.67 (0.51-0.84) | 0.67 (0.51-0.83) | 0.02 (0.02-0.04) | 0.99 (0.99-1.00) |
|  | APRICOT-M | 0.72 (0.69-0.75) [a] | 0.04 (0.03-0.07) | 0.73 (0.65-0.83) | 0.62 (0.53-0.67) | 0.02 (0.02-0.03) | 0.99 (0.99-1.00) |
| **MV** |  |  |  |  |  |  |  |
| **Development** |  |  |  |  |  |  |  |
|  | CatBoost | 0.68 (0.67-0.69) | 0.02 (0.01-0.02) | 0.67 (0.54-0.74) | 0.60 (0.53-0.73) | 0.01 (0.01-0.01) | 1.00 (1.00-1.00) |
|  | GRU | 0.81 (0.80-0.82) | 0.03 (0.03-0.03) | 0.70 (0.65-0.74) | 0.77 (0.74-0.81) | 0.02 (0.01-0.02) | 1.00 (1.00-1.00) |
|  | Transformer | 0.84 (0.83-0.85) | 0.13 (0.11-0.15) | 0.76 (0.70-0.78) | 0.75 (0.74-0.81) | 0.02 (0.01-0.02) | 1.00 (1.00-1.00) |
|  | APRICOT-T | 0.85 (0.84-0.86) | 0.13 (0.12-0.15) | 0.74 (0.68-0.80) | 0.78 (0.71-0.83) | 0.02 (0.01-0.02) | 1.00 (1.00-1.00) |
|  | APRICOT-M | 0.85 (0.84-0.86) [a,b] | 0.14 (0.12-0.15) [a,b] | 0.73 (0.68-0.78) | 0.79 (0.74-0.83) [a] | 0.02 (0.02-0.02) [a] | 1.00 (1.00-1.00) |
| **External** |  |  |  |  |  |  |  |
|  | CatBoost | 0.74 (0.74-0.75) | 0.14 (0.13-0.15) | 0.64 (0.60-0.67) | 0.71 (0.67-0.74) | 0.01 (0.01-0.01) | 1.00 (1.00-1.00) |
|  | GRU | 0.77 (0.77-0.77) | 0.02 (0.02-0.02) | 0.65 (0.64-0.67) | 0.76 (0.74-0.76) | 0.02 (0.02-0.02) | 1.00 (1.00-1.00) |
|  | Transformer | 0.81 (0.81-0.82) | 0.19 (0.18-0.20) | 0.68 (0.66-0.70) | 0.80 (0.78-0.82) | 0.02 (0.02-0.02) | 1.00 (1.00-1.00) |
| | APRICOT-T | 0.82 (0.82-0.83) | 0.27 (0.26-0.28) | 0.67 (0.64-0.69) | 0.83 (0.82-0.86) | 0.02 (0.02-0.03) | 1.00 (1.00-1.00) |
| | APRICOT-M | 0.82 (0.82-0.83) [a,b,c] | 0.23 (0.23-0.24) [a,b,c,d] | 0.65 (0.64-0.69) | 0.85 (0.82-0.87) [a,b,c] | 0.03 (0.02-0.03) [a,b,c,d] | 1.00 (1.00-1.00) |
| **Temporal** | | | | | | | |
| | CatBoost | 0.73 (0.72-0.74) | 0.02 (0.02-0.02) | 0.72 (0.66-0.77) | 0.66 (0.60-0.72) | 0.01 (0.01-0.01) | 1.00 (1.00-1.00) |
| | GRU | 0.82 (0.81-0.83) | 0.11 (0.09-0.12) | 0.69 (0.67-0.73) | 0.81 (0.78-0.83) | 0.02 (0.02-0.02) | 1.00 (1.00-1.00) |
| | Transformer | 0.86 (0.85-0.86) | 0.12 (0.11-0.14) | 0.77 (0.71-0.80) | 0.80 (0.77-0.85) | 0.02 (0.02-0.03) | 1.00 (1.00-1.00) |
| | APRICOT-T | 0.87 (0.86-0.88) | 0.11 (0.10-0.12) | 0.76 (0.73-0.81) | 0.83 (0.79-0.86) | 0.03 (0.02-0.03) | 1.00 (1.00-1.00) |
| | APRICOT-M | 0.87 (0.87-0.88) [a,b,c] | 0.12 (0.11-0.14) [a] | 0.79 (0.75-0.83) [a,b] | 0.81 (0.77-0.85) [a] | 0.02 (0.02-0.03) [a,d] | 1.00 (1.00-1.00) |
| **Prospective** | | | | | | | |
| | CatBoost | 0.71 (0.66-0.74) | 0.03 (0.02-0.04) | 0.78 (0.69-0.89) | 0.61 (0.48-0.65) | 0.02 (0.01-0.02) | 1.00 (1.00-1.00) |
| | GRU | 0.78 (0.73-0.83) | 0.06 (0.03-0.08) | 0.67 (0.52-0.83) | 0.78 (0.60-0.90) | 0.03 (0.02-0.04) | 1.00 (1.00-1.00) |
| | Transformer | 0.80 (0.76-0.84) | 0.06 (0.03-0.09) | 0.70 (0.59-0.86) | 0.79 (0.63-0.83) | 0.03 (0.02-0.04) | 1.00 (1.00-1.00) |
| | APRICOT-T | 0.80 (0.75-0.83) | 0.06 (0.04-0.10) | 0.70 (0.61-0.83) | 0.79 (0.67-0.83) | 0.02 (0.02-0.03) | 1.00 (1.00-1.00) |
| | APRICOT-M | 0.72 (0.66-0.76) [c,d] | 0.05 (0.02-0.08) | 0.67 (0.42-0.81) | 0.66 (0.52-0.88) | 0.01 (0.01-0.03) [b,c] | 1.00 (0.99-1.00) |
| **VP** | | | | | | | |
| **Development** | | | | | | | |
| | CatBoost | 0.75 (0.74-0.76) | 0.02 (0.02-0.03) | 0.67 (0.61-0.74) | 0.69 (0.62-0.74) | 0.01 (0.01-0.01) | 1.00 (1.00-1.00) |
| | GRU | 0.80 (0.79-0.81) | 0.02 (0.02-0.02) | 0.76 (0.70-0.79) | 0.72 (0.70-0.78) | 0.01 (0.01-0.01) | 1.00 (1.00-1.00) |
| | Transformer | 0.79 (0.78-0.81) | 0.02 (0.02-0.02) | 0.69 (0.65-0.73) | 0.77 (0.73-0.81) | 0.01 (0.01-0.01) | 1.00 (1.00-1.00) |
| | APRICOT-T | 0.83 (0.81-0.84) | 0.02 (0.02-0.03) | 0.71 (0.66-0.76) | 0.80 (0.75-0.83) | 0.01 (0.01-0.02) | 1.00 (1.00-1.00) |
| | APRICOT-M | 0.83 (0.82-0.84) [a,b,c] | 0.03 (0.03-0.03) [a,b,c,d] | 0.75 (0.73-0.78) [a,c] | 0.78 (0.76-0.80) [a,b] | 0.01 (0.01-0.01) | 1.00 (1.00-1.00) |
| **External** | | | | | | | |
| | CatBoost | 0.71 (0.70-0.71) | 0.02 (0.02-0.02) | 0.65 (0.64-0.67) | 0.67 (0.66-0.69) | 0.02 (0.02-0.02) | 0.99 (0.99-1.00) |
| | GRU | 0.76 (0.75-0.76) | 0.03 (0.03-0.03) | 0.76 (0.72-0.79) | 0.63 (0.59-0.66) | 0.02 (0.02-0.02) | 1.00 (1.00-1.00) |
| | Transformer | 0.75 (0.74-0.75) | 0.06 (0.06-0.06) | 0.67 (0.64-0.69) | 0.70 (0.68-0.73) | 0.02 (0.02-0.02) | 1.00 (1.00-1.00) |
| | APRICOT-T | 0.82 (0.81-0.82) | 0.14 (0.14-0.15) | 0.72 (0.69-0.74) | 0.76 (0.74-0.79) | 0.03 (0.03-0.03) | 1.00 (1.00-1.00) |
| | APRICOT-M | 0.82 (0.81-0.82) [a,b,c] | 0.14 (0.13-0.15) [a,b,c] | 0.73 (0.71-0.75) [a,c] | 0.75 (0.73-0.77) [a,b,c] | 0.03 (0.03-0.03) [a,b,c] | 1.00 (1.00-1.00) [a] |
| **Temporal** | | | | | | | |
| | CatBoost | 0.66 (0.66-0.67) | 0.02 (0.02-0.02) | 0.58 (0.53-0.63) | 0.68 (0.63-0.72) | 0.02 (0.02-0.02) | 0.99 (0.99-0.99) |
| | GRU | 0.71 (0.70-0.72) | 0.06 (0.05-0.06) | 0.60 (0.58-0.62) | 0.74 (0.72-0.74) | 0.02 (0.02-0.02) | 0.99 (0.99-0.99) |
| | Transformer | 0.70 (0.69-0.71) | 0.04 (0.04-0.04) | 0.58 (0.53-0.62) | 0.73 (0.70-0.78) | 0.02 (0.02-0.03) | 0.99 (0.99-0.99) |
| | APRICOT-T | 0.73 (0.72-0.73) | 0.06 (0.05-0.06) | 0.59 (0.56-0.64) | 0.76 (0.71-0.80) | 0.03 (0.02-0.03) | 0.99 (0.99-0.99) |
| | APRICOT-M | 0.74 (0.73-0.74) [a,b,c,d] | 0.07 (0.06-0.08) [a,c] | 0.64 (0.60-0.68) | 0.73 (0.69-0.76) | 0.02 (0.02-0.03) [d] | 0.99 (0.99-1.00) |
| | **Prospective** | | | | | | |
| | CatBoost | 0.62 (0.58-0.66) | 0.02 (0.02-0.03) | 0.70 (0.51-0.82) | 0.52 (0.37-0.70) | 0.02 (0.01-0.02) | 0.99 (0.99-1.00) |
| | GRU | 0.62 (0.57-0.66) | 0.04 (0.02-0.06) | 0.53 (0.33-0.74) | 0.68 (0.44-0.85) | 0.02 (0.01-0.03) | 0.99 (0.99-0.99) |
| | Transformer | 0.62 (0.58-0.67) | 0.02 (0.02-0.03) | 0.57 (0.39-0.73) | 0.65 (0.51-0.81) | 0.02 (0.01-0.03) | 0.99 (0.99-0.99) |
| | APRICOT-T | 0.72 (0.68-0.75) | 0.05 (0.04-0.08) | 0.59 (0.48-0.70) | 0.76 (0.62-0.86) | 0.03 (0.02-0.04) | 0.99 (0.99-1.00) |
| | APRICOT-M | 0.71 (0.67-0.74) [a,b,c] | 0.04 (0.03-0.07) | 0.61 (0.54-0.80) | 0.74 (0.56-0.78) [a] | 0.03 (0.02-0.03) [a] | 0.99 (0.99-1.00) |
| **CRRT** | | | | | | | |
| | **Development** | | | | | | |
| | CatBoost | 0.80 (0.77-0.82) | 0.00 (0.00-0.00) | 0.78 (0.66-0.92) | 0.70 (0.55-0.81) | 0.00 (0.00-0.00) | 1.00 (1.00-1.00) |
| | GRU | 0.85 (0.83-0.87) | 0.00 (0.00-0.00) | 0.85 (0.78-0.91) | 0.73 (0.69-0.80) | 0.00 (0.00-0.00) | 1.00 (1.00-1.00) |
| | Transformer | 0.82 (0.79-0.85) | 0.00 (0.00-0.00) | 0.74 (0.66-0.89) | 0.78 (0.62-0.81) | 0.00 (0.00-0.00) | 1.00 (1.00-1.00) |
| | APRICOT-T | 0.89 (0.87-0.91) | 0.00 (0.00-0.01) | 0.87 (0.79-0.95) | 0.78 (0.73-0.84) | 0.00 (0.00-0.00) | 1.00 (1.00-1.00) |
| | APRICOT-M | 0.89 (0.86-0.91) [a,b,c] | 0.00 (0.00-0.01) | 0.83 (0.74-0.94) | 0.79 (0.67-0.87) | 0.00 (0.00-0.00) | 1.00 (1.00-1.00) |
| | **External** | | | | | | |
| | CatBoost | 0.84 (0.83-0.85) | 0.01 (0.01-0.01) | 0.78 (0.70-0.82) | 0.76 (0.72-0.85) | 0.00 (0.00-0.01) | 1.00 (1.00-1.00) |
| | GRU | 0.82 (0.82-0.83) | 0.00 (0.00-0.01) | 0.80 (0.74-0.87) | 0.71 (0.63-0.77) | 0.00 (0.00-0.00) | 1.00 (1.00-1.00) |
| | Transformer | 0.82 (0.82-0.83) | 0.01 (0.00-0.01) | 0.76 (0.72-0.80) | 0.75 (0.73-0.79) | 0.00 (0.00-0.00) | 1.00 (1.00-1.00) |
| | APRICOT-T | 0.87 (0.86-0.88) | 0.01 (0.01-0.01) | 0.84 (0.81-0.87) | 0.77 (0.74-0.78) | 0.00 (0.00-0.00) | 1.00 (1.00-1.00) |
| | APRICOT-M | 0.88 (0.87-0.89) [a,b,c] | 0.01 (0.01-0.01) [b] | 0.83 (0.79-0.86) [c] | 0.79 (0.76-0.82) [b] | 0.00 (0.00-0.00) | 1.00 (1.00-1.00) |
| | **Temporal** | | | | | | |
| | CatBoost | 0.91 (0.91-0.92) | 0.05 (0.04-0.05) | 0.81 (0.78-0.85) | 0.87 (0.83-0.89) | 0.02 (0.01-0.02) | 1.00 (1.00-1.00) |
| | GRU | 0.92 (0.91-0.93) | 0.04 (0.04-0.05) | 0.89 (0.84-0.92) | 0.82 (0.80-0.86) | 0.01 (0.01-0.02) | 1.00 (1.00-1.00) |
| | Transformer | 0.93 (0.92-0.93) | 0.04 (0.03-0.04) | 0.88 (0.83-0.93) | 0.83 (0.78-0.87) | 0.01 (0.01-0.02) | 1.00 (1.00-1.00) |
| | APRICOT-T | 0.95 (0.95-0.96) | 0.06 (0.06-0.07) | 0.91 (0.89-0.95) | 0.88 (0.83-0.89) | 0.02 (0.01-0.02) | 1.00 (1.00-1.00) |
| | APRICOT-M | 0.96 (0.95-0.96) [a,b,c,d] | 0.07 (0.06-0.08) [a,b,c] | 0.91 (0.88-0.95) [a] | 0.90 (0.86-0.92) [b,c] | 0.02 (0.02-0.03) [b,c] | 1.00 (1.00-1.00) |
| | **Prospective** | | | | | | |
| | CatBoost | 0.96 (0.94-0.99) | 0.00 (0.00-0.01) | 1.00 (1.00-1.00) | 0.95 (0.93-0.99) | 0.00 (0.00-0.01) | 1.00 (1.00-1.00) |
| | GRU | 0.94 (0.88-0.99) | 0.00 (0.00-0.01) | 1.00 (1.00-1.00) | 0.91 (0.88-0.99) | 0.00 (0.00-0.01) | 1.00 (1.00-1.00) |
| | Transformer | 0.81 (0.68-0.95) | 0.00 (0.00-0.00) | 1.00 (1.00-1.00) | 0.74 (0.68-0.95) | 0.00 (0.00-0.00) | 1.00 (1.00-1.00) |
| | APRICOT-T | 0.96 (0.94-0.98) | 0.00 (0.00-0.01) | 1.00 (1.00-1.00) | 0.94 (0.94-0.98) | 0.00 (0.00-0.01) | 1.00 (1.00-1.00) |
| | APRICOT-M | 0.96 (0.94-0.99) | 0.00 (0.00-0.01) | 1.00 (1.00-1.00) | 0.95 (0.93-0.99) | 0.00 (0.00-0.01) | 1.00 (1.00-1.00) |

Abbreviations: AUPRC: area under the precision-recall curve; AUROC: area under the receiver operating characteristic curve; CRRT: Continuous Renal Replacement Therapy; MV: Mechanical Ventilation; NPV: negative predictive value PPV: positive predictive value; VP: Vasopressors.
Metrics are shown as the median across 100-iteration bootstrap with 95% Confidence Intervals in parenthesis. P-values are based on pairwise Wilcoxon rank sum tests, with every baseline model compared to APRICOT-M. [a] p-value < 0.05 compared to CatBoost. [b] p-value < 0.05 compared to GRU. [c] p-value < 0.05 compared to Transformer. [d] p-value < 0.05 compared to APRICOT-T.

**Supplementary Table S7. Subgroup bias analysis on APRICOT-M model for each prediction task in validation cohorts.**

| Task/Validation cohort | | AUROC (95% CI) | AUPRC (95% CI) | Sensitivity (95% CI) | Specificity (95% CI) | PPV (95% CI) | NPV (95% CI) |
|---|---|---|---|---|---|---|---|
| **Discharge** | | | | | | | |
| *Development* | | | | | | | |
| | All | 0.87 (0.86-0.87) | 0.03 (0.03-0.03) | 0.84 (0.81-0.88) | 0.74 (0.70-0.77) | 0.02 (0.01-0.02) | 1.00 (1.00-1.00) |
| | Young | 0.86 (0.86-0.87) | 0.03 (0.03-0.04) | 0.84 (0.81-0.86) | 0.73 (0.72-0.76) | 0.02 (0.02-0.02) | 1.00 (1.00-1.00) |
| | Old | 0.87 (0.86-0.87)* | 0.03 (0.03-0.03) | 0.85 (0.82-0.88) | 0.74 (0.72-0.76) | 0.01 (0.01-0.01) | 1.00 (1.00-1.00) |
| | Female | 0.86 (0.85-0.88) | 0.03 (0.02-0.03) | 0.85 (0.82-0.86) | 0.75 (0.74-0.76) | 0.01 (0.01-0.02) | 1.00 (1.00-1.00) |
| | Male | 0.87 (0.87-0.87) | 0.03 (0.03-0.03) | 0.81 (0.75-0.83) | 0.77 (0.75-0.83) | 0.02 (0.02-0.02)* | 1.00 (1.00-1.00) |
| | Black | 0.85 (0.85-0.86) | 0.02 (0.02-0.03) | 0.83 (0.80-0.93) | 0.76 (0.67-0.78) | 0.01 (0.01-0.02) | 1.00 (1.00-1.00) |
| | Other | 0.86 (0.86-0.87) | 0.03 (0.02-0.03) | 0.85 (0.77-0.93) | 0.73 (0.64-0.79) | 0.01 (0.01-0.01) | 1.00 (1.00-1.00) |
| | White | 0.87 (0.87-0.87)** | 0.03 (0.03-0.03)* | 0.85 (0.84-0.87) | 0.74 (0.71-0.75) | 0.02 (0.02-0.02)** | 1.00 (1.00-1.00) |
| *External* | | | | | | | |
| | All | 0.87 (0.87-0.88) | 0.04 (0.03-0.04) | 0.86 (0.84-0.87) | 0.74 (0.73-0.75) | 0.02 (0.02-0.02) | 1.00 (1.00-1.00) |
| | Young | 0.86 (0.86-0.87) | 0.05 (0.04-0.05) | 0.87 (0.87-0.88) | 0.71 (0.70-0.71) | 0.02 (0.02-0.02) | 1.00 (1.00-1.00) |
| | Old | 0.88 (0.87-0.88)* | 0.03 (0.02-0.03)* | 0.84 (0.82-0.87)* | 0.76 (0.73-0.79)* | 0.01 (0.01-0.01) | 1.00 (1.00-1.00) |
| | Female | 0.87 (0.87-0.88) | 0.04 (0.04-0.04) | 0.85 (0.83-0.86) | 0.75 (0.74-0.77) | 0.02 (0.02-0.02) | 1.00 (1.00-1.00) |
| | Male | 0.87 (0.87-0.87) | 0.03 (0.03-0.03) | 0.86 (0.83-0.87) | 0.74 (0.73-0.77) | 0.02 (0.01-0.02) | 1.00 (1.00-1.00) |
| | Black | 0.86 (0.86-0.87) | 0.03 (0.03-0.03) | 0.83 (0.80-0.86) | 0.75 (0.72-0.78) | 0.01 (0.01-0.02) | 1.00 (1.00-1.00) |
| | Other | 0.89 (0.89-0.89) | 0.04 (0.03-0.04) | 0.85 (0.84-0.87) | 0.78 (0.76-0.78) | 0.02 (0.02-0.02) | 1.00 (1.00-1.00) |
| | White | 0.87 (0.87-0.87)* | 0.04 (0.03-0.04)* | 0.84 (0.83-0.86) | 0.75 (0.74-0.76)* | 0.02 (0.02-0.02)* | 1.00 (1.00-1.00) |
| *Temporal* | | | | | | | |
| | All | 0.80 (0.80-0.81) | 0.05 (0.05-0.05) | 0.76 (0.72-0.82) | 0.70 (0.64-0.74) | 0.03 (0.03-0.03) | 1.00 (1.00-1.00) |
| | Young | 0.81 (0.80-0.82) | 0.05 (0.05-0.05) | 0.80 (0.77-0.82) | 0.68 (0.66-0.70) | 0.03 (0.03-0.03) | 1.00 (1.00-1.00) |
| | Old | 0.80 (0.80-0.80) | 0.05 (0.05-0.05) | 0.77 (0.69-0.80) | 0.69 (0.66-0.76) | 0.03 (0.03-0.03) | 1.00 (1.00-1.00) |
| | Female | 0.80 (0.80-0.81) | 0.05 (0.04-0.05) | 0.77 (0.73-0.80) | 0.70 (0.68-0.72) | 0.03 (0.03-0.03) | 1.00 (1.00-1.00) |
| | Male | 0.80 (0.80-0.81) | 0.05 (0.05-0.05) | 0.77 (0.75-0.82) | 0.69 (0.64-0.71) | 0.03 (0.03-0.03) | 1.00 (1.00-1.00) |
| | Black | 0.80 (0.79-0.80) | 0.04 (0.04-0.05) | 0.76 (0.70-0.81) | 0.70 (0.66-0.75) | 0.03 (0.03-0.03) | 1.00 (1.00-1.00) |
| | Other | 0.79 (0.78-0.80) | 0.06 (0.05-0.06) | 0.73 (0.68-0.82) | 0.73 (0.65-0.78) | 0.03 (0.03-0.04) | 1.00 (0.99-1.00) |
| | White | 0.81 (0.80-0.81)** | 0.05 (0.05-0.05)** | 0.78 (0.75-0.84) | 0.68 (0.63-0.72) | 0.03 (0.03-0.03) | 1.00 (1.00-1.00) |
| **Stable** | | | | | | | |
| *Development* | | | | | | | |
| | All | 0.94 (0.94-0.95) | 0.95 (0.95-0.95) | 0.96 (0.95-0.96) | 0.84 (0.83-0.84) | 0.91 (0.91-0.91) | 0.91 (0.91-0.92) |
| | Young | 0.95 (0.95-0.95) | 0.95 (0.95-0.95) | 0.96 (0.96-0.96) | 0.84 (0.84-0.84) | 0.91 (0.91-0.91) | 0.92 (0.92-0.93) |
| | Old | 0.94 (0.94-0.94) | 0.95 (0.95-0.95) | 0.96 (0.95-0.96) | 0.83 (0.82-0.83)* | 0.91 (0.91-0.91) | 0.92 (0.91-0.92) |
| | Female | 0.94 (0.94-0.94) | 0.95 (0.95-0.95) | 0.96 (0.95-0.97) | 0.83 (0.83-0.84) | 0.91 (0.91-0.92) | 0.91 (0.90-0.93) |
| | Male | 0.95 (0.94-0.95)* | 0.95 (0.95-0.95) | 0.95 (0.95-0.96) | 0.84 (0.83-0.84)* | 0.91 (0.91-0.91) | 0.91 (0.91-0.92) |
| | Black | 0.95 (0.95-0.95) | 0.96 (0.96-0.96) | 0.96 (0.95-0.97) | 0.85 (0.85-0.86) | 0.92 (0.92-0.93) | 0.93 (0.91-0.94) |
| | Other | 0.96 (0.96-0.96) | 0.96 (0.96-0.96) | 0.94 (0.94-0.95) | 0.87 (0.86-0.87) | 0.90 (0.90-0.91) | 0.92 (0.92-0.92) |
| | White | 0.94 (0.94-0.94) | 0.95 (0.95-0.95) | 0.96 (0.96-0.96)* | 0.82 (0.82-0.82)** | 0.91 (0.91-0.91)** | 0.91 (0.91-0.92)** |
| **External** | | | | | | | |
| | All | 0.95 (0.94-0.95) | 0.95 (0.95-0.95) | 0.95 (0.95-0.95) | 0.86 (0.85-0.86) | 0.91 (0.91-0.91) | 0.91 (0.91-0.91) |
| | Young | 0.95 (0.95-0.95) | 0.95 (0.95-0.95) | 0.95 (0.95-0.95) | 0.86 (0.86-0.86) | 0.91 (0.91-0.91) | 0.92 (0.92-0.92) |
| | Old | 0.94 (0.94-0.94) | 0.95 (0.95-0.95) | 0.95 (0.94-0.95) | 0.85 (0.85-0.85) | 0.91 (0.91-0.91) | 0.90 (0.90-0.91)* |
| | Female | 0.94 (0.94-0.94) | 0.95 (0.95-0.95) | 0.95 (0.94-0.95) | 0.85 (0.85-0.86) | 0.92 (0.92-0.92) | 0.90 (0.90-0.91) |
| | Male | 0.95 (0.95-0.95)* | 0.95 (0.95-0.95) | 0.95 (0.95-0.95) | 0.86 (0.86-0.86)* | 0.91 (0.91-0.91) | 0.91 (0.91-0.91)* |
| | Black | 0.94 (0.94-0.94) | 0.95 (0.95-0.95) | 0.94 (0.94-0.94) | 0.85 (0.85-0.86) | 0.91 (0.91-0.91) | 0.90 (0.89-0.90) |
| | Other | 0.96 (0.96-0.96) | 0.95 (0.95-0.95) | 0.95 (0.95-0.95) | 0.88 (0.88-0.88) | 0.91 (0.91-0.91) | 0.93 (0.93-0.93) |
| | White | 0.94 (0.94-0.94) | 0.95 (0.95-0.95) | 0.95 (0.95-0.95)* | 0.85 (0.85-0.85) | 0.91 (0.91-0.92) | 0.91 (0.90-0.91)** |
| **Temporal** | | | | | | | |
| | All | 0.95 (0.95-0.95) | 0.97 (0.97-0.97) | 0.97 (0.96-0.97) | 0.85 (0.85-0.86) | 0.95 (0.95-0.95) | 0.90 (0.88-0.91) |
| | Young | 0.95 (0.95-0.95) | 0.97 (0.97-0.97) | 0.97 (0.97-0.97) | 0.86 (0.86-0.86) | 0.95 (0.95-0.95) | 0.91 (0.90-0.91) |
| | Old | 0.94 (0.94-0.95)* | 0.97 (0.97-0.97) | 0.96 (0.96-0.96) | 0.85 (0.85-0.85) | 0.95 (0.95-0.95) | 0.88 (0.88-0.89)* |
| | Female | 0.95 (0.94-0.95) | 0.97 (0.97-0.97) | 0.97 (0.96-0.97) | 0.85 (0.84-0.85) | 0.95 (0.95-0.95) | 0.89 (0.88-0.90) |
| | Male | 0.95 (0.95-0.95) | 0.97 (0.97-0.97) | 0.97 (0.97-0.97) | 0.85 (0.85-0.86) | 0.95 (0.95-0.95) | 0.91 (0.91-0.91)* |
| | Black | 0.95 (0.95-0.95) | 0.97 (0.97-0.97) | 0.97 (0.96-0.97) | 0.87 (0.87-0.88) | 0.95 (0.95-0.95) | 0.90 (0.90-0.92) |
| | Other | 0.95 (0.95-0.95) | 0.97 (0.97-0.97) | 0.96 (0.95-0.96) | 0.86 (0.85-0.87) | 0.95 (0.95-0.96) | 0.88 (0.86-0.89) |
| | White | 0.95 (0.94-0.95) | 0.97 (0.97-0.97) | 0.97 (0.97-0.97)* | 0.84 (0.84-0.84)** | 0.95 (0.95-0.95) | 0.90 (0.90-0.91)* |
| **Unstable** | | | | | | | |
| **Development** | | | | | | | |
| | All | 0.95 (0.95-0.95) | 0.94 (0.94-0.94) | 0.85 (0.84-0.85) | 0.95 (0.95-0.96) | 0.91 (0.90-0.92) | 0.92 (0.92-0.92) |
| | Young | 0.96 (0.96-0.96) | 0.95 (0.94-0.95) | 0.86 (0.85-0.86) | 0.95 (0.95-0.96) | 0.91 (0.91-0.92) | 0.92 (0.92-0.92) |
| | Old | 0.95 (0.95-0.95) | 0.94 (0.94-0.94)* | 0.84 (0.84-0.84)* | 0.95 (0.95-0.95) | 0.90 (0.90-0.91)* | 0.92 (0.92-0.92) |
| | Female | 0.95 (0.95-0.95) | 0.94 (0.94-0.94) | 0.84 (0.84-0.85) | 0.96 (0.95-0.96) | 0.91 (0.90-0.92) | 0.92 (0.92-0.92) |
| | Male | 0.95 (0.95-0.95) | 0.94 (0.94-0.94) | 0.85 (0.84-0.85)* | 0.95 (0.95-0.96)* | 0.91 (0.91-0.92) | 0.92 (0.92-0.92) |
| | Black | 0.96 (0.96-0.96) | 0.95 (0.95-0.95) | 0.87 (0.87-0.87) | 0.96 (0.96-0.96) | 0.92 (0.92-0.92) | 0.93 (0.93-0.93) |
| | Other | 0.96 (0.96-0.96) | 0.96 (0.96-0.96) | 0.88 (0.87-0.88) | 0.94 (0.93-0.94) | 0.91 (0.90-0.92) | 0.91 (0.90-0.92) |
| | White | 0.95 (0.95-0.95) | 0.93 (0.93-0.93) | 0.83 (0.82-0.83)** | 0.96 (0.96-0.96)* | 0.91 (0.91-0.92)* | 0.92 (0.92-0.92) |
| **External** | | | | | | | |
| | All | 0.95 (0.95-0.95) | 0.94 (0.94-0.94) | 0.87 (0.87-0.87) | 0.95 (0.94-0.95) | 0.91 (0.90-0.91) | 0.92 (0.92-0.92) |
| | Young | 0.96 (0.96-0.96) | 0.94 (0.94-0.94) | 0.88 (0.87-0.88) | 0.95 (0.95-0.95) | 0.92 (0.91-0.92) | 0.93 (0.92-0.93) |
| | Old | 0.95 (0.95-0.95) | 0.93 (0.93-0.93) | 0.86 (0.86-0.86)* | 0.94 (0.94-0.95)* | 0.90 (0.90-0.90)* | 0.92 (0.92-0.92)* |
| | Female | 0.95 (0.95-0.95) | 0.93 (0.93-0.94) | 0.87 (0.86-0.87) | 0.95 (0.94-0.95) | 0.90 (0.90-0.90) | 0.93 (0.93-0.93) |
| | Male | 0.95 (0.95-0.95) | 0.94 (0.94-0.94)* | 0.87 (0.87-0.87) | 0.95 (0.95-0.95) | 0.91 (0.91-0.91)* | 0.92 (0.92-0.92) |
| | Black | 0.95 (0.95-0.95) | 0.93 (0.93-0.93) | 0.86 (0.86-0.87) | 0.94 (0.94-0.94) | 0.89 (0.89-0.90) | 0.92 (0.92-0.92) |
| | Other | 0.96 (0.96-0.96) | 0.95 (0.95-0.95) | 0.89 (0.89-0.89) | 0.95 (0.95-0.95) | 0.93 (0.93-0.93) | 0.92 (0.92-0.92) |
| | White | 0.95 (0.95-0.95) | 0.93 (0.93-0.93) | 0.86 (0.86-0.86) | 0.95 (0.94-0.95)* | 0.90 (0.90-0.90)* | 0.92 (0.92-0.93) |
| **Temporal** | | | | | | | |
| | All | 0.97 (0.97-0.97) | 0.95 (0.95-0.95) | 0.89 (0.89-0.89) | 0.97 (0.96-0.97) | 0.89 (0.88-0.90) | 0.97 (0.96-0.97) |
| | Young | 0.97 (0.97-0.97) | 0.95 (0.95-0.95) | 0.90 (0.90-0.90) | 0.97 (0.97-0.97) | 0.90 (0.89-0.90) | 0.97 (0.97-0.97) |
| | Old | 0.97 (0.97-0.97) | 0.95 (0.95-0.95) | 0.89 (0.88-0.89)* | 0.96 (0.96-0.97)* | 0.88 (0.87-0.89)* | 0.96 (0.96-0.97)* |
| | Female | 0.97 (0.97-0.97) | 0.95 (0.95-0.95) | 0.88 (0.88-0.89) | 0.97 (0.96-0.97) | 0.88 (0.87-0.89) | 0.96 (0.96-0.97) |
| | Male | 0.97 (0.97-0.97) | 0.95 (0.95-0.95) | 0.90 (0.89-0.90)* | 0.96 (0.96-0.97)* | 0.89 (0.89-0.90) | 0.97 (0.96-0.97)* |
| | Black | 0.97 (0.97-0.98) | 0.96 (0.96-0.96) | 0.90 (0.90-0.90) | 0.97 (0.97-0.97) | 0.91 (0.90-0.91) | 0.97 (0.97-0.97) |
| | Other | 0.97 (0.97-0.97) | 0.95 (0.94-0.95) | 0.89 (0.88-0.90) | 0.96 (0.95-0.97) | 0.87 (0.85-0.90) | 0.97 (0.96-0.97) |
| | White | 0.97 (0.97-0.97) | 0.95 (0.95-0.95) | 0.88 (0.88-0.89)* | 0.97 (0.96-0.97) | 0.89 (0.89-0.90)* | 0.96 (0.96-0.96)* |
| **Deceased** | | | | | | | |
| **Development** | | | | | | | |
| | All | 0.93 (0.92-0.94) | 0.14 (0.12-0.16) | 0.86 (0.82-0.88) | 0.88 (0.86-0.91) | 0.02 (0.01-0.02) | 1.00 (1.00-1.00) |
| | Young | 0.94 (0.93-0.94) | 0.10 (0.09-0.12) | 0.84 (0.83-0.86) | 0.91 (0.90-0.93) | 0.02 (0.01-0.02) | 1.00 (1.00-1.00) |
| | Old | 0.93 (0.93-0.93)* | 0.16 (0.15-0.17)* | 0.87 (0.84-0.89)* | 0.86 (0.85-0.88)* | 0.02 (0.02-0.02) | 1.00 (1.00-1.00) |
| | Female | 0.94 (0.94-0.94) | 0.15 (0.13-0.18) | 0.89 (0.87-0.90) | 0.88 (0.88-0.88) | 0.02 (0.02-0.02) | 1.00 (1.00-1.00) |
| | Male | 0.93 (0.92-0.93)* | 0.11 (0.10-0.13)* | 0.84 (0.83-0.85)* | 0.87 (0.85-0.89) | 0.01 (0.01-0.02)* | 1.00 (1.00-1.00) |
| | Black | 0.94 (0.93-0.95) | 0.15 (0.12-0.19) | 0.91 (0.87-0.94) | 0.88 (0.86-0.91) | 0.02 (0.01-0.02) | 1.00 (1.00-1.00) |
| | Other | 0.94 (0.93-0.94) | 0.14 (0.11-0.17) | 0.89 (0.87-0.91) | 0.86 (0.86-0.87) | 0.02 (0.01-0.02) | 1.00 (1.00-1.00) |
| | White | 0.93 (0.92-0.93)* | 0.13 (0.11-0.14) | 0.85 (0.82-0.87)** | 0.88 (0.85-0.90) | 0.02 (0.01-0.02) | 1.00 (1.00-1.00) |
| **External** | | | | | | | |
| | All | 0.95 (0.94-0.95) | 0.18 (0.16-0.19) | 0.86 (0.84-0.88) | 0.89 (0.87-0.91) | 0.02 (0.02-0.03) | 1.00 (1.00-1.00) |
| | Young | 0.95 (0.94-0.96) | 0.15 (0.13-0.16) | 0.86 (0.85-0.88) | 0.89 (0.87-0.90) | 0.02 (0.01-0.02) | 1.00 (1.00-1.00) |
| | Old | 0.95 (0.94-0.95) | 0.18 (0.18-0.19)* | 0.86 (0.84-0.87) | 0.89 (0.87-0.90) | 0.02 (0.02-0.03) | 1.00 (1.00-1.00) |
| | Female | 0.95 (0.95-0.95) | 0.18 (0.17-0.19) | 0.85 (0.83-0.87) | 0.90 (0.88-0.92) | 0.02 (0.02-0.03) | 1.00 (1.00-1.00) |
| | Male | 0.95 (0.95-0.95) | 0.17 (0.17-0.18) | 0.85 (0.84-0.88) | 0.90 (0.88-0.92) | 0.02 (0.02-0.03) | 1.00 (1.00-1.00) |
| | Black | 0.95 (0.94-0.96) | 0.20 (0.17-0.22) | 0.85 (0.81-0.90) | 0.91 (0.88-0.93) | 0.02 (0.02-0.03) | 1.00 (1.00-1.00) |
| | Other | 0.95 (0.94-0.95) | 0.19 (0.18-0.20) | 0.89 (0.87-0.91) | 0.87 (0.85-0.88) | 0.02 (0.02-0.03) | 1.00 (1.00-1.00) |
| | White | 0.95 (0.95-0.95) | 0.17 (0.17-0.17)** | 0.85 (0.85-0.86)* | 0.90 (0.89-0.90)* | 0.02 (0.02-0.02) | 1.00 (1.00-1.00) |
| **Temporal** | | | | | | | |
| | All | 0.98 (0.97-0.98) | 0.42 (0.39-0.45) | 0.91 (0.88-0.94) | 0.94 (0.92-0.96) | 0.03 (0.02-0.04) | 1.00 (1.00-1.00) |
| | Young | 0.99 (0.98-0.99) | 0.47 (0.45-0.49) | 0.94 (0.92-0.97) | 0.95 (0.94-0.96) | 0.03 (0.02-0.03) | 1.00 (1.00-1.00) |
| | Old | 0.97 (0.97-0.97)* | 0.40 (0.36-0.42)* | 0.90 (0.89-0.91)* | 0.94 (0.94-0.94) | 0.03 (0.03-0.04) | 1.00 (1.00-1.00) |
| | Female | 0.98 (0.97-0.99) | 0.47 (0.45-0.49) | 0.93 (0.91-0.96) | 0.95 (0.95-0.97) | 0.04 (0.03-0.05) | 1.00 (1.00-1.00) |
| | Male | 0.98 (0.97-0.98) | 0.39 (0.37-0.41)* | 0.91 (0.89-0.93) | 0.93 (0.91-0.95) | 0.02 (0.02-0.03)* | 1.00 (1.00-1.00) |
| | Black | 0.98 (0.97-0.98) | 0.39 (0.34-0.49) | 0.93 (0.91-0.95) | 0.94 (0.93-0.95) | 0.03 (0.02-0.03) | 1.00 (1.00-1.00) |
| | Other | 0.97 (0.96-0.98) | 0.40 (0.36-0.46) | 0.92 (0.85-0.99) | 0.90 (0.82-0.98) | 0.04 (0.01-0.08) | 1.00 (1.00-1.00) |
| | White | 0.98 (0.98-0.98) | 0.44 (0.41-0.46) | 0.91 (0.89-0.92) | 0.95 (0.95-0.95) | 0.03 (0.03-0.04) | 1.00 (1.00-1.00) |
| **Unstable-Stable** | | | | | | | |
| **Development** | | | | | | | |
| | All | 0.74 (0.73-0.74) | 0.10 (0.09-0.10) | 0.65 (0.63-0.67) | 0.70 (0.68-0.72) | 0.04 (0.04-0.04) | 0.99 (0.99-0.99) |
| | Young | 0.73 (0.73-0.74) | 0.09 (0.08-0.10) | 0.63 (0.60-0.65) | 0.71 (0.69-0.74) | 0.04 (0.04-0.04) | 0.99 (0.99-0.99) |
| | Old | 0.74 (0.74-0.75)* | 0.10 (0.09-0.10) | 0.66 (0.64-0.67)* | 0.70 (0.69-0.70) | 0.04 (0.04-0.04) | 0.99 (0.99-0.99) |
| | Female | 0.73 (0.73-0.74) | 0.09 (0.08-0.09) | 0.63 (0.61-0.65) | 0.71 (0.69-0.73) | 0.04 (0.04-0.04) | 0.99 (0.99-0.99) |
| | Male | 0.74 (0.74-0.75)* | 0.10 (0.10-0.11)* | 0.67 (0.66-0.67)* | 0.69 (0.68-0.70) | 0.04 (0.04-0.04) | 0.99 (0.99-0.99) |
| | Black | 0.71 (0.71-0.72) | 0.08 (0.07-0.08) | 0.60 (0.58-0.63) | 0.74 (0.73-0.75) | 0.03 (0.03-0.03) | 0.99 (0.99-0.99) |
| | Other | 0.76 (0.75-0.77) | 0.13 (0.11-0.14) | 0.65 (0.62-0.69) | 0.73 (0.69-0.76) | 0.05 (0.05-0.05) | 0.99 (0.99-0.99) |
| | White | 0.74 (0.73-0.74)** | 0.09 (0.09-0.09)** | 0.67 (0.66-0.67)* | 0.68 (0.67-0.69)** | 0.04 (0.04-0.04)* | 0.99 (0.99-0.99) |
| **External** | | | | | | | |
| | All | 0.79 (0.79-0.79) | 0.14 (0.13-0.14) | 0.65 (0.63-0.67) | 0.80 (0.77-0.82) | 0.07 (0.06-0.07) | 0.99 (0.99-0.99) |
| | Young | 0.80 (0.80-0.80) | 0.14 (0.14-0.15) | 0.67 (0.65-0.68) | 0.79 (0.78-0.81) | 0.06 (0.06-0.07) | 0.99 (0.99-0.99) |
| | Old | 0.79 (0.79-0.79) | 0.13 (0.13-0.14)* | 0.65 (0.64-0.66)* | 0.80 (0.78-0.81) | 0.07 (0.07-0.07)* | 0.99 (0.99-0.99) |
| | Female | 0.79 (0.78-0.79) | 0.13 (0.12-0.13) | 0.66 (0.64-0.67) | 0.79 (0.78-0.81) | 0.06 (0.06-0.07) | 0.99 (0.99-0.99) |
| | Male | 0.79 (0.79-0.80) | 0.14 (0.14-0.15)* | 0.65 (0.63-0.66) | 0.81 (0.79-0.82) | 0.07 (0.07-0.08)* | 0.99 (0.99-0.99) |
| | Black | 0.78 (0.78-0.79) | 0.12 (0.11-0.13) | 0.64 (0.63-0.65) | 0.80 (0.80-0.81) | 0.06 (0.05-0.06) | 0.99 (0.99-0.99) |
| | Other | 0.81 (0.80-0.81) | 0.15 (0.14-0.15) | 0.69 (0.65-0.71) | 0.79 (0.77-0.82) | 0.07 (0.07-0.08) | 0.99 (0.99-0.99) |
| | White | 0.79 (0.78-0.79)** | 0.14 (0.13-0.14)** | 0.65 (0.63-0.66)* | 0.79 (0.78-0.81) | 0.07 (0.06-0.07)* | 0.99 (0.99-0.99) |
| **Temporal** | | | | | | | |
| | All | 0.91 (0.91-0.92) | 0.44 (0.42-0.45) | 0.79 (0.77-0.81) | 0.88 (0.87-0.89) | 0.11 (0.10-0.12) | 1.00 (1.00-1.00) |
| | Young | 0.92 (0.92-0.92) | 0.46 (0.44-0.48) | 0.81 (0.80-0.83) | 0.88 (0.86-0.89) | 0.11 (0.09-0.12) | 1.00 (1.00-1.00) |
| | Old | 0.91 (0.91-0.91) | 0.41 (0.41-0.42)* | 0.77 (0.75-0.78)* | 0.88 (0.87-0.89) | 0.10 (0.10-0.11) | 1.00 (1.00-1.00) |
| | Female | 0.91 (0.90-0.91) | 0.43 (0.41-0.44) | 0.78 (0.77-0.79) | 0.87 (0.87-0.88) | 0.10 (0.09-0.10) | 1.00 (1.00-1.00) |
|  | Male | 0.92 (0.92-0.92)* | 0.44 (0.43-0.45) | 0.80 (0.78-0.81)* | 0.89 (0.88-0.91)* | 0.12 (0.11-0.14)* | 1.00 (1.00-1.00) |
|  | Black | 0.91 (0.91-0.92) | 0.44 (0.41-0.46) | 0.81 (0.79-0.83) | 0.86 (0.85-0.89) | 0.09 (0.08-0.10) | 1.00 (1.00-1.00) |
|  | Other | 0.93 (0.91-0.94) | 0.45 (0.42-0.47) | 0.78 (0.76-0.81) | 0.91 (0.91-0.92) | 0.14 (0.13-0.16) | 1.00 (1.00-1.00) |
|  | White | 0.91 (0.91-0.92)* | 0.43 (0.41-0.43) | 0.79 (0.78-0.79) | 0.89 (0.89-0.89)** | 0.11 (0.11-0.11)** | 1.00 (1.00-1.00) |
| **Stable-Unstable** |  |  |  |  |  |  |  |
| **Development** |  |  |  |  |  |  |  |
|  | All | 0.84 (0.83-0.85) | 0.12 (0.11-0.14) | 0.75 (0.68-0.81) | 0.75 (0.69-0.81) | 0.02 (0.01-0.02) | 1.00 (1.00-1.00) |
|  | Young | 0.85 (0.83-0.85) | 0.13 (0.11-0.15) | 0.73 (0.68-0.80) | 0.78 (0.73-0.82) | 0.02 (0.02-0.02) | 1.00 (1.00-1.00) |
|  | Old | 0.83 (0.83-0.84)* | 0.12 (0.11-0.14) | 0.76 (0.74-0.77) | 0.73 (0.72-0.75)* | 0.02 (0.02-0.02) | 1.00 (1.00-1.00) |
|  | Female | 0.83 (0.82-0.84) | 0.09 (0.08-0.10) | 0.72 (0.64-0.79) | 0.77 (0.71-0.84) | 0.02 (0.01-0.02) | 1.00 (1.00-1.00) |
|  | Male | 0.84 (0.84-0.85) | 0.15 (0.15-0.16)* | 0.76 (0.75-0.77) | 0.75 (0.75-0.76) | 0.02 (0.02-0.02) | 1.00 (1.00-1.00) |
|  | Black | 0.84 (0.82-0.87) | 0.15 (0.12-0.20) | 0.77 (0.71-0.81) | 0.77 (0.74-0.79) | 0.02 (0.01-0.02) | 1.00 (1.00-1.00) |
|  | Other | 0.87 (0.86-0.88) | 0.09 (0.07-0.09) | 0.82 (0.79-0.87) | 0.73 (0.67-0.77) | 0.02 (0.01-0.02) | 1.00 (1.00-1.00) |
|  | White | 0.83 (0.83-0.84)* | 0.13 (0.13-0.13)* | 0.74 (0.73-0.76)* | 0.74 (0.73-0.77) | 0.02 (0.02-0.02) | 1.00 (1.00-1.00) |
| **External** |  |  |  |  |  |  |  |
|  | All | 0.82 (0.81-0.82) | 0.22 (0.21-0.22) | 0.66 (0.64-0.69) | 0.81 (0.78-0.83) | 0.03 (0.03-0.03) | 1.00 (1.00-1.00) |
|  | Young | 0.81 (0.81-0.81) | 0.18 (0.17-0.19) | 0.65 (0.63-0.66) | 0.82 (0.81-0.83) | 0.03 (0.02-0.03) | 1.00 (1.00-1.00) |
|  | Old | 0.81 (0.81-0.81) | 0.23 (0.22-0.24)* | 0.66 (0.65-0.67) | 0.80 (0.79-0.81)* | 0.03 (0.03-0.03) | 1.00 (1.00-1.00) |
|  | Female | 0.80 (0.80-0.80) | 0.16 (0.16-0.17) | 0.66 (0.64-0.67) | 0.78 (0.78-0.80) | 0.02 (0.02-0.03) | 1.00 (1.00-1.00) |
|  | Male | 0.82 (0.82-0.83)* | 0.24 (0.24-0.25)* | 0.67 (0.66-0.68) | 0.82 (0.81-0.83)* | 0.03 (0.03-0.03)* | 1.00 (1.00-1.00) |
|  | Black | 0.79 (0.79-0.81) | 0.14 (0.14-0.15) | 0.65 (0.62-0.68) | 0.79 (0.78-0.81) | 0.02 (0.02-0.02) | 1.00 (1.00-1.00) |
|  | Other | 0.83 (0.82-0.83) | 0.25 (0.24-0.25) | 0.73 (0.67-0.75) | 0.76 (0.74-0.82) | 0.03 (0.02-0.03) | 1.00 (1.00-1.00) |
|  | White | 0.81 (0.81-0.81)** | 0.21 (0.21-0.22)** | 0.65 (0.63-0.66)* | 0.82 (0.80-0.83)** | 0.03 (0.03-0.03)* | 1.00 (1.00-1.00) |
| **Temporal** |  |  |  |  |  |  |  |
|  | All | 0.78 (0.77-0.78) | 0.08 (0.07-0.09) | 0.67 (0.64-0.72) | 0.72 (0.68-0.76) | 0.02 (0.02-0.03) | 1.00 (0.99-1.00) |
|  | Young | 0.78 (0.78-0.79) | 0.08 (0.07-0.09) | 0.67 (0.63-0.69) | 0.74 (0.72-0.76) | 0.02 (0.02-0.03) | 1.00 (1.00-1.00) |
|  | Old | 0.78 (0.77-0.79) | 0.08 (0.07-0.08) | 0.67 (0.63-0.71) | 0.73 (0.68-0.77) | 0.03 (0.02-0.03)* | 1.00 (0.99-1.00) |
|  | Female | 0.77 (0.76-0.78) | 0.07 (0.06-0.07) | 0.67 (0.64-0.70) | 0.72 (0.68-0.75) | 0.02 (0.02-0.02) | 1.00 (1.00-1.00) |
|  | Male | 0.78 (0.78-0.78) | 0.08 (0.08-0.09)* | 0.65 (0.62-0.67) | 0.75 (0.74-0.78) | 0.03 (0.03-0.03)* | 0.99 (0.99-1.00)* |
|  | Black | 0.81 (0.80-0.81) | 0.08 (0.07-0.09) | 0.70 (0.62-0.76) | 0.77 (0.70-0.83) | 0.03 (0.02-0.03) | 1.00 (1.00-1.00) |
|  | Other | 0.78 (0.77-0.81) | 0.10 (0.08-0.13) | 0.81 (0.72-0.89) | 0.60 (0.49-0.68) | 0.02 (0.02-0.02) | 1.00 (1.00-1.00) |
|  | White | 0.77 (0.76-0.77)* | 0.08 (0.07-0.08) | 0.64 (0.63-0.66)* | 0.74 (0.73-0.74)* | 0.03 (0.02-0.03)* | 0.99 (0.99-1.00)** |
| **MV** |  |  |  |  |  |  |  |
| **Development** |  |  |  |  |  |  |  |
|  | All | 0.85 (0.84-0.86) | 0.14 (0.12-0.15) | 0.73 (0.68-0.78) | 0.79 (0.74-0.83) | 0.02 (0.02-0.02) | 1.00 (1.00-1.00) |
| | Young | 0.86 (0.85-0.86) | 0.15 (0.13-0.16) | 0.75 (0.70-0.80) | 0.79 (0.74-0.85) | 0.02 (0.01-0.02) | 1.00 (1.00-1.00) |
| | Old | 0.84 (0.84-0.85)* | 0.12 (0.11-0.13)* | 0.73 (0.68-0.78) | 0.79 (0.75-0.82) | 0.02 (0.02-0.02) | 1.00 (1.00-1.00) |
| | Female | 0.84 (0.83-0.85) | 0.10 (0.08-0.11) | 0.72 (0.70-0.75) | 0.79 (0.77-0.81) | 0.02 (0.02-0.02) | 1.00 (1.00-1.00) |
| | Male | 0.86 (0.85-0.86)* | 0.16 (0.15-0.17)* | 0.72 (0.69-0.75) | 0.82 (0.79-0.84) | 0.02 (0.02-0.02) | 1.00 (1.00-1.00) |
| | Black | 0.84 (0.82-0.86) | 0.14 (0.11-0.19) | 0.71 (0.66-0.77) | 0.81 (0.80-0.87) | 0.02 (0.01-0.02) | 1.00 (1.00-1.00) |
| | Other | 0.85 (0.84-0.87) | 0.08 (0.07-0.10) | 0.75 (0.71-0.81) | 0.80 (0.77-0.83) | 0.02 (0.02-0.02) | 1.00 (1.00-1.00) |
| | White | 0.85 (0.84-0.85) | 0.14 (0.14-0.15)* | 0.73 (0.70-0.76) | 0.79 (0.74-0.82) | 0.02 (0.02-0.02) | 1.00 (1.00-1.00) |
| **External** | | | | | | | |
| | All | 0.82 (0.82-0.83) | 0.23 (0.23-0.24) | 0.65 (0.64-0.69) | 0.85 (0.82-0.87) | 0.03 (0.02-0.03) | 1.00 (1.00-1.00) |
| | Young | 0.82 (0.82-0.82) | 0.20 (0.18-0.20) | 0.68 (0.65-0.71) | 0.82 (0.79-0.85) | 0.02 (0.02-0.02) | 1.00 (1.00-1.00) |
| | Old | 0.83 (0.82-0.83)* | 0.25 (0.24-0.26)* | 0.65 (0.64-0.66) | 0.86 (0.84-0.87)* | 0.03 (0.03-0.03)* | 1.00 (1.00-1.00) |
| | Female | 0.81 (0.80-0.81) | 0.18 (0.17-0.19) | 0.67 (0.66-0.69) | 0.80 (0.79-0.82) | 0.02 (0.02-0.02) | 1.00 (1.00-1.00) |
| | Male | 0.83 (0.83-0.84)* | 0.26 (0.25-0.27)* | 0.67 (0.64-0.70) | 0.85 (0.82-0.87)* | 0.03 (0.03-0.03)* | 1.00 (1.00-1.00) |
| | Black | 0.79 (0.78-0.80) | 0.14 (0.13-0.16) | 0.65 (0.63-0.67) | 0.81 (0.80-0.81) | 0.02 (0.02-0.02) | 1.00 (1.00-1.00) |
| | Other | 0.83 (0.82-0.84) | 0.27 (0.26-0.28) | 0.72 (0.67-0.74) | 0.80 (0.78-0.85) | 0.02 (0.02-0.03) | 1.00 (1.00-1.00) |
| | White | 0.82 (0.82-0.83)* | 0.23 (0.23-0.24)** | 0.66 (0.65-0.67)* | 0.85 (0.84-0.85)** | 0.03 (0.03-0.03)** | 1.00 (1.00-1.00) |
| **Temporal** | | | | | | | |
| | All | 0.87 (0.87-0.88) | 0.12 (0.11-0.14) | 0.79 (0.75-0.83) | 0.81 (0.77-0.85) | 0.02 (0.02-0.03) | 1.00 (1.00-1.00) |
| | Young | 0.88 (0.87-0.88) | 0.12 (0.11-0.13) | 0.78 (0.77-0.79) | 0.83 (0.82-0.84) | 0.03 (0.03-0.03) | 1.00 (1.00-1.00) |
| | Old | 0.87 (0.87-0.88)* | 0.12 (0.11-0.14) | 0.79 (0.76-0.81) | 0.80 (0.78-0.83)* | 0.02 (0.02-0.03)* | 1.00 (1.00-1.00) |
| | Female | 0.86 (0.85-0.87) | 0.11 (0.09-0.13) | 0.80 (0.74-0.83) | 0.78 (0.75-0.85) | 0.02 (0.02-0.03) | 1.00 (1.00-1.00) |
| | Male | 0.88 (0.88-0.88)* | 0.12 (0.11-0.13) | 0.80 (0.79-0.81) | 0.81 (0.81-0.82) | 0.03 (0.03-0.03)* | 1.00 (1.00-1.00) |
| | Black | 0.87 (0.87-0.88) | 0.10 (0.08-0.13) | 0.82 (0.81-0.82) | 0.82 (0.82-0.82) | 0.02 (0.02-0.02) | 1.00 (1.00-1.00) |
| | Other | 0.90 (0.88-0.92) | 0.19 (0.13-0.23) | 0.75 (0.68-0.84) | 0.90 (0.84-0.95) | 0.04 (0.03-0.06) | 1.00 (1.00-1.00) |
| | White | 0.87 (0.87-0.88)* | 0.13 (0.12-0.13)** | 0.77 (0.74-0.84) | 0.82 (0.76-0.85)* | 0.03 (0.02-0.03)* | 1.00 (1.00-1.00) |
| **VP** | | | | | | | |
| **Development** | | | | | | | |
| | All | 0.83 (0.82-0.84) | 0.03 (0.03-0.03) | 0.75 (0.73-0.78) | 0.78 (0.76-0.80) | 0.01 (0.01-0.01) | 1.00 (1.00-1.00) |
| | Young | 0.82 (0.80-0.83) | 0.03 (0.02-0.03) | 0.72 (0.70-0.75) | 0.79 (0.78-0.80) | 0.01 (0.01-0.01) | 1.00 (1.00-1.00) |
| | Old | 0.83 (0.82-0.84) | 0.03 (0.03-0.03) | 0.74 (0.71-0.76) | 0.78 (0.75-0.82) | 0.01 (0.01-0.02) | 1.00 (1.00-1.00) |
| | Female | 0.82 (0.81-0.83) | 0.03 (0.02-0.03) | 0.74 (0.70-0.76) | 0.79 (0.77-0.81) | 0.01 (0.01-0.01) | 1.00 (1.00-1.00) |
| | Male | 0.83 (0.82-0.84) | 0.03 (0.03-0.04) | 0.72 (0.69-0.75) | 0.80 (0.77-0.83) | 0.01 (0.01-0.02) | 1.00 (1.00-1.00) |
| | Black | 0.80 (0.75-0.83) | 0.01 (0.01-0.02) | 0.74 (0.65-0.82) | 0.76 (0.69-0.82) | 0.01 (0.00-0.01) | 1.00 (1.00-1.00) |
| | Other | 0.83 (0.83-0.84) | 0.03 (0.02-0.04) | 0.75 (0.72-0.79) | 0.78 (0.72-0.83) | 0.02 (0.01-0.02) | 1.00 (1.00-1.00) |
| | White | 0.83 (0.82-0.83) | 0.03 (0.02-0.03)* | 0.74 (0.73-0.75) | 0.79 (0.76-0.81) | 0.01 (0.01-0.02)* | 1.00 (1.00-1.00) |
| **External** | | | | | | | |
|  | All | 0.82 (0.81-0.82) | 0.14 (0.13-0.15) | 0.73 (0.71-0.75) | 0.75 (0.73-0.77) | 0.03 (0.03-0.03) | 1.00 (1.00-1.00) |
|  | Young | 0.83 (0.82-0.83) | 0.11 (0.11-0.11) | 0.75 (0.73-0.77) | 0.76 (0.75-0.77) | 0.02 (0.02-0.03) | 1.00 (1.00-1.00) |
|  | Old | 0.81 (0.81-0.81)* | 0.15 (0.15-0.16)* | 0.72 (0.70-0.74)* | 0.75 (0.73-0.78) | 0.03 (0.03-0.03)* | 1.00 (1.00-1.00) |
|  | Female | 0.80 (0.80-0.81) | 0.11 (0.10-0.11) | 0.71 (0.69-0.73) | 0.75 (0.75-0.77) | 0.03 (0.03-0.03) | 1.00 (1.00-1.00) |
|  | Male | 0.83 (0.82-0.83)* | 0.16 (0.15-0.17)* | 0.75 (0.70-0.77) | 0.75 (0.73-0.81) | 0.03 (0.03-0.04) | 1.00 (1.00-1.00) |
|  | Black | 0.81 (0.80-0.81) | 0.10 (0.09-0.10) | 0.73 (0.67-0.78) | 0.74 (0.70-0.78) | 0.02 (0.02-0.02) | 1.00 (1.00-1.00) |
|  | Other | 0.82 (0.81-0.82) | 0.15 (0.14-0.15) | 0.74 (0.73-0.75) | 0.74 (0.72-0.75) | 0.03 (0.03-0.03) | 1.00 (1.00-1.00) |
|  | White | 0.82 (0.81-0.82)* | 0.14 (0.13-0.15)* | 0.71 (0.70-0.72)* | 0.78 (0.77-0.79)* | 0.03 (0.03-0.03)* | 1.00 (1.00-1.00) |
| **Temporal** |  |  |  |  |  |  |  |
|  | All | 0.74 (0.73-0.74) | 0.07 (0.06-0.08) | 0.64 (0.60-0.68) | 0.73 (0.69-0.76) | 0.02 (0.02-0.03) | 0.99 (0.99-1.00) |
|  | Young | 0.74 (0.73-0.75) | 0.07 (0.06-0.08) | 0.67 (0.66-0.69) | 0.70 (0.69-0.70) | 0.02 (0.02-0.02) | 1.00 (0.99-1.00) |
|  | Old | 0.74 (0.73-0.74) | 0.07 (0.06-0.07) | 0.64 (0.62-0.66)* | 0.72 (0.70-0.75) | 0.03 (0.02-0.03)* | 0.99 (0.99-0.99)* |
|  | Female | 0.74 (0.74-0.75) | 0.06 (0.06-0.07) | 0.67 (0.65-0.71) | 0.71 (0.67-0.74) | 0.02 (0.02-0.03) | 1.00 (1.00-1.00) |
|  | Male | 0.73 (0.72-0.74) | 0.07 (0.07-0.08)* | 0.62 (0.58-0.65)* | 0.73 (0.71-0.75) | 0.03 (0.02-0.03)* | 0.99 (0.99-0.99) |
|  | Black | 0.76 (0.76-0.78) | 0.07 (0.06-0.08) | 0.67 (0.64-0.68) | 0.74 (0.73-0.75) | 0.02 (0.02-0.03) | 1.00 (1.00-1.00) |
|  | Other | 0.74 (0.72-0.76) | 0.09 (0.07-0.12) | 0.66 (0.63-0.68) | 0.72 (0.69-0.77) | 0.03 (0.03-0.04) | 0.99 (0.99-0.99) |
|  | White | 0.73 (0.72-0.74)* | 0.07 (0.06-0.07) | 0.64 (0.59-0.68) | 0.71 (0.67-0.76) | 0.02 (0.02-0.03)* | 0.99 (0.99-0.99) |
| **CRRT** |  |  |  |  |  |  |  |
| **Development** |  |  |  |  |  |  |  |
|  | All | 0.89 (0.86-0.91) | 0.00 (0.00-0.01) | 0.83 (0.74-0.94) | 0.79 (0.67-0.87) | 0.00 (0.00-0.00) | 1.00 (1.00-1.00) |
|  | Young | 0.90 (0.88-0.92) | 0.01 (0.00-0.01) | 0.83 (0.79-0.90) | 0.83 (0.79-0.86) | 0.00 (0.00-0.00) | 1.00 (1.00-1.00) |
|  | Old | 0.87 (0.85-0.89)* | 0.00 (0.00-0.00)* | 0.89 (0.84-0.95) | 0.71 (0.68-0.77)* | 0.00 (0.00-0.00) | 1.00 (1.00-1.00) |
|  | Female | 0.88 (0.86-0.91) | 0.01 (0.00-0.01) | 0.85 (0.75-0.95) | 0.80 (0.72-0.87) | 0.00 (0.00-0.00) | 1.00 (1.00-1.00) |
|  | Male | 0.89 (0.88-0.90) | 0.00 (0.00-0.00)* | 0.84 (0.81-0.89) | 0.79 (0.71-0.83) | 0.00 (0.00-0.00) | 1.00 (1.00-1.00) |
|  | Black | 0.89 (0.86-0.92) | 0.00 (0.00-0.00) | 0.90 (0.85-0.95) | 0.80 (0.74-0.85) | 0.00 (0.00-0.00) | 1.00 (1.00-1.00) |
|  | Other | 0.85 (0.84-0.85) | 0.01 (0.01-0.01) | 0.91 (0.88-0.92) | 0.66 (0.65-0.66) | 0.00 (0.00-0.00) | 1.00 (1.00-1.00) |
|  | White | 0.91 (0.90-0.91)* | 0.00 (0.00-0.00) | 0.90 (0.85-0.98) | 0.77 (0.70-0.83)* | 0.00 (0.00-0.00) | 1.00 (1.00-1.00) |
| **External** |  |  |  |  |  |  |  |
|  | All | 0.88 (0.87-0.89) | 0.01 (0.01-0.01) | 0.83 (0.79-0.86) | 0.79 (0.76-0.82) | 0.00 (0.00-0.00) | 1.00 (1.00-1.00) |
|  | Young | 0.88 (0.87-0.88) | 0.01 (0.01-0.01) | 0.81 (0.79-0.86) | 0.81 (0.76-0.83) | 0.01 (0.00-0.01) | 1.00 (1.00-1.00) |
|  | Old | 0.87 (0.87-0.88)* | 0.01 (0.01-0.01) | 0.83 (0.81-0.87) | 0.78 (0.73-0.81) | 0.00 (0.00-0.00)* | 1.00 (1.00-1.00) |
|  | Female | 0.90 (0.89-0.90) | 0.01 (0.01-0.02) | 0.87 (0.87-0.88) | 0.79 (0.79-0.79) | 0.00 (0.00-0.00) | 1.00 (1.00-1.00) |
|  | Male | 0.87 (0.86-0.87)* | 0.01 (0.01-0.01) | 0.81 (0.79-0.83)* | 0.79 (0.78-0.81) | 0.00 (0.00-0.00) | 1.00 (1.00-1.00) |
|  | Black | 0.88 (0.86-0.89) | 0.01 (0.01-0.01) | 0.85 (0.79-0.89) | 0.79 (0.75-0.83) | 0.00 (0.00-0.01) | 1.00 (1.00-1.00) |
|  | Other | 0.89 (0.88-0.89) | 0.01 (0.01-0.01) | 0.87 (0.86-0.89) | 0.78 (0.77-0.79) | 0.01 (0.00-0.01) | 1.00 (1.00-1.00) |
|  | White | 0.88 (0.87-0.89) | 0.01 (0.01-0.01) | 0.81 (0.77-0.85)* | 0.81 (0.77-0.83) | 0.00 (0.00-0.00)* | 1.00 (1.00-1.00) |
| **Temporal** |  |  |  |  |  |  |  |

| | | | | | | |
|---|---|---|---|---|---|---|
| All | 0.96 (0.95-0.96) | 0.07 (0.06-0.08) | 0.91 (0.88-0.95) | 0.90 (0.86-0.92) | 0.02 (0.02-0.03) | 1.00 (1.00-1.00) |
| Young | 0.96 (0.95-0.96) | 0.06 (0.06-0.07) | 0.89 (0.88-0.90) | 0.92 (0.91-0.92) | 0.03 (0.02-0.03) | 1.00 (1.00-1.00) |
| Old | 0.96 (0.95-0.96) | 0.07 (0.06-0.08) | 0.91 (0.89-0.92)* | 0.89 (0.89-0.91)* | 0.02 (0.02-0.03)* | 1.00 (1.00-1.00) |
| Female | 0.96 (0.96-0.96) | 0.07 (0.06-0.08) | 0.93 (0.91-0.94) | 0.89 (0.88-0.91) | 0.02 (0.02-0.02) | 1.00 (1.00-1.00) |
| Male | 0.95 (0.95-0.96)* | 0.07 (0.06-0.07) | 0.90 (0.86-0.94) | 0.89 (0.85-0.92) | 0.02 (0.02-0.03) | 1.00 (1.00-1.00) |
| Black | 0.97 (0.96-0.97) | 0.08 (0.07-0.08) | 0.94 (0.92-0.96) | 0.90 (0.87-0.91) | 0.02 (0.02-0.03) | 1.00 (1.00-1.00) |
| Other | 0.96 (0.94-0.97) | 0.09 (0.06-0.11) | 0.94 (0.88-0.98) | 0.88 (0.87-0.91) | 0.03 (0.02-0.03) | 1.00 (1.00-1.00) |
| White | 0.96 (0.95-0.96)* | 0.06 (0.06-0.07)** | 0.91 (0.87-0.93) | 0.89 (0.87-0.92) | 0.02 (0.02-0.03)* | 1.00 (1.00-1.00) |

Abbreviations: AUPRC: area under the precision-recall curve; AUROC: area under the receiver operating characteristic curve; CRRT: Continuous Renal Replacement Therapy; MV: Mechanical Ventilation; NPV: negative predictive value PPV: positive predictive value; VP: Vasopressors.

Metrics are shown as the median across 100-iteration bootstrap with 95% Confidence Intervals in parenthesis. P-values are based on pairwise Wilcoxon rank sum tests, with age, gender, and race groups being compared amongst them. * p-value < 0.05 compared to one group within its respective category. ** p-value < 0.05 compared to two groups within its respective category.

# SUPPLEMENTARY FIGURES

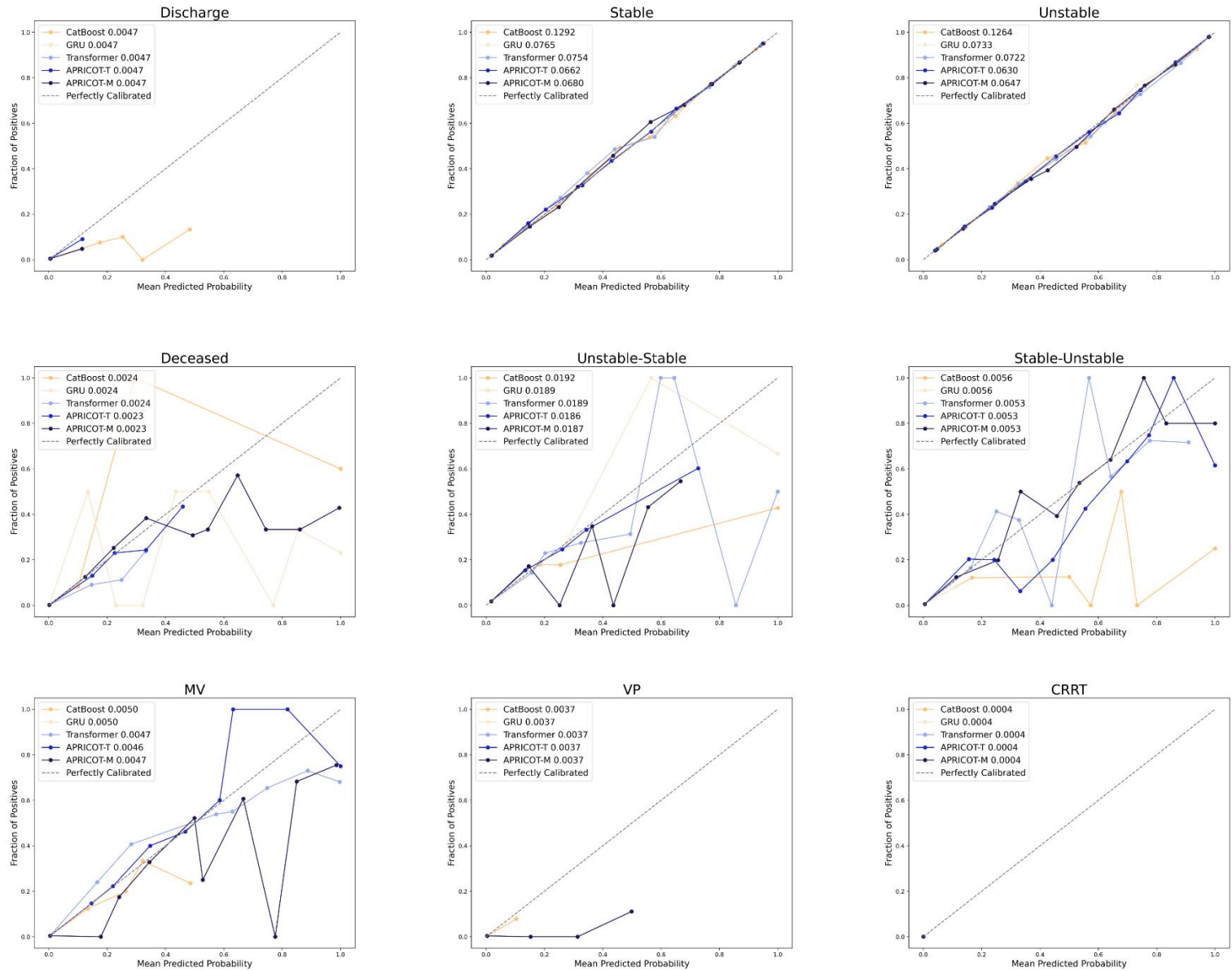

**Supplementary Figure S1. Calibration curve for all models in development cohort.**

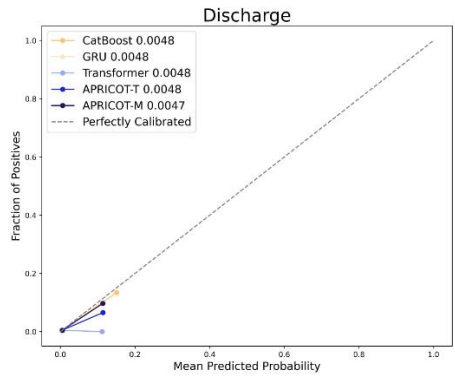
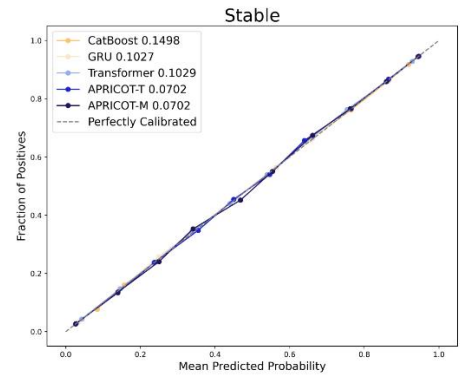
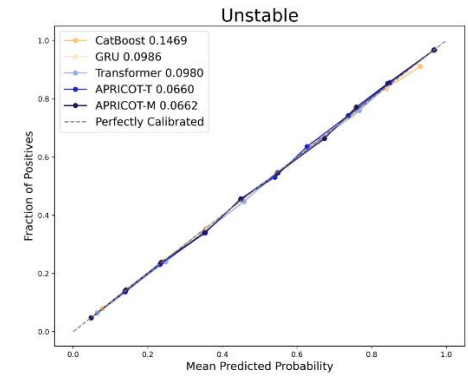
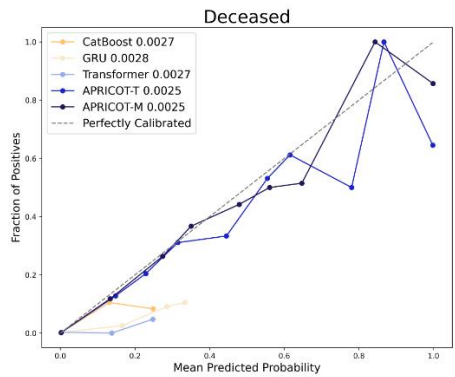
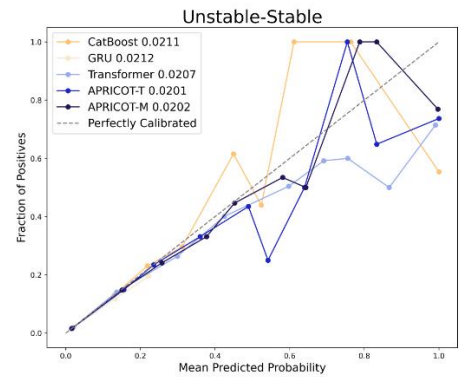
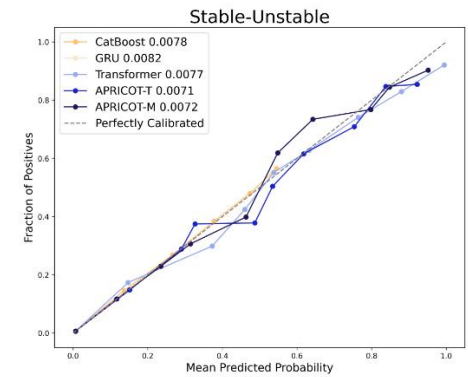
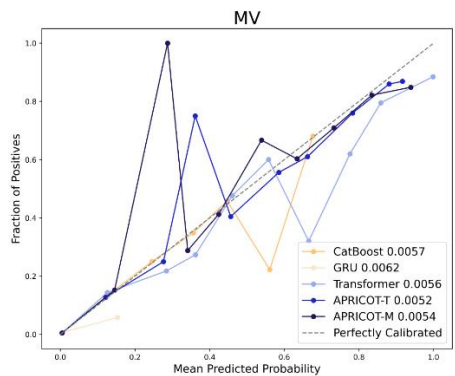
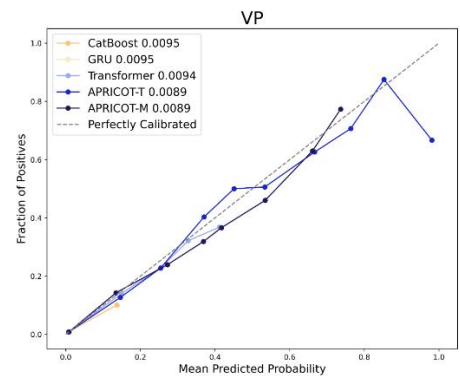
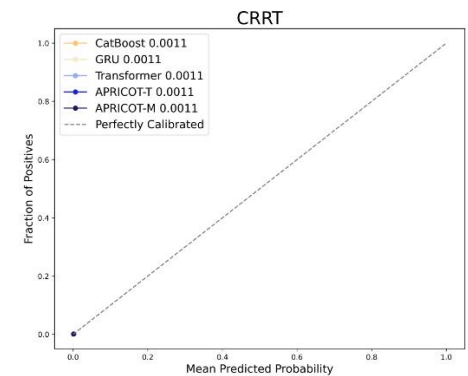

**Supplementary Figure S2. Calibration curve for all models in external cohort.**

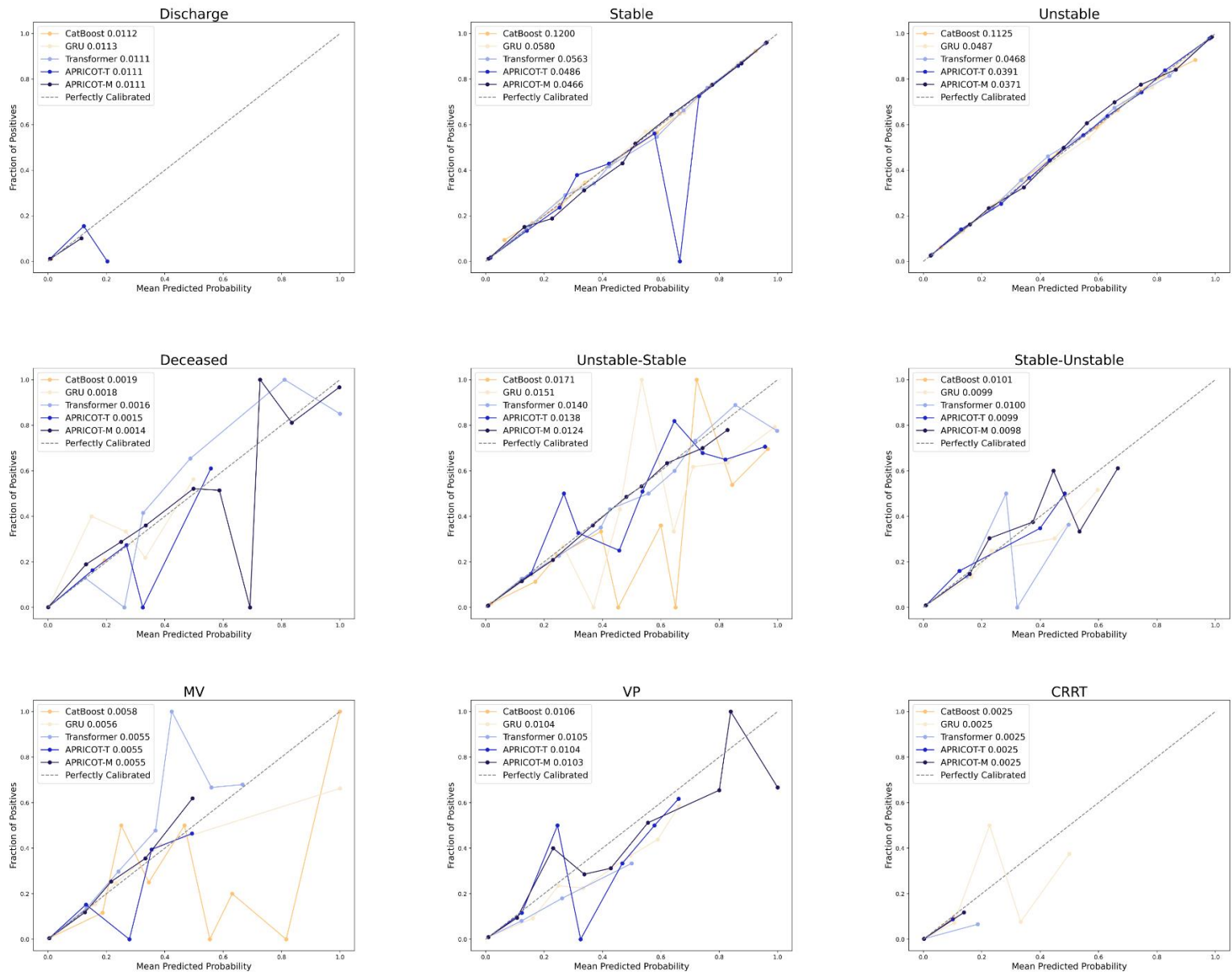

**Supplementary Figure S3. Calibration curve for all models in temporal cohort.**

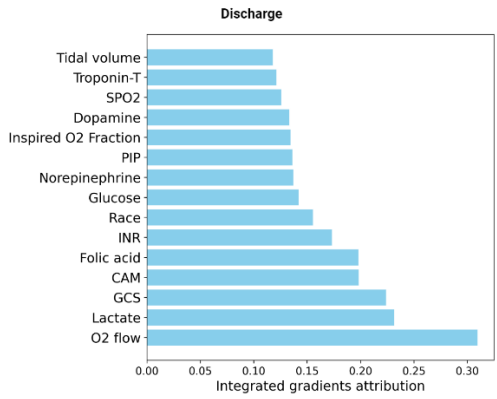 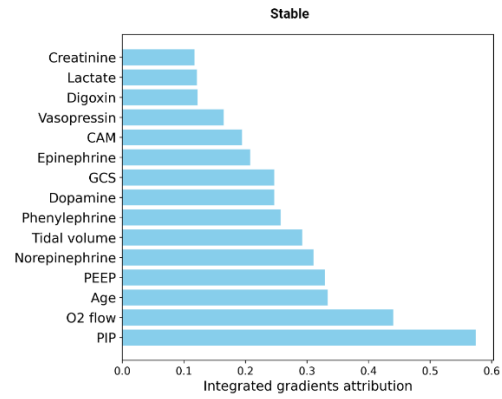 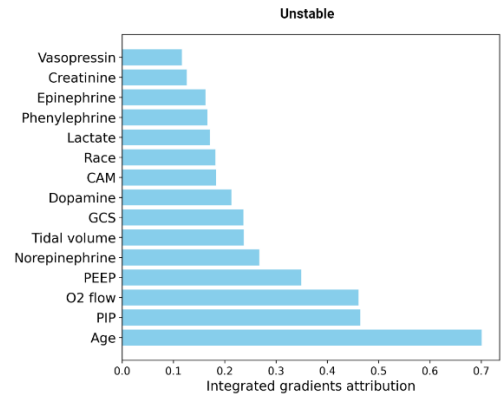
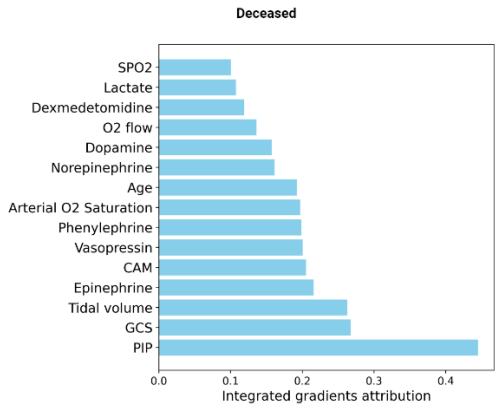 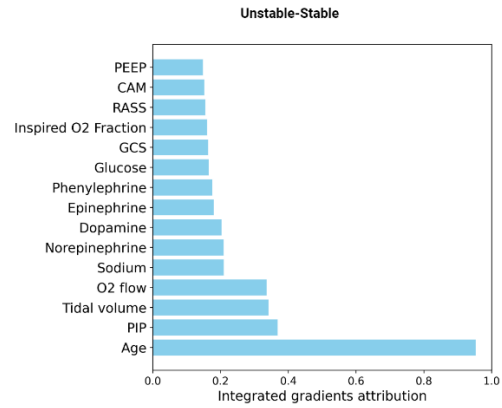 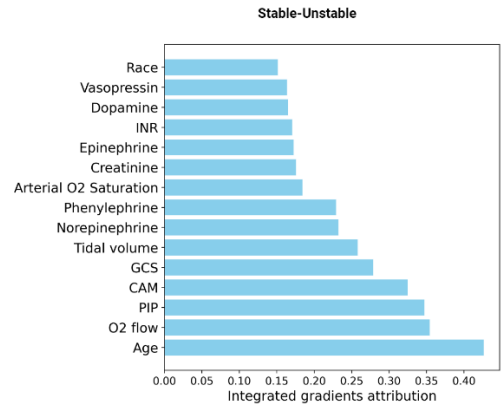
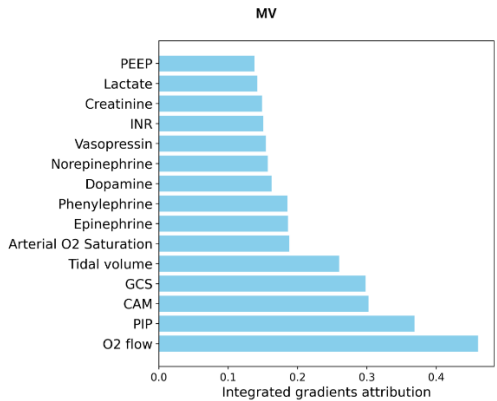 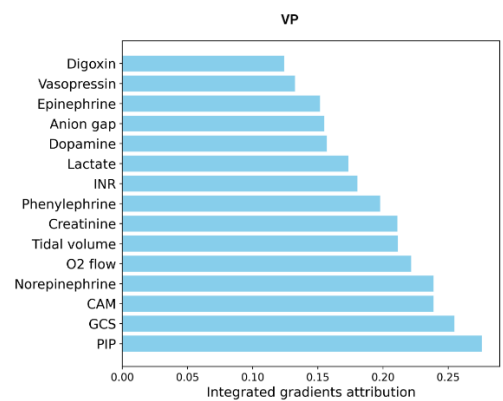 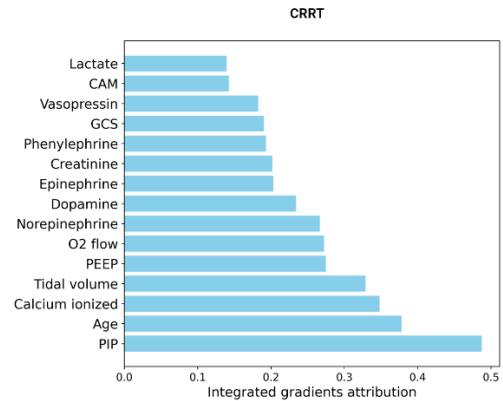

**Supplementary Figure S4. Development cohort most relevant features according to integrated gradients attributions for each prediction task.**

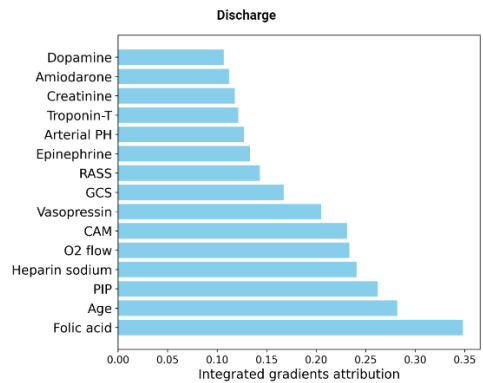
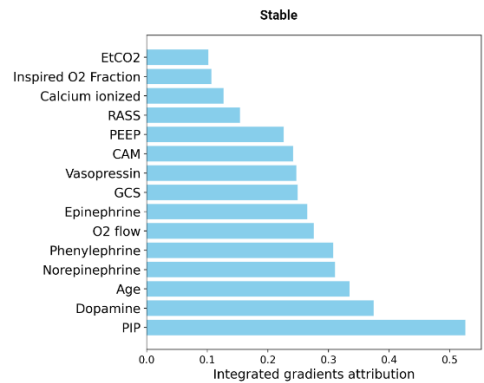
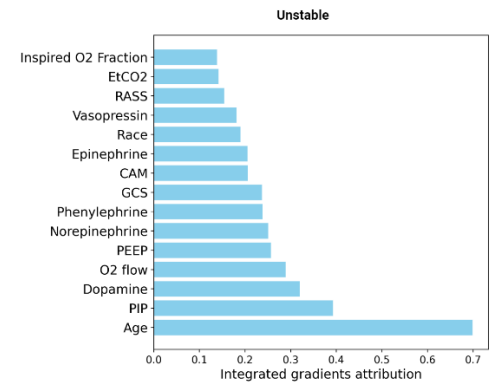
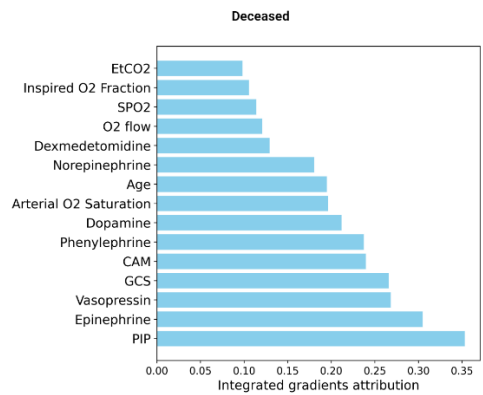
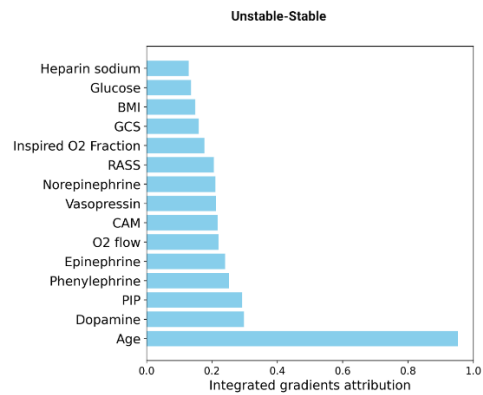
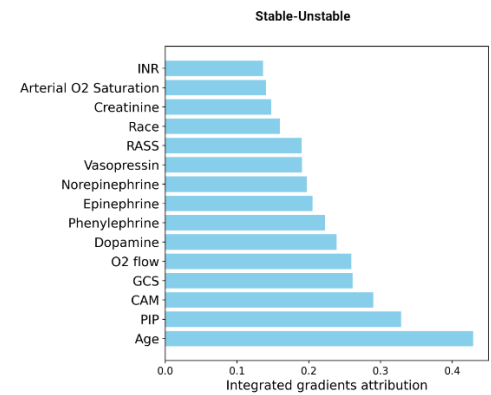
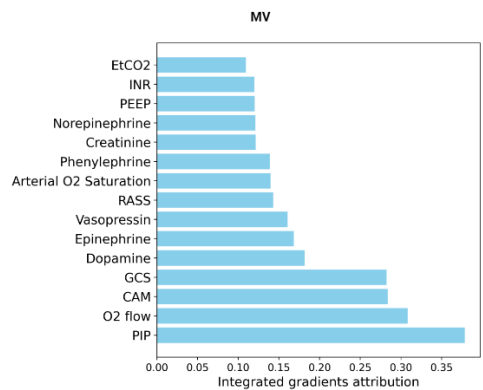
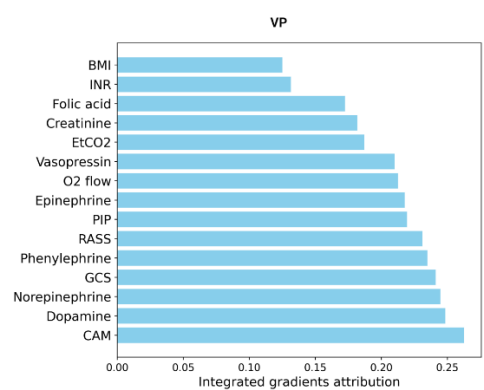
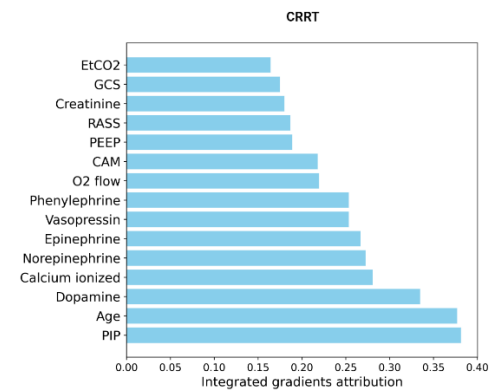

**Supplementary Figure S5. External cohort most relevant features according to integrated gradients attributions for each prediction task.**

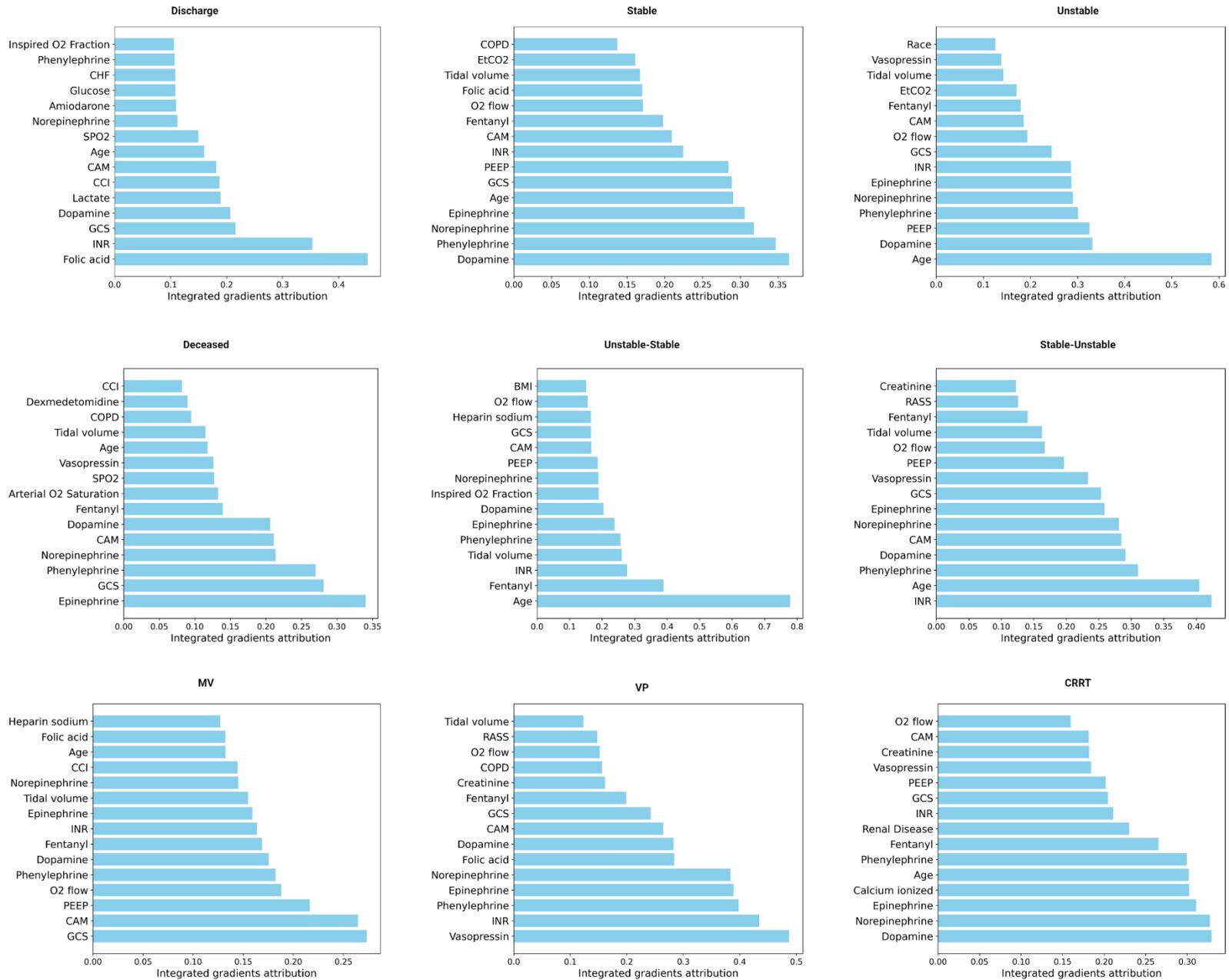

**Supplementary Figure S6. Temporal cohort most relevant features according to integrated gradients attributions for each prediction task.**

# TRIPOD Checklist: Prediction Model Development and Validation

| Section/Topic | | | Checklist Item | Page |
|---|---|---|---|---|
| **Title and abstract** | | | | |
| Title | 1 | D;V | Identify the study as developing and/or validating a multivariable prediction model, the target population, and the outcome to be predicted. | 1 |
| Abstract | 2 | D;V | Provide a summary of objectives, study design, setting, participants, sample size, predictors, outcome, statistical analysis, results, and conclusions. | 1 |
| **Introduction** | | | | |
| Background and objectives | 3a | D;V | Explain the medical context (including whether diagnostic or prognostic) and rationale for developing or validating the multivariable prediction model, including references to existing models. | 2 |
| | 3b | D;V | Specify the objectives, including whether the study describes the development or validation of the model or both. | 2-3 |
| **Methods** | | | | |
| Source of data | 4a | D;V | Describe the study design or source of data (e.g., randomized trial, cohort, or registry data), separately for the development and validation data sets, if applicable. | 3-4 |
| | 4b | D;V | Specify the key study dates, including start of accrual; end of accrual; and, if applicable, end of follow-up. | 3-4 |
| Participants | 5a | D;V | Specify key elements of the study setting (e.g., primary care, secondary care, general population) including number and location of centres. | 3-4 |
| | 5b | D;V | Describe eligibility criteria for participants. | 3-4 |
| | 5c | D;V | Give details of treatments received, if relevant. | N/A |
| Outcome | 6a | D;V | Clearly define the outcome that is predicted by the prediction model, including how and when assessed. | 6-7 |
| | 6b | D;V | Report any actions to blind assessment of the outcome to be predicted. | N/A |
| Predictors | 7a | D;V | Clearly define all predictors used in developing or validating the multivariable prediction model, including how and when they were measured. | 6-7, S1 |
| | 7b | D;V | Report any actions to blind assessment of predictors for the outcome and other predictors. | N/A |
| Sample size | 8 | D;V | Explain how the study size was arrived at. | 3-4 |
| Missing data | 9 | D;V | Describe how missing data were handled (e.g., complete-case analysis, single imputation, multiple imputation) with details of any imputation method. | 6-7 |
| Statistical analysis methods | 10a | D | Describe how predictors were handled in the analyses. | 8-9 |
| | 10b | D | Specify type of model, all model-building procedures (including any predictor selection), and method for internal validation. | 9-10 |
| | 10c | V | For validation, describe how the predictions were calculated. | 9 |
| | 10d | D;V | Specify all measures used to assess model performance and, if relevant, to compare multiple models. | 9, 12 |
| | 10e | V | Describe any model updating (e.g., recalibration) arising from the validation, if done. | 11 |
| Risk groups | 11 | D;V | Provide details on how risk groups were created, if done. | 11 |
| Development vs. validation | 12 | V | For validation, identify any differences from the development data in setting, eligibility criteria, outcome, and predictors. | 4-5 |
| **Results** | | | | |
| Participants | 13a | D;V | Describe the flow of participants through the study, including the number of participants with and without the outcome and, if applicable, a summary of the follow-up time. A diagram may be helpful. | 9-10 |
| | 13b | D;V | Describe the characteristics of the participants (basic demographics, clinical features, available predictors), including the number of participants with missing data for predictors and outcome. | 12-13, S2-21 |
| | 13c | V | For validation, show a comparison with the development data of the distribution of important variables (demographics, predictors and outcome). | 12-13, S2-21 |
| Model development | 14a | D | Specify the number of participants and outcome events in each analysis. | 12-13 |
| | 14b | D | If done, report the unadjusted association between each candidate predictor and outcome. | S2-21 |
| Model specification | 15a | D | Present the full prediction model to allow predictions for individuals (i.e., all regression coefficients, and model intercept or baseline survival at a given time point). | |
| | 15b | D | Explain how to the use the prediction model. | |
| Model performance | 16 | D;V | Report performance measures (with CIs) for the prediction model. | 10-11, S17-21 |
| Model-updating | 17 | V | If done, report the results from any model updating (i.e., model specification, model performance). | N/A |
| **Discussion** | | | | |
| Limitations | 18 | D;V | Discuss any limitations of the study (such as nonrepresentative sample, few events per predictor, missing data). | 25-26 |
| Interpretation | 19a | V | For validation, discuss the results with reference to performance in the development data, and any other validation data. | 24-25 |
| | 19b | D;V | Give an overall interpretation of the results, considering objectives, limitations, results from similar studies, and other relevant evidence. | 24-26 |
| Implications | 20 | D;V | Discuss the potential clinical use of the model and implications for future research. | 26 |
| **Other information** | | | | |
| Supplementary information | 21 | D;V | Provide information about the availability of supplementary resources, such as study protocol, Web calculator, and data sets. | S1-42 |
| Funding | 22 | D;V | Give the source of funding and the role of the funders for the present study. | 26 |